%% file: main.tex
\documentclass{book}
\include{kapitel/header}

\begin{document} 

\selectlanguage{english}
 
\include{kapitel/titelseite}
\blankpage
\pagenumbering{roman} 
\tableofcontents
\cleardoublepage
\pagenumbering{arabic}
\include{kapitel/einleitung}
\include{kapitel/kapitel2}
\include{kapitel/gdl}

\include{kapitel/gnn}
 
\include{kapitel/invariantswhy}
\include{kapitel/localpooling}
\include{kapitel/conclusion	}

\appendix
\include{kapitel/anhang}
\listoffigures
\addcontentsline{toc}{chapter}{List of Figures}
\cleardoublepage

\bibliographystyle{gerplain}
\bibliography{literatur/diplom}
\addcontentsline{toc}{chapter}{\bibname}
\thispagestyle{myheadings}
\markboth{}{ERKLÄRUNG}
\addcontentsline{toc}{chapter}{Declaration}
\cleardoublepage
\end{document}

%% file: kapitel/header.tex
\usepackage[a4paper,left=3.5cm,right=2.5cm,bottom=3.5cm,top=3cm]{geometry}
\usepackage[T1]{fontenc}

\usepackage[german,english]{babel}
\usepackage{tikz} 
\usepackage{amsmath,amssymb,subfig}
\usepackage{svg}
\newcommand{\mybox}[2]{
    \begin{figure}[h] 
        \centering
    \begin{tikzpicture} 
        \node[anchor=text,text width=\columnwidth-1.2cm, draw, rounded corners, line width=1pt, fill=gray!20, inner sep=5mm] (big) {\\#2};
        \node[draw, rounded corners, line width=.5pt, fill=gray!60, anchor=west, xshift=5mm] (small) at (big.north west) {#1};
    \end{tikzpicture}
    \end{figure}
} 

\DeclareMathOperator*{\argmin}{arg\,min}
\DeclareMathOperator{\grp}{\mathbb{G}}
\DeclareMathOperator*{\orbit}{orbit}
\DeclareMathOperator*{\gcong}{\cong_{\grp}}
\DeclareMathOperator{\Ex}{\mathbb{E}}

\DeclareMathOperator{\ft}{f_{\Theta}}
\DeclareMathOperator{\sig}{sig}
\DeclareMathOperator{\real}{\mathbb{R}}
\DeclareMathOperator*{\rep}{rep}
\DeclareMathOperator{\softmax}{softmax}

\newcommand\dist[1]{\lvert\lvert{#1} \rvert\rvert}
\newcommand\ddx[2]{\frac{\partial {#1}}{\partial {#2}}}

\usepackage[amsmath,thmmarks]{ntheorem}

\usepackage[utf8]{inputenc}
\usepackage[T1]{fontenc}

\usepackage[plain,chapter]{algorithm}
\usepackage{algorithmic}

\usepackage{enumerate}

\usepackage{bibgerm}

\usepackage{url}

\usepackage[margin=0pt,font=small,labelfont=bf]{caption}

\theoremseparator{.}
\theoremstyle{change}

\theoremstyle{nonumberplain}
\renewtheorem{theorem*}{Theorem}
\renewtheorem{satz*}{Satz}
\renewtheorem{lemma*}{Lemma}
\renewtheorem{korollar*}{Korollar}
\renewtheorem{proposition*}{Proposition}
\theorembodyfont{\upshape}
\theoremstyle{change}

\theoremstyle{nonumberplain}
\renewtheorem{definition*}{Definition}
\theoremheaderfont{\itshape}

\theoremsymbol{\ensuremath{\Box}}

\theoremsymbol{}
\theoremstyle{change}
\theoremheaderfont{\bfseries}

\theoremstyle{nonumberplain}
\renewtheorem{bemerkung*}{Bemerkung}
\renewtheorem{beispiel*}{Beispiel}
\renewtheorem{problem*}{Problem}

\floatname{algorithm}{Algorithmus}

\usepackage{xspace}

\newcommand{\blankpage}{
 \clearpage{\pagestyle{empty}\cleardoublepage}
}

%% file: kapitel/titelseite.tex
\begin{titlepage}
\definecolor{TUGreen}{rgb}{0.517,0.721,0.094}
\vspace*{-2cm}
\newlength{\links}
\setlength{\links}{-1.5cm}
\sffamily
\hspace*{\links}
\begin{minipage}{12.5cm}
\includegraphics[width=8cm]{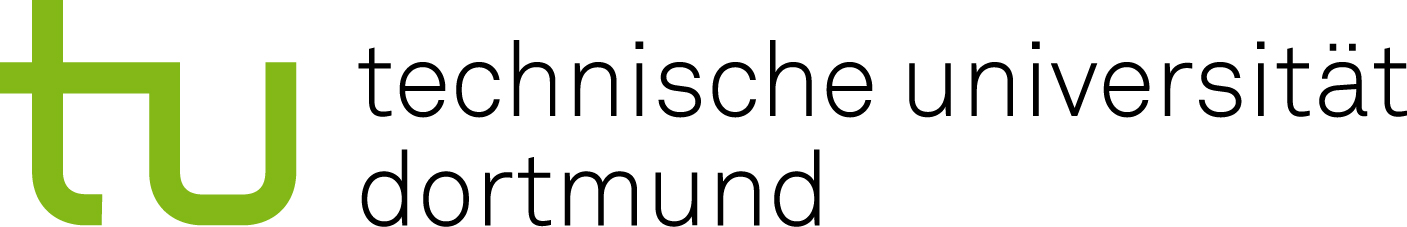}
\end{minipage}

\vspace*{4cm}

\hspace*{\links}
\hspace*{-0.2cm}
\begin{minipage}{9cm}
\large
\begin{center}
{\Large Master's Thesis} \\
\vspace*{1cm}
\textbf{A Structural Approach to the Design of Domain Specific Neural Network Architectures} \\
\vspace*{1cm}
Gerrit Nolte\\
December 2021\end{center}
\end{minipage}
\normalsize
\vspace*{5.5cm}


\vspace*{2.1cm}

\hspace*{\links}
\begin{minipage}[b]{5cm}
\raggedright
Advisors: \\
Prof. Dr. Bernhard Steffen \\
Alnis Murtovi, M.Sc.\\
\end{minipage}

\vspace*{2.5cm}
\hspace*{\links}
\begin{minipage}[b]{8cm}
\raggedright
Technische Universität Dortmund \\
Fakultät für Informatik\\
Lehrstuhl für Programmiersysteme (5)\\
http://ls5-www.cs.tu-dortmund.de
\end{minipage}
\begin{minipage}[b]{8cm}

\end{minipage}

\end{titlepage}

%% file: kapitel/einleitung.tex
\chapter{Introduction and Motivation}
Few fields have made rapid advances in the way that the field of deep learning has managed to in the last decades. One of the reasons for this advancement certainly lies in the rise of ever more powerful parallel hardware that supports neural networks very well. However, arguably even more important advancements have been made in the design of neural network models themselves.\\
Instead of relying on the, at this point mostly historic, fully connected neural networks, better architectures dominate in practice. For example, convolutional neural networks achieve state of the art results in image recognition \cite{dai2021coatnet}. In language processing tasks, transformers dominate \cite{wolf2019huggingface} and in graph based tasks, graph neural networks are the model of choice \cite{scarselli2008graph}.\\
The powerful new idea behind these architectures is the idea of using neural networks that are specifically made for the domain they are supposed to be used in. While the ancestors of modern neural networks, fully connected neural networks, can be used for all of these tasks in principle, practice demonstrates that neural networks that were build by using the explicit human knowledge of domain experts heavily outperform the architectures that were not build with domain knowledge in mind. As a consequence, the idea of neural networks as general purpose tools seems outdated, instead, one requires a well thought out model for one's specific task to match modern results.\\
However, while domain specific neural networks have enjoyed large success in practice, they are still faced with the problem that building \textit{good} neural networks for a given domain is a terrifyingly difficult task. For example, in the case of graph neural networks the road from their first theoretical conception to widespread adoption took almost a decade and many refinements to the model itself \cite{zhou2020graph, scarselli2008graph}. The reason for this lies in the fact, that, to build a good domain specific neural network, one needs to be an expert both of the domain and of deep learning as a whole.\\
As a consequence, much is missing in the creation of domain specific neural networks.
First, there does not seem to exist a notion of what actually makes a domain specific neural network \textit{good} and what principles it should adhere to. Second, as a consequence, there does not yet exist any structured approach to the design of neural networks that adhere to said principle, much less an actually automatic approach.\\
Thus, the questions that require answering are: ''What do good domain specific neural networks look like?'' and ''How can we construct them?''.\\
The authors of \cite{geo} aim to make a first step towards a structured approach to creating domain specific neural networks which they call ''geometric deep learning''\footnote{This name was most likely chosen due to the origins of this work in geometric computer graphics tasks.}. In doing so, they propose both some criteria that they believe are common to successful domain specific neural networks and an attempt at a somewhat more structured approach to the design of domain specific neural networks.\\
It is the aim of this work to discuss the ideas of \cite{geo}. We want to evaluate, how we can answer the two core questions of this work using the ideas of geometric deep learning and how they impact the performance of neural networks.
\section{Structure of this Work}
The core questions of this work are: ''what do good domain specific neural networks look like?'' and ''how can we construct them'' and how can geometric deep learning help in this regard. Of course, we have to find an answer to the first one before we can move on to the second.\\
This work will be split into four different parts.\\
To answer the question ''what do good domain specific neural networks look like?'' we first have to clarify the setting in which we aim to apply them. Thus, the first chapter is intended to clearly define the general setting that we are working with and make explicit many of the assumptions that are usually implicitly contained in work on deep learning.\\
With the background set, the second chapter will introduce the criteria of geometric deep learning that the authors believe are characteristic for good domain specific neural networks. To do this, we will first take a look at the most non-domain specific neural network, the fully connected neural network, and identify its strengths, that we would like to keep for domain specific neural networks, and its shortcomings that geometric deep learning seeks to fix. Then, we will have a look at the  main ideas of deep learning, three criteria that the authors believe should guide the design of domain specific neural networks. These criteria are ''Task Separation'', ''Locality'' and ''Invariance''.\\ Afterwards, we will start dealing with the question of how neural networks that adhere to those principles can be built. To do this, we will first showcase these principles on the example of deep sets, a domain specific neural network architecture for sets that could be considered, and from there motivate the geometric deep learning blue print, the proposed approach to the design of domain specific neural networks in the sense of geometric deep learning.\\
With the ideas of geometric deep learning introduced, we will apply them in a case study to the example of graph neural networks. Here, we will evaluate both whether graph neural networks fit the principles of geometric deep learning and whether the geometric deep learning blueprint would be useful in deriving them. In this chapter, we hope to evaluate how practical it is to construct neural networks that adhere to the geometric deep learning blueprint. Graph neural networks go beyond the case study of deep sets as they are much more complex and a more mature model.\\
Lastly, in the fifth chapter, we want to discuss the impact of the ideas of geometric deep learning, specifically the idea of invariances theoretically. To get a complete picture, we will view these impacts from three different angles. In this chapter, we seek to evaluate whether the effort of building neural networks that adhere to the geometric deep learning blueprint is actually justified by the performance benefits of the neural networks that we attain.\\
This work will close with a final summary, discussion and an outlook regarding new research avenues in the field of domain specific deep learning architectures.

%% file: kapitel/kapitel2.tex
\chapter{Background and General Setting}
\label{approx}
The core question of this work is ''what do \textit{good} neural network architectures look like and how do we build them''. But before we can answer the question of ''what is a good neural network'', we first have to define what the terms \textit{good} and \textit{neural network} actually mean, which is deceptively hard to answer in both cases.\\
Due to the high level of generality, deep learning approaches can be used in a myriad of different ways, from classical supervised learning to unsupervised learning and reinforcement learning \cite{dl,dayan2008reinforcement}.\\
This makes it almost impossible to make any meaningful statement about the quality of any deep learning architecture. After all, it is very hard to disprove that there does not exist \textit{some} context in which any architecture might be useful and most definitions of terms in deep learning are too fuzzy to properly evaluate.\\
Thus, we first have to clearly define the problem that we want to solve, so that we can make an argument whether neural network architectures work well in this task. Thus, this work will focus on the perhaps most basic and well understood use case of neural networks, supervised learning or, more succinctly put, function approximation.\\
The following sections are heavily based on the works of \cite{dl} and \cite{minka2000bayesian} for deep learning and statistical concepts respectively.\\
The general setting that we consider is as follows:\\
Suppose there exists high dimensional space of data $\Omega$, i.e. some space of data that can be encoded as $\real^i$ for $i$ large, some space of possible classification results $C$ and an unknown function $f^*: \Omega \rightarrow C$ belonging to some known class of function $F$ that is to be approximated. Note that deep learning generally assumes that $\Omega$ and $C$ are real vector spaces. If they are not, additional techniques are needed to transform them into real vector spaces. Techniques to do this are discussed in the appendix. Further, let $D=\{(x_i,y_i=f^*(x_i) \}$ be some set sample points (also: training set) that are known. The task is now to find a function $f$ that most closely resembles $f^*$ using the data $D$. Depending on the context, this can mean multiple different things as we will later elaborate on more in-depth.\\
In the context of this work, deep learning is understood as a subdivision of the more general field of function approximation and neural networks are more well understood as a class of approximators that work in the very general, well understood framework of function approximation.\\ 
Thus, in the following sections, we will first review fundamental concepts of function approximation and statistical modeling before introducing neural networks, a type of function estimator that is particularly well suited to some problems.\\
\mybox{General Setting}{Our general setting is this: Estimate a function $f^* : \Omega \rightarrow C$ with a high dimensional domain $\Omega$ given a set $D$ containing some samples of $f^*$.}
\section{Parameterized Estimators, Risk and Loss Functions}
\label{losses}
\begin{figure}[htbp]
\centerline{\includegraphics[scale=.25]{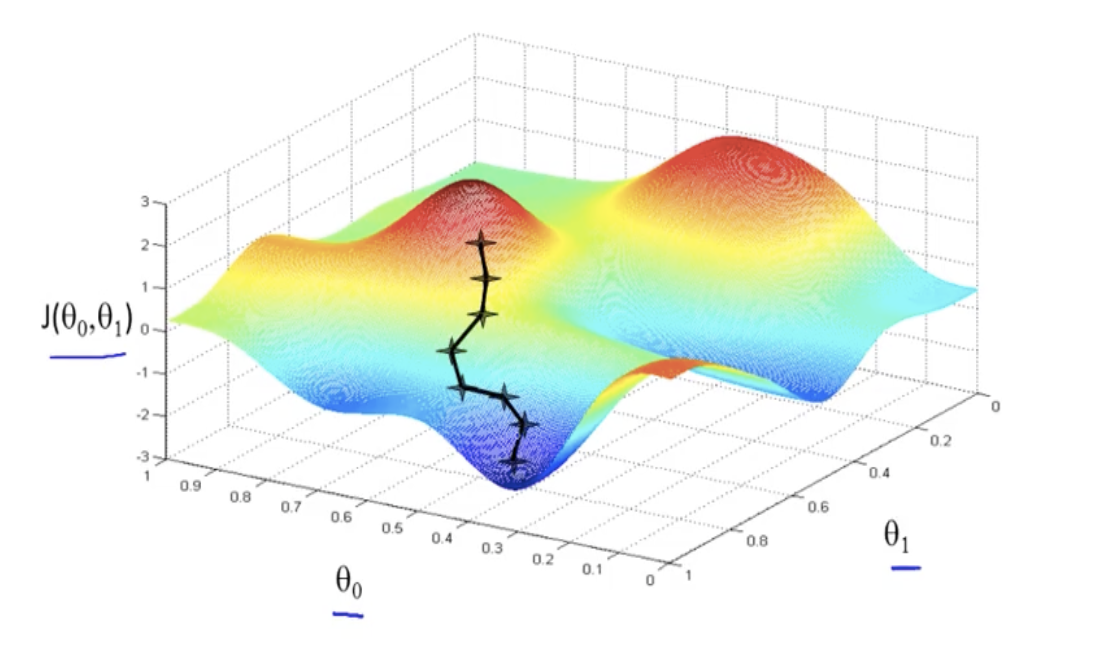}}
\caption{Gradient descent running on a simple example two dimensional function. Note the greedy nature of gradient based optimization, always tending to the nearest local minimum and taking steps in the largest local direction of improvement. Example taken from \cite{gdesc}}.
\label{WL}
\end{figure}
If one is presented with the task described in the previous section, one usually chooses an estimator that they then want to fit to the training data $D$. Intuitively, one apriori chooses a family of candidate functions  $F$ and, after seeing the data, chooses one function $f \in F$ that one believes is best for the task at hand. For example, if one chooses a linear regression model, $F$ corresponds to the family of linear functions $\Omega \rightarrow C$ and the chosen function $f$ after observing data $D$ is given as the one linear function that minimizes the average squared distance of samples and predictions.\\
There exist a multitude  of very well understood and efficient methods to do this such as the polynomial interpolation or spline interpolation to approximate a function from samples \cite{de1990multivariate, carslaw1921introduction}. These methods are generally much more interpretable and controllable than neural networks are. However, these more theoretically well founded approaches tend to suffer from an exponential growth in runtime as the number of dimensions increases. For example, the necessary number of terms for both spline interpolation and multivariate polynomial interpolation grows exponentially with regards to the number of dimensions one considers \cite{habermann2007multidimensional, gasca2000polynomial}. Therefore, we seek an approach that can handle a large number of dimensions, and input samples for that matter, without scaling exponentially in either memory or time. One way to achieve this is by the usage of parameterized estimators and gradient based training. \\
For gradient based estimator training \cite{dl}, one considers parameterized estimator classes. That is, one assumes that there exists a parameter space $\Theta=R^q$ such that the space of possible estimators $F$ can be written as $F = \{ f_{\theta} \mid  \theta \in \Theta \}$.  With this, training is simplified: we just have to choose the right parameters such that we obtain a good $f_{\theta}$. Thus, memory usage is guaranteed to remain constant through training when one uses parameterized estimators as we only need to keep $\theta$ in memory and the size of $\theta$ is unchanging.\\
With a slight abuse of notation, we will write $f$ both for a concrete estimator with a fixed parameterset $f_{\theta}$ and also sometimes for the class of estimators $F$ induced by changing the parameters, for example when referring to the functions $f$ could approximate. This matches the intuition behind most estimators: First, one defines an empty model without parameters and then assigns through some training process parameters to it.\\ 
If the context does not make this distinction obvious, we will use $f_*=F$ to refer to the unparameterized estimator and the class of parameterized estimators that result from it, and $f_{\theta}$ to refer to an estimator with a fixed set of parameters $\theta$.\\
As an example of parameterized estimators, let us consider the very simple case of linear models. A linear estimator \cite{minka2000bayesian} is given as:
\begin{equation}
f_{\theta}(x)= \theta^T x
\end{equation}
Linear models are very easy to interpret. The prediction of a linear model is the sum of its inputs where the $i$-th input is weighted by the parameter $\theta_i$. In this case, the possible approximable functions are given $f_*=\{ f \; \mid \; f \in \real^n \rightarrow \real^m, \text{f linear}\}$ and one concrete estimator $f_{\theta}$ is just one singular linear function.\\
Now we will consider the application of linear estimators to approximate a function $f^*$ based on some samples $D=\{(x_i, y_i=f^*(x_i)\}$. First off, we have to first define what the estimator function should ideally be like. Usually, this is done by defining a loss function $L(y,\hat{y})$ that describes how costly a prediction of $\hat{y}$ is when the actual result would have been $y$. Then, we want our estimator to have the lowest expected loss over all possible datasets, alternatively called risk or generalization error, among all possible estimators.
\begin{equation}
R(f,f^*)=\Ex_{x \in \Omega}(L(f(x), f^*(x) )
\end{equation}
If one could find a function $f$ that minimizes this risk function, one would obtain the ideal estimator with respect to the loss function $L$. Unfortunately, $R$ is almost never tractable in practice as we do not have access to $f^*$ and the true data generating distribution $P(x)$. However, if one assumes that $D$ was sampled according to the true data generating distribution, i.e. for each $(x_i, y_i) \in D$, $x_i$ was sampled from $P(x)$, we can approximate the true risk with the empiric risk or training error:
\begin{equation}
\hat{R}(f,f^*)=\sum_{(x_i,y_i) \in D}(L(f(x_i), y_i )
\end{equation}
One might at first assume that this is a very good estimator for the risk $R$ as it generally holds that the approximation of an expected value has variance proportional to $1/n$ where $n$ is the number of samples. However, while this is true, to correctly determine $\argmin_f R(f,f^*)$ we need an estimator for all possible estimators $f$. It is of course very unlikely that our estimator is reliable for all function values $f$ and it is thus not necessarily the case that optimizing $\hat{R}$ corresponds well with optimizing $R$. Therefore, an estimator that achieves a good $\hat{R}$, i.e., performs well on the training data, is not necessarily a good estimator with respect to $R$, i.e. the general performance on the entire task, and building one that performs well on the entire possible dataset is a very hard problem and there does not exist a general way to do it.
\subsection{Determining a Sensible Loss Function}
Of course, it is of critical importance to construct a sensible loss function that one optimizes. If the loss function one chose does not adequately describe the costs of a wrong prediction, the resulting estimator will necessarily be unfit for ones task even if the optimization process works perfectly. There is however a very well understood and structured way of constructing such a loss function that is based in the concept of a maximum likelihood estimator \cite{eliason1993maximum}.\\
This approach works by adding some stochastic noise to our estimator.Thus, an estimator is not just a function, but rather a probability density function, giving not just one estimation $f(x)$ but a density describing $P(y|x)$. 
The idea behind this is the following: If our estimator is not random, then our guess for $f_{\theta}$ is only ever true or false (either $f(x)=f^*(x)$ or not). This binary classification makes learning hard, as small changes to the estimator tend not to change it. However, for a stochastic estimator, $P(y|x)$ is generally non-zero. Therefore, this gives an idea of \textit{how likely} an estimator is to be true, which gives a direction for optimization.\\
Returning to our example of a linear classifier, we can add some normally distributed noise $\epsilon \sim N(0, \sigma)$ to the classifier and attain
\begin{equation}
P(y|x,\theta)= N(\theta^T x,\sigma) \propto e^{\frac{-1}{2\sigma^2}(x- \theta^T x)^2}
\end{equation}
which is the standard presentation of the linear regression model.\\
As mentioned before, this allows us to not only quantify whether our predictor was wrong on some task but quantify how far it was off the true target\footnote{It is of course necessary to find a good guess as to what noise makes sense to assume based on the task. Usually, this depends on the type of data, i.e. normal distributions for continuous tasks and binomial distributions for discrete tasks.}. Now, a maximum likelihood estimator is any estimator $f$ that maximizes the probability of observing the data if it was generated by $f_{\theta}$ and not $f^*$
\begin{equation}
P(D|f) = \Pi_{(x_i,y_i) \in D} P(y_i|x_i,f)
\end{equation}
It can be shown that if the number of samples increases towards infinity and our assumptions on $f^*$ hold, i.e. the error $\epsilon_f$ that we assume to be added to our predictor is distributed equivalently to the noise in our data, the maximum likelihood estimator converges to the optimal estimator for $\theta$.\\
Therefore, one just has to maximize the likelihood and can hope to attain a good value for $\theta$ as long as the nature of the noise $\epsilon$ is sensible for the task at hand. If one aims to maximize the likelihood, one can equivalently minimize its negative logarithm
\begin{equation}
L(x,y)=-\log(P(y|x,f)) 
\end{equation}
yielding a useful formulation for the loss function. Returning to our example of the linear model, this yields the following loss function (up to additive constants that do not matter for optimization):
\begin{equation}
L(x,y) \propto (\theta^Tx-y)^2
\end{equation}
which is the standard mean squared error loss that is frequently used in many machine learning tasks \cite{dl}.\\
\mybox{Loss Functions}{Parameterized estimators solve the memory bottleneck: As we only consider estimators of the form $f_{\theta}$ for $\theta \in \Theta=\real^k$, we can only hold $\theta$ in memory and therefore have constant memory usage. Loss functions describe how well the our estimator fits the data. The ideal estimator is the one estimator that minimizes the expected loss, also called risk.
As the risk is not computable, as we only have access to the training set $D$, the estimator is chosen to minimize $\hat{R}$, which is the mean loss over the dataset. Loss functions are attained by adding stochastic noise to the estimator and using the $-log(P(y|x,f))$ as the loss function. Thus, if $f^*$ is noisy, the estimator is a maximum likelihood estimator which converges to $f^*$ given enough data, the correct noise and that a global optimum is reached in optimization.}
\subsection{Gradient Based Optimization}
After having defined a loss function and the corresponding empiric risk $\hat{R}$, one still has to find the one estimator $f_{\theta}$, or equivalently its corresponding parameter vector $\theta$, that actually minimizes the empiric risk.\\
Finding a global optimum is a very complex problem for most nonlinear estimators. For example, even finding an optimal set of parameters for a very simple neural network is NP-hard  \cite{manurangsi2018computational}.\\
Thus, we restrict ourselves to efficient algorithms that find local optima. A useful algorithm to do this is gradient descent. One can use the gradient descent algorithm or one of its many variants such as ADAM \cite{kingma2014adam} that are heavily used in practice.\\
Gradient descent works as an iterative algorithm, starting with a random guess\footnote{Of course, in practice one does not initialize randomly. Sophisticated initialization strategies exist (see: \cite{glorot2010understanding}), but we will not cover them as they are not of huge interest to and exceed the scope of this work. } for parameters $\theta_0$ that is iteratively defined through multiple timesteps. In gradient descent and all of its variants, the next guess for parameters at timestep $t+1$ is given by:
\begin{equation}
\label{gdupdate}
\theta_{t+1} = \theta_{t} - \alpha \nabla_{\theta} \hat{R}(f_{\theta_{t}},D)
\end{equation}
for some small learning rate $\alpha$. The variants of gradient descent only differ in the choice of $\alpha$.\\
The idea behind this is very simple. The gradient $\nabla_{\theta} \hat{R}(f_{\theta_{t}},D)$ is known to always point into the direction of steepest local ascent. That means, that for infinitesimal steps the gradient points into the locally best direction to decrease the empiric error
\begin{equation}
e(\theta)= \hat{R}(f_{\theta_{t}},D)
\end{equation}
Using a taylor expansion of $e(\theta)$ yields:
\begin{equation}
e(\theta + \delta \theta)\approx e(\theta)+ \nabla_{\theta} e \delta \theta
\end{equation}  
and substituting the update described in Equation \ref{gdupdate} with $\delta \theta = -\alpha\nabla_{\theta_t}$ yields
\begin{align*}
e(\theta_{t+1})&=e(\theta_t + \nabla \theta_t) \approx e(\theta_t)- \nabla_{\theta_t} e \nabla_{\theta_t} e 
\\&= e(\theta_t)- \dist{\nabla_{\theta_t} e}^2 < e(\theta_t)
\end{align*}
Therefore, one can  see that, if the learning rates $\alpha$ are small enough for the taylor approximation to be accurate, the error is guaranteed to decrease in each iteration of gradient descent.\\
The idea behind this is relatively easy: One can visualize the function $e(\theta)$ as a plotted graph. At each step, $\theta_i$ is moved into the direction where the $e$ most steeply descents. Akin to a ball rolling down a hill, the value of $\theta_i$ moves down the graph of $e$. As the value of $e(\theta_i)$ improves at each step, the procedure is guaranteed to eventually find a local minimum or keep on improving until infinity. The last option does not really occur in practice as loss functions of the form $\-log(P(D| f_{\theta}))$ have a lower bound of $0$. As one rarely desires a loss of exactly zero, one can simply stop the process early if it does not converge and keep improving.\\
While gradient descent is guaranteed to improve the original error of a function and is also guaranteed to converge in all practical applications if the learning rate is small enough, there are some problems with it that one has to keep in mind. First off, gradient descent does not generally find a global optimum but rather only some local optimum of the function. This is of course due to the local nature of gradient descent: gradient descent only ever greedily moves toward the direction that is optimal locally and thus can never guarantee to reach a globally optimal solution. Moreover, out of all the possible local optima of the function $e$, gradient descent tends to choose local optima that are close to its starting point. The reason for this is very simple: gradient descent only ever takes local steps and thus is much more likely to reach a nearby local optimum. As in very powerful estimators there exist many local optima (as powerful estimators can fit the entire training data set in many different ways), this is a very important property of gradient descent as it describes which of the possible solutions it tends to choose. 
\mybox{Gradient Based Optimization}{In this work we consider only neural networks that are trained by using gradient based optimization methods to minimize a loss function. Gradient  based optimization methods are approximative function minimizers that find the parameters $\theta$ that minimize some function depending on $\theta$. Gradient based methods are very general; they only require $f_{\theta}$ to be differentiable w.r.t. $\theta$.  Further, they simply require $O(t*d)$ steps, where $d$ is the time it takes to compute the differential $\ddx{f_{\theta}}{d\theta}$ and $t$ is the number of training steps.  Gradient based optimization always finds a local minimum, given that the learning rate $\alpha$ is sufficiently small. However, in general gradient based optimization does not find a global optimum.}
\section{Regularization and Inductive Biases}
\begin{figure}[htbp]
\centerline{\includegraphics[scale=.8]{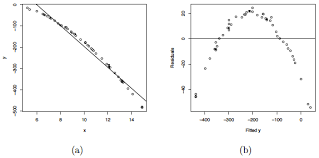}}
\caption{A linear model estimating a linear function (left) and a quadratic function (right). The first estimation task is solved well, the second is not. That is of course because the inductive biases present in the model assume a linear behavior of $f^*$ that is not actually present.}
\label{inductive}
\end{figure}
In some ways, training an estimator is a cursed task. If one observes some set of samples $D=\{(x_i,y_i)\}$, any estimator that classifies all samples correctly, i.e. $f(x_i)=y_i$ for all $i$, could in theory be the true function $f^*$. The set of all functions that classify the data correctly
\begin{equation}
	 \{f \; | \; f : \Omega \rightarrow C, \; \forall i\; f(x_i)= y_i\; \}
\end{equation}
is usually uncountably infinite. If one had no additional information, one would have to choose some arbitrary function from $S(D)$. The probability of correctly guessing $f^*$ from an uncountably infinite set is exactly 0 and for almost all tasks so is the chance of guessing a function $f$ that is sufficiently close to $f^*$.\\
 Yet, every machine learning algorithm does in fact choose one of the functions contained in $S(D)$ and most of them do so in a way that is at least somewhat useful in practice.\\ 
The \textit{inductive biases} of a  machine learning algorithm are defined as the rules and assumptions that are encoded in the algorithm's choice of estimator. For example, when one uses a linear model, there is an implicit assumption that the function $f^*$ is linear and thus a linear function is chosen from $S(D)$. As there is, given enough sample points, only one such linear function that correctly classifies all samples, this function is uniquely determined.\\
Of course, the inductive biases of a model heavily rely on human expert knowledge, even though they are not usually explicitly discussed. As the no-free lunch theorems famously state, there is no way for any machine learning algorithm to be better than any other unless it makes use of human knowledge of the domain $\Omega$ and the approximated function $f^*$ \cite{alpaydin2019maschinelles}.\\ Thus, a machine learning algorithm is \textit{good}, if and only if its inductive biases are useful in the domain where it is used. Consider Figure \ref{inductive}. Observe that in a linear setting a linear model works extremely well while it does not work at all in the approximation of a quadratic function as the inductive biases do not match the problem at hand.\\
Note that there is a certain duality of inductive biases of a machine learning algorithm and the assumptions that one makes for the function $f^*$ in that inductive biases should be based on said assumptions and frequently mirror them, i.e. if we assume linear behavior of $f^*$, we generally want $f$ to also behave linearly.\\
For most machine learning algorithm, there exist a certain set of inductive biases that are present in the algorithm itself. In a linear estimator, this assumption is that a linear function is a good estimator of $f^*$. In a fourier-series estimator, the assumption is that a linear combination of sin- and cosine functions is a good estimator. However, on top those inductive biases there are some that a user can generally fine-tune.\\
Those actively user induced inductive biases are what is generally referred to as regularization and are a subset of the full set of inductive biases. As defined by \cite{dl}, regularization refers to any modification we make to a learning algorithm that is intended to reduce its risk on the entire, unknown dataset but not its training error. This can be done in a multitude of different ways but for this section we will inspect the addition of terms to the loss function.\\
The core idea behind this is to add additional loss to the predictions of the estimator if the estimator does not obey the structure that its designer wants to enforce. Thus, the training algorithm needs to find a tradeoff between optimizing its training error and adhering to its intended structure, or, if possible, choosing the function that optimizes the training loss while most closely adhering to the regularization target.\\ 
To illustrate this, we will briefly review the L2-regularization scheme. L2-regularization replaces the usual loss $L$ with a modified loss $\hat{L}$:
\begin{equation}
\label{l2}
\hat{L}(x,t)=L(x,t)+ \lambda\dist{\theta}^2 
\end{equation}
where the euclidean norm (or L2 norm) of the vector of parameters of $f$ is added to the loss function $L$.\\
This formulation yields higher loss values for estimators with very uneven weight distributions that are far away from zero while only slightly increasing loss for estimators with parameter values that only slightly deviate from zero. Recall the example of linear estimators
\begin{equation}
f_{\theta}(x) = \theta^T x
\end{equation}
 The intuition behind enforcing a low L2-norm of the parameters is relatively simple: parameters are pushed towards zero and towards a relatively even distribution. Thus, function values of $f_{\Theta}$ do not change as much if $x$ is changed and a singular variable $x_i$ holds less influence over the estimation result.\\
Depending on the task, this can be a very good choice of regularization. For example, when doing image recognition a small change in one pixel is almost always close to meaningless. Large areas of the image have to change as a whole to induce meaningful difference. However, when one considers a chess board, the absence of one piece can drastically change the games structure and one would be misguided to use L2 regularization in this context. In general, L2 regularization leads to "less complex" estimators\footnote{Note that L2 regularization regularizes the parameters and its effect is thus tied to the class of estimators where this is used. However, as we will see later on, the effect L2 regularization has on linear models extends to neural networks as well, albeit in a different way.} and is thus intuitively unfit for more complex tasks. This vividly showcases the very domain specific nature of inductive biases and thus regularization.\\
There exist a multitude of other regularization methods such as dropout or early stopping \cite{prechelt1998early, baldi2013understanding} but all of them pursue the same target: Guide the classifier into a direction that is deemed useful not by the data but by the designer of the estimator.\\
For regularization methods that work by adding another term to the loss, there exists a very intuitive interpretation based on bayesian statistics \cite{polson2010shrink} that can easily be adapted to more general settings. In a bayesian setting, one would define a prior probability measure over the possible concrete estimators $P(f)$. Then, one would generally desire to use the estimator with the highest posterior probability:
\begin{equation}
P(f|D)\propto P(f)*P(D|f)
\end{equation}
This identity follows by the bayesian theorem. Of course, maximizing this quantity is equivalent to minimizing its negative logarithm: 
\begin{equation}
-\log P(f|D) = -\log P(f)  -\log P(D|f) + c
\end{equation}
where $c$ is some norming constant that is of little relevance to the optimization itself. Returning to the derivation of loss functions in Section \ref{losses}, observe that $-\log P(D|f)$ is just the loss function. Thus:
\begin{equation}
-\log P(f|D) = -\log P(f) - L(f|D) + c
\end{equation}
Comparing this with Equation \ref{l2}, it can be seen that the negative log of the prior $-\log P(f)$ plays the exact same role in this example as the added regularization loss. Hence, regularization can be interpreted as performing a maximum a-posteriori estimation of the function $f$ \footnote{Note that there exists some technical difficulties in a statistical sense that are omitted here. In fact, in a machine learning setting, one  maximizes over the probability of the parameters $\theta$, not the probability of the corresponding function $f_{\theta}$. While those are closely linked, it is possible that there exist $\theta_1$, $\theta_2$ such that $f_{\theta_1}=f_{\theta_2}$. In this case, it is in theory possible that the global maximum for $\theta$ does not correspond with the global minimum for $f_{\theta}$} with a prior given by $P(f)=e^{-\rho(f)}$ where $\rho$ is the loss term. Seen this way, a regularization term in the loss function encodes a prior belief (i.e. a belief that is held before seeing the actual data) about the candidate functions for $f^*$ and thus is a relatively soft constraint. It does not dictate that estimators with large parameter values are impossible, but makes them less likely to be chosen.\\
With this in mind, recall the formulation of the L2-loss. Here, the regularization term is given by:
\begin{equation}
\rho(f_{\theta})= \dist{\theta}^2
\end{equation} 
Thus, the corresponding prior belief is given by:
\begin{equation}
P(f)=e^{\rho(f_{\theta})}= e^{-\dist{\theta}^2}
\end{equation}
which is, up to normalizing constants, equivalent to a normal distribution over the estimator's parameters.\\
There are however even harder inductive biases than those that are induced by additional terms in the loss functions. For example, one might add strict requirements such that $P(f)=0$ for some specific functions $f$. One could of course implement this by simply setting the loss term to infinity for those functions, always preventing them from being chosen as the final result. However, this necessarily leads to numerical problems and makes it impossible to calculate a meaningful gradient of the loss function.\\
Thus, more strict requirements are usually built into the architecture of the estimator. For example, when using fourier transform one might have the assumption that the true function $f^*$ only contains broad, low frequency patterns and that all high frequencies occuring in the data are noise. Thus, one would directly omit all of the higher frequencies directly instead of posing a prior on higher frequencies being less likely to exist and including this prior in a mathematical optimization.\\
It is important to keep in mind that inductive biases and by extension regularization are not something that one can actively choose to use. The simple act of choosing an estimator from a class of possible estimators that describe the given data implies that one has some inductive bias that determines the chosen function. Thus, it is not a choice to include an inductive bias but rather a choice of what inductive bias should be chosen and it is the most critical part of any machine learning algorithm to include the correct inductive bias for the task at hand. \\
\mybox{Inductive Biases and Regularization}{Due to the no-free lunch theorem, all machine learnings have the same average generalization error when computed over arbitrary problems. Therefore, the question ``is a function approximator good?'' boils down to human decision making and cannot be answered in general. ``Inductive Biases'' is an informal term that denotes the way that an estimator uses the training data set to generalize. To answer, whether an estimator is useful, can only be done by evaluating its inductive biases by a human.    }
\section{Deep Learning}
\label{pl:deeplearning}
This section will briefly review the most important deep learning concepts for this work. It will start off by defining the most important terms and notation referring to deep learning before describing both the building blocks and the basics of deep learning models that will be used throughout the remainder of this work.
\subsection{General Terms and Definitions}
Before introducing the relevant deep learning concepts for this work, it is very important to do some terminological ground work. As this work is fundamentally concerned with different neural network architectures, it requires a specification of what actually is a neural network architecture. The term "neural network" is very loosely defined, referring to a multitude of very different models that appear to have very little to no crossover; compare for example the very different structures of transformer models, fully connected neural networks and boltzmann machines that are all considered deep learning models \cite{vaswani2017attention, dl}.\\
As to not to impose structural limitations on what a neural network can look like,we choose a very radical definition here and define deep learning as "function approximation using gradient based optimization" and therefore neural networks as "functions that can be trained using gradient based optimizers". Indeed, all of the aforementioned models use gradient based optimizers for training. To enable gradient based optimization, it is necessary that the neural network is both parameterized with some set of parameters $\theta$ and also differentiable with respect to said parameters. Therefore, we define a function to be \textit{trainable}, if it is parameterized and differentiable with respect to its parameters.\\
As we defined neural networks loosely as "function approximators trained using gradient based optimization", we can more concretely define neural networks as trainable, continuous functions $f : \Omega \rightarrow C$ where $\Omega$ is the space of our possible data and $C$ is the space of possible classes. \\
Note that there do exist approaches that do not use gradient descent to train neural networks such as genetic approaches \cite{gupta1999comparing} and would thus not fall under our characterization of deep learning. Still, we believe this definition to be sensible as almost all of the approaches relevant to this work use gradient based optimization and most, if not all, of the deep learning approaches presented in this work were built to work with gradient based optimization and thus fit the moniker of a trainable function.\\
We denote concrete sets of parameters as $\theta$ and the space of all possible parameter configurations for a neural networks as $\Theta$.\\
We will denote the space of all functions that some neural network $f$ can represent with some set of parameters $\theta$ as $\rep(f)=\{f_{\theta} | \theta \in \Theta \}$.
Further, we define a neural network architecture as a family of neural networks $F$ and we call a neural network architecture a universal approximator for some function space $S$, if one can choose $f \in F$ such that the representable functions $rep(f)$ lie arbitrarily dense in $S$\footnote{For example, in fully connected neural networks it is well known that the representable function of a neural network lie arbitrarily dense in the space $\real^n \rightarrow \real^m$ if the number of neurons and layers tends to infinity\cite{hornik1989multilayer}}.\\
Note that as deep learning lies in the crossroads of computer science, statistics and maths neural networks can be understood in many different ways, as graph based models, functions etc. and thus can be defined differently from the definition given here. For this work, we are mostly concerned with neural networks as function approximators as most ideas of geometric deep learning are motivated by this point of view \cite{geo}. Thus, we will generally abstract from the concrete graph based structure wherever possible and instead focus on the space of functions that a neural network can approximate.\\
In the following, we will review some neural network architectures that are relevant to this work and describe their use cases and domains of interest.
This section is heavily based on \cite{dl} but adapted to better fit the ideas presented in \cite{geo}.
\subsection{Fully Connected Neural Networks}
\begin{figure}[htbp]
\centerline{\includegraphics[scale=.5]{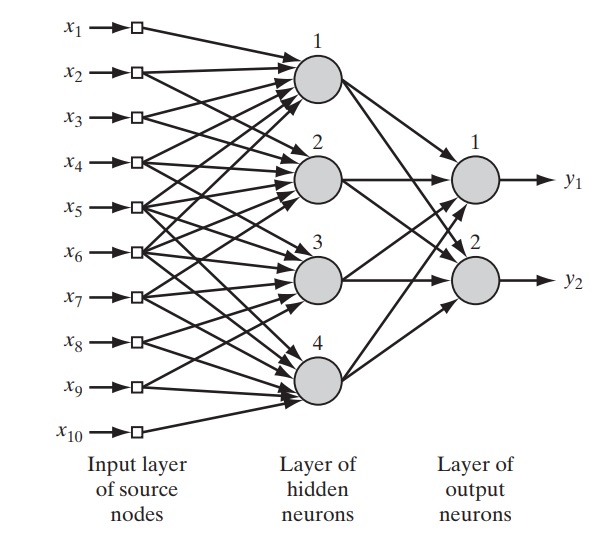}}
\caption{An example of a fully connected neural network. Note the layered architecture and the strictly forward facing direction of the edges. Also observe that all neurons are connected to all neurons in the succeeding layer. Illustration from \cite{haykin2010neural}}.
\label{activation_func}
\end{figure}
Fully connected neural networks ($FC$) are one of the very first deep learning models and have gone by many names such as feedforward neural networks or multilayer perceptrons \cite{dl}. Fully connected networks can be visualized as a graph $G=(V,E)$ partitioned into multiple layers $V=\coprod_i l_i$\footnote{The symbol $\coprod$ refers here to union of disjoint sets.} of vertices (called "neurons") that are connected by edges $e_{i,j}$  with weights $w_{i,j}$ where nodes are connected by an edge if and only if they belong to adjacent layers\footnote{This characterizes the eponymous fully connected-ness of the network}.\\
As input, the fully connected neural network takes arbitrary vectors $u=(u_1, ...,u_n)^T \in \real^n$ and outputs again vectors $\real^m$ for some $n$, $m$. The computation of the fully connected neural network happens layer by layer, computing a so called "activation" $a(v)$ for each neuron $v \in V$.\\
In the first layer, activations are set to match the input:
\begin{equation}
a(v_{0,i})=u_i
\end{equation}  
Afterwards, activations are computed one layer at a time, each building on the layer that came before it. At layer $k+1$, activations are given by:
\begin{equation}
a(v_{k+1,j})=\phi(\sum_{i=0}^{dim(k)} a(v_i)w_{i,j} +b_j)
\end{equation}
\begin{figure}[htbp]
\centerline{\includegraphics[scale=.25]{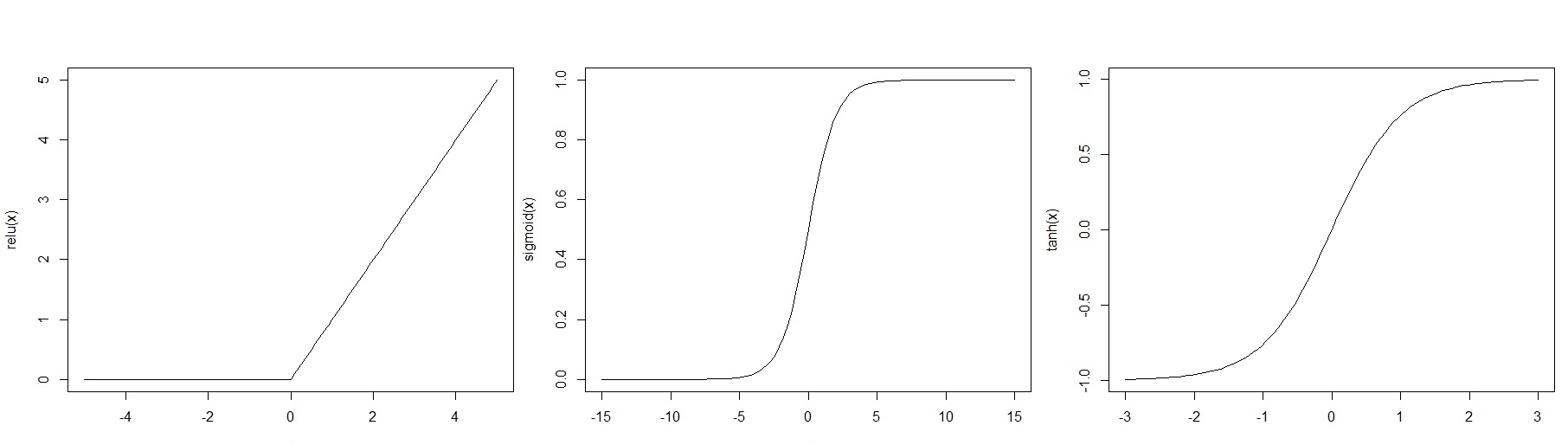}}
\caption{Some of the most popular activation function choices for $\phi$. Note that the activation function can in principle be chosen to be any function as long as it is differentiable and, preferably, monotone. The rectified linear unit (ReLU) is the de facto standard choice for activation functions at time of writing \cite{dl} and will thus be use for every example in this work.}
\label{activation_func}
\end{figure}
where $\phi$ is some nonlinear, differentiable, predetermined function called the activation function. If $\phi$ was omitted, the neural network would collapse to a simple linear model. In this way, the input is propagated layer-by-layer through the network, eventually reaching the last layer. The activations in the last layer are interpreted as the output; one can thus view the neural network as a function $f$ such that $f(u)= \{a(v_{L,1},..., a(v_{L,n}) \}.$\\
For the purposes of this work, it is perhaps more intuitive to view these computations in a vector-valued sense and instead of singular neurons inspect the effects that each layer has on its specific input. Of course, given input $u$, the vector $l_u=(a_{0,1},...,a_{0,n})$ of stacked activations $a(v)$ in the first layer is given by:
\begin{equation}
l_0(i)=u
\end{equation}
Similarly, the later layers can be represented as
\begin{equation}
l_{i+1}=\phi(W_{i+1} l_i+B_{i+1})
\end{equation}
Where $W_{i+1}$ and $B_{i+1}$ represent the matrix of stacked weights $w_{i,j}$ and the vector of stacked biases $b_j$ respectively and $\phi$ is applied pointwise to each element of its input argument. There are two different things happening in this equation: First, there $l_i$ is passed into an affine transformation with $W_{i+1}$ and $B_{i+1}$. Afterwards, a nonlinear function is applied. The full function represented by the neural network is thus given by:
\begin{equation}
f :=   \phi \circ A_L .... \phi \circ A_2 \phi \circ A_1
\end{equation}
where $A_L$ are the affine transformations described before and $\phi$ is the nonlinear activation function. Thus, a fully connected neural network is in its essence given by an alternating concatentation of the nonlinear activation function $\phi$ and affine transformations $A_i$. As $\phi$ is previously fixed and $A_i$ are the only parameterized parts of $f$, they are the trained functions; $\phi$ is only used to induce a sufficiently nonlinear structure to the function.\\
This structure makes fully connected neural networks extremely expressive. As shown in \cite{hornik1989multilayer}, fully connected neural networks can arbitrarily closely approximate any given continuous function on a compact subset of $R^n$ if enough layers and/or neurons are provided\footnote{It can be shown that both an arbitrary number of layers with fixed width or an arbitrary number of neurons in one singular layer are sufficient for universal approximation\cite{pinkus1999approximation, park2020minimum}}.\\
Thus, the representable space of functions for fully connected neural networks is given as: 
\begin{equation}
rep(FC) = \{g \; \mid \; g \in \real^n \rightarrow \real^m, \text{g continuous}\} 
\end{equation}
\section*{Deep Learning Summary}
 In this section, we introduced deep learning. Due to the vagueness of the term deep learning, it is almost impossible to find a coherent definition. Therefore, we just let every function that can be trained by gradient descent be called a ``neural network''. The question of ``What is a neural network that, if trained with gradient based optimization yields good results in our domain?'' thus becomes: ``What is a trainable function that, if trained with gradient based optimization yields good results in our domain?".\\
Then, we introduced fully connected neural networks as a baseline for neural networks. Fully connected neural networks are the most basic form of neural networks. Most interestingly, they are able to approximate every real valued, continuous function to arbitrary precisison.\\ 
On one hand, this guarantees that fully connected neural networks can, in theory, approximate any continuous function on a compact subset of $\real^k$ and thus the function $f^*$, if it is continuous. On the flipside, we might know of some function $g$ that can not feasibly be the correct guess for $f^*$ but our neural network might still approximate $g$. As noted before, we can only evaluate whether this is good based on the problem where we seek to use neural networks. Further, this also means that there are some unanswered questions as to why neural networks are good: If they can approximate any continuous function in theory, how can we be sure that they approximate useful functions?
\mybox{Deep Learning and Fully Connected Neural Networks}{In the context of this work, deep learning is function approximation using neural networks. Neural networks are functions that can be trained by gradient descent and thus are parameterized and differentiable with respect to their parameters. Fully connected neural networks are one such function $\real^k \rightarrow \real^l$. Fully connected neural networks are comprised of alternating affine transformations with trainable parameters and fixed nonlinear activation functions, most frequently the ReLU function. Fully connected neural networks $f$ can approximate every continous function $g$ on a compact subset of $\real^k$ such that $\sup_{ x \in \real^k} \dist{f(x)-g(x)} \leq \epsilon$ for each $\epsilon >0$ if given enough layers and neurons.}
\section{Inductive Biases of Fully Connected Neural Networks}
\begin{figure}
\centering
\includegraphics[scale=0.7]{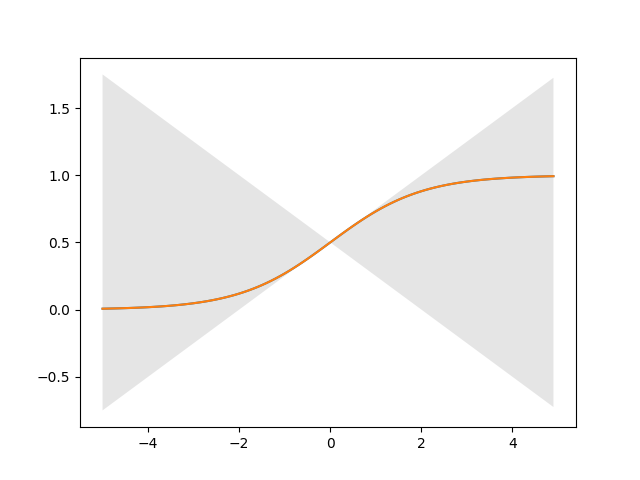}
\caption{The sigmoid activation function $\sigma$ which is lipschitz with $L(\sigma)=\frac{1}{4	}$. The grey area indicates the bounds for the function that are induced by the lipschitz constant at $x=0$. The lipschitz constant of $\frac{1}{4}$ defines that the function can only ever change at most with rate $\frac{1}{4}$. Check how the function never leaves the boundaries defined this way. }
\end{figure}
\begin{figure}[!tbp]
  \centering
  \subfloat[2 layers]{\includegraphics[width=0.5\textwidth]{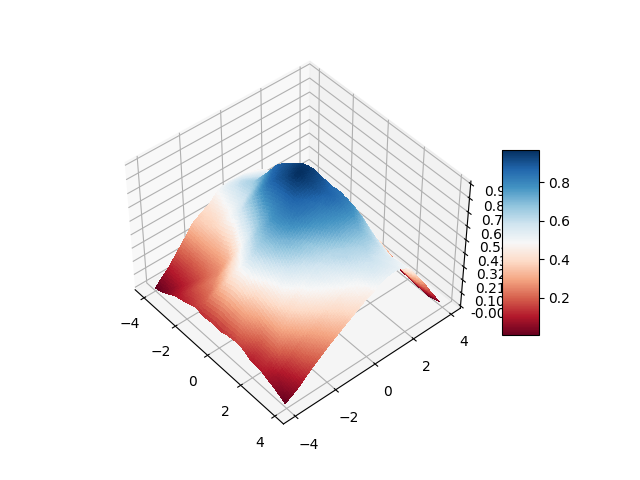}\label{fig:f1}}
  \hfill
  \subfloat[5 layers]{\includegraphics[width=0.5\textwidth]{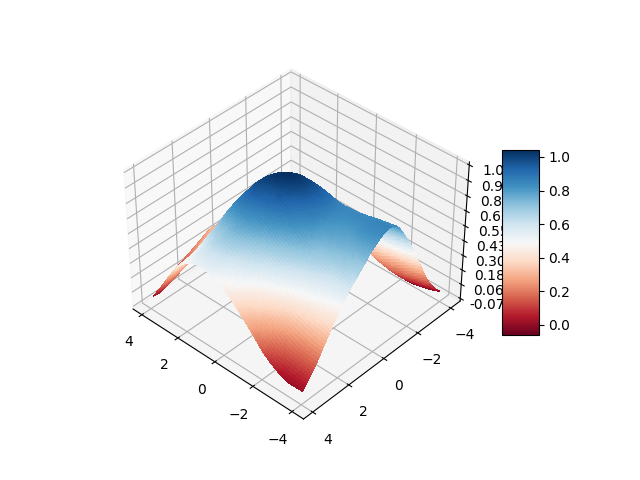}\label{fig:f2}}
    \hfill
  \subfloat[15 layers]{\includegraphics[width=0.5\textwidth]{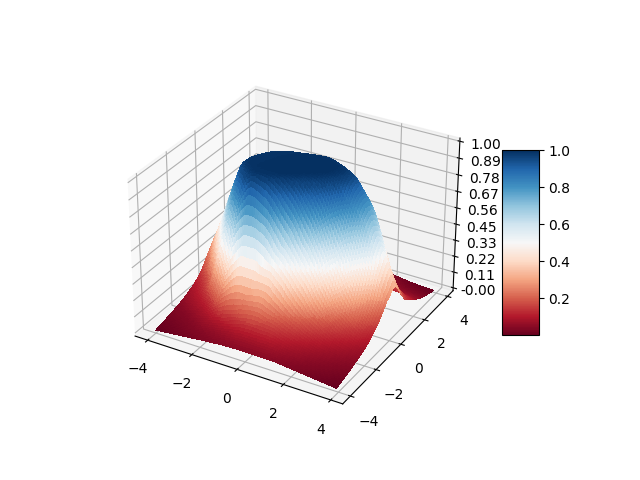}\label{fig:f2}}
    \hfill
  \subfloat[25 layers]{\includegraphics[width=0.5\textwidth]{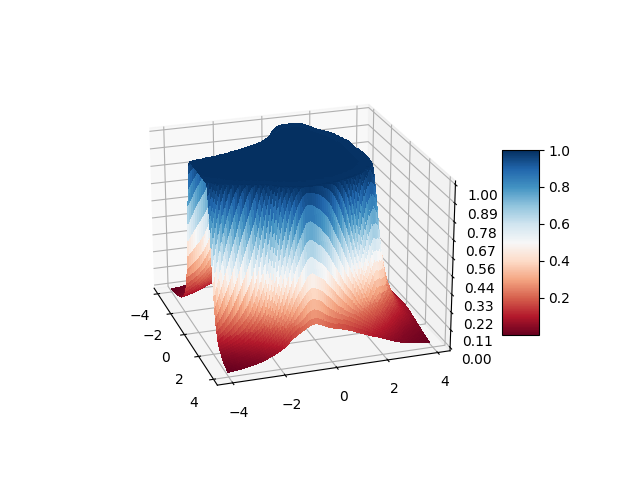}\label{fig:f2}}
    \hfill
  \subfloat[35 layers]{\includegraphics[width=0.5\textwidth]{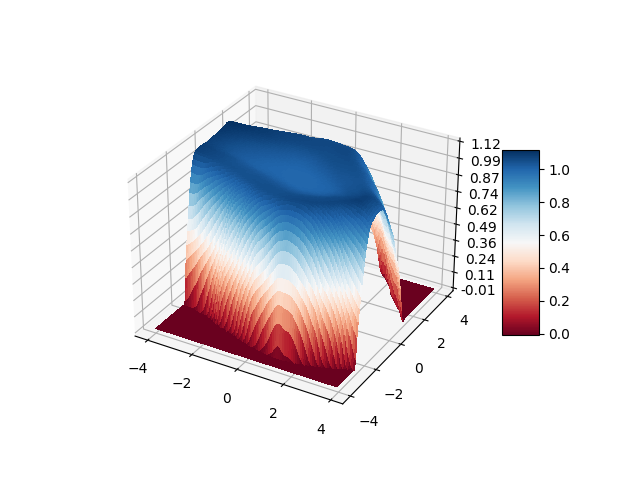}\label{fig:f2}}
\caption{Neural networks with a different number of layers learning the same task. Each layer is given the same data points. One data point at each corner (i.e. $(x,y)$ where $|x|=|y|=4$ } with value $0$ and one in the middle at $(0,0)$ with value $1$. Each network solves this, but interpolates in different ways. Most notable, the neural networks with higher numbers of layers approximate much more unsteady functions with higher lipschitz constants. These functions have much sharper 'cliffs', especially compared to networks with 2 or 5 layers. Note that there is no benefit to this behavior, all models achieve an error of $0$, the reason for this behavior lies only in the inductive biases given by the number of layers.
\end{figure}

Now that we have cleared up the setting that we want to work in, we can start answering the question of ''what is a good domain specific neural network''. As we described throughout this section, this question is more specifically posed as: 
\begin{center}
What is a trainable function that can be trained well with gradient based optimization and whose behavior matches our domain knowledge?
\end{center}
To answer this, we will proceed as follows: We will first examine the behavior of fully connected neural networks as a baseline. Fully connected neural networks have performed decently well in classification tasks even without being explicitly based on domain knowledge. Thus, it stands to reason that they possess some properties that make them useful classifiers that we would like to maintain for our domain specific neural network architectures. We are especially interested in their inductive biases, i.e., how does the neural network generalize to unseen datapoints. Then, we will examine whether fully connected neural networks have shortcomings in specific domains and how they can be fixed using domain knowledge and the ideas of geometric deep learning.\\
First off, the most important and most general inductive bias introduced by neural networks is the fact that they are able to approximate arbitrary continuous functions. That is, each function that a neural network can learn is necessarily continuous. Thus, there is some semblance of euclidean similarity that is assumed over the inputs: By continuity, if some inputs $x$ and $y$ are close in a euclidean sense, we can assume that $f(x)\approx f(y)$. This is a core concept of neural networks. They generalize (at least in part) by assuming that nearby points have similar values  for $f^*$.\\
While continuity in itself tells us that neural networks treat nearby points similarly, it does not tell us \textit{how} near these points have to be to be treated as adequately similar. There are however some stronger results regarding the lipschitz continuity of neural networks based on the works of \cite{scaman2018lipschitz}.\\
A function $f$ is called lipschitz continuous if there exists some constant $L$ such that for all $x,y \in \Omega$
\begin{equation}
\dist{f(x)-f(y)} \leq L \dist{x-y}
\end{equation}
The smallest constant $L$ for which this is the case being called the lipschitz constant of $f$, $L(f)$. It is clear from definition that lipschitz continuity is a stronger requirement on a function $f$ than continuity: A function with lipschitz constant $L(f)$ can only change its value by at most $L(f)$ for each unit length step taken in the domain $\Omega$. As a consequence, lipschitz continuity can be reframed as an upper bound on the gradient and thus the partial derivatives:
\begin{equation}
\sup_{x \in \Omega} \dist{\nabla_x f} = L(f)
\end{equation}
and as a consequence
\begin{equation}
\ddx{f(x)}{x_i} \leq L(f)
\end{equation}
A lipschitz constant of a neural network is very interesting with regards to the inductive biases present in a neural network. The lower the lipschitz constant, the more nearby points are similar. In consequence, if the neural network is trained to perfectly estimate each training point, i.e. $f(x_i)=y_i$ for all $(x_i,y_i) \in D$ it can easily be seen that for each point $p \in \Omega$ for which $\dist{x_i-p} = d$ it holds that
\begin{equation}
f(x_i)-dL(f) \leq f(p)\leq f(x_i)+ dL(f)
\end{equation}
Therefore, in a lipschitz continuous estimator that perfectly fits the training data \footnote{This is not guaranteed to happen, but modern, powerful neural networks can fit arbitrary datasets in most applications to high precision as shown by \cite{zhang2021understanding}.}, the training data points $y_i$ restrict the possible values for the points in their euclidean neighborhood.\\
The missing piece to relate this to actual neural network is the examination of lipschitz constants \footnote{Note again that neural networks are always lipschitz continuous and thus there exists some lipschitz constant for each neural network.} of fully connected neural networks. We will use results from \cite{scaman2018lipschitz} to show how the lipschitz constants of fully connected neural networks behave. Recall that
\begin{equation}
\sup_{x \in \Omega} \dist{\nabla_x f} = L(f)
\end{equation}
Therefore, to find the lipschitz constant $L(f)$, it suffices to find the one $x \in \Omega$ that maximizes the length of the gradient $\nabla_x f$. For a more intuitive understanding, suppose that we are tasked with a neural network $f$ with one output neuron, i.e. the range of $f$ is $\real$ such that the gradient $\nabla_x f$ is indeed a vector.\\
As done in \cite{scaman2018lipschitz}, we can inspect the lipschitz constants for each neuron in the network. For the input layer, each neurons activation is equivalent to the given input:
\begin{equation}
v_{0,i}=x_i
\end{equation}
Therefore, the gradient $\nabla_x v_{0,i}$ can be easily seen to be
\begin{equation}
\nabla_x a_{0,i} = (0,0,...,1,...,0)^T
\end{equation}
where the i-th entry of this vector is equal to 1, the lipschitz constant of the entry nodes thus being 1.\\
For a neuron in the k+1-th layer, the activation is given as:
\begin{equation}
a(v_{k+1,j})=\phi(\sum_{i=0}^{dim(k)} a(v_i)w_{i,j} +b_j)
\end{equation}
The  chain rule can be applied here to attain $\nabla_x v_{k+1,j}$:
\begin{align}
\nabla_x v_{k+1,j} &=
\nabla_x\phi(\sum_{i=0}^{dim(k)} a(v_i)w_{i,j} +b_j) \\&=
\phi'\left( \sum_{i=0}^{dim(k)} a(v_i)w_{i,j} +b_j\right) 
\sum_{i=0}^{dim(k)} w_{i,j}\nabla_x a(v_{k,i})
\end{align}
To find the lipschitz constant of the neural network, i.e.  $ \sup_x\dist{\nabla_x v_{k+1,j} }$, we have to find an upper bound for $\dist{\nabla_x v_{k+1,j}}$. Using the triangle inequality:
\begin{equation}
\dist{\nabla_x v_{k+1,j}} \leq \phi'\left( \sum_{i=0}^{dim(k)} a(v_i)w_{i,j} +b_j\right) 
\sum_{i=0}^{dim(k)} w_{i,j}\dist{\nabla_x a(v_{k,i})}
\end{equation}
If we can find an upper bound over all $x$ for the right hand side of this equation, we can derive an upper bound for the lipschitz constant $L(f)$.\\
One can easily see that $\nabla_x a(v_{k,i})$ is bounded by $L(v_{k,i})$, the lipschitz constants of the previous layers, by definition. Similarly, the quantity $\phi'\left( \sum_{i=0}^{dim(k)} a(v_i)w_{i,j} +b_j\right)$ is bounded by the lipschitz constant of $\phi$, for example this is $L(\phi)=1$ for the ReLU activation function.\\
Thus, we attain:
\begin{equation}
L(v_{k+1,j}) \leq L(\phi)\sum_{i=0}^{dim(k)} |w_{i,j}| L(v_{k,i})
\end{equation} 
Given a fully connected neural network where all weights are the same, this inequality can be reached.\\
Observe the recursive pattern in this equation: The upper bound for the lipschitz constants of a neuron in layer $k$ is given as a weighted sum of the lipschitz constant in the previous layer. As long as weights are greater than $1$ in sum for each neuron, the sum of lipschitz constants in layer $k+1$ is greater than in the previous layers. Thus, at each layer the average lipschitz constant increases, usually drastically as $dim(k)$ and $w_{i,j}$ can be large. To further simplify this, let us assume that $L(v_{k,i})$ is uniform for all nodes in the previous layer such that $L(v_{k,i})=\overline{L(v)}$ and also assume that $|w_{i,j}|$ are bounded by some value $\overline{w}$\footnote{This is reasonable for two reasons: First, backpropagation tends to yield local minima in the near vicinity of initialization, usually close to 0. Second, regularization often entails shrinking the $w_{i,j}$}. Then, we attain:
\begin{equation}
L(v_{k+1,j}) \leq L(\phi) \dim(k) \bar{w} \overline{L(v)}
\end{equation} 
This shows that the upper bound on on the lipschitz constant of each neuron can increase exponentially through the layers and linearly when more neurons are added to one singular layer\cite{pmlr-v9-glorot10a}. This matches similar results concerning the complexity of functions that can be approximated by fully connected neural networks such as \cite{bianchini2014complexity}. Note that the findings in this section can be extended to almost arbitrary layered neural network architectures as shown in \cite{scaman2018lipschitz}.\\
This yields a very salient way to interpret the inductive bias of fully connected neural network architectures with respect to the number of its layers and neurons. Depending on the numbers of layers and neurons\footnote{Of course, if the weights in a neural network are very large even shallow networks can achieve a high lipschitzness. This is usually not the case, however, for reasons pointed out in footnote 4.}, the resulting neural networks tend to have bigger or smaller lipschitz constants, varying more or less for very similar inputs.\\ 
Thus, larger networks can learn much more heavily varying functions and, in signal theoretic terms, capture higher frequencies present in the data. Conversely, smaller neural networks can only learn very steady, slowly varying functions with lower frequencies. If one has an abundance of data, larger networks are thus preferable as smaller networks might not be able to fit all of the trainings examples. Conversely, for little data, smaller networks are preferable as each datapoint affects the neural network outputs in a longer range, extrapolating to more possible values. By changing up the number of neurons/layers, one can thus also change the inductive bias present in the neural network, allowing for higher or lower lipschitz constants. \\
\begin{figure}[!tbp]
  \centering
  \subfloat[$\lambda=.001$]{\includegraphics[width=0.7\textwidth]{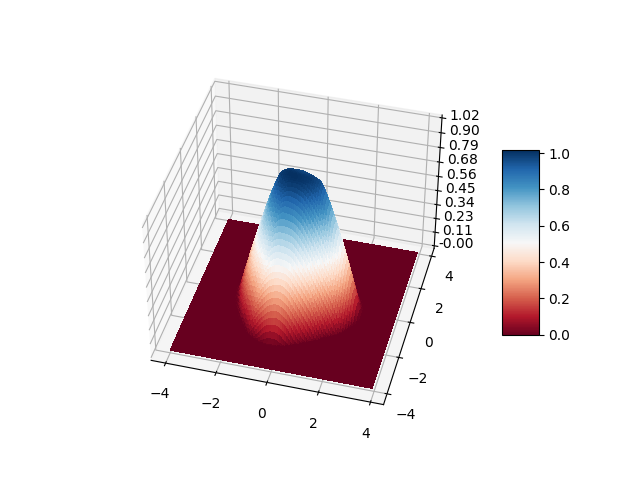}\label{fig:f1}}
  \hfill
  \subfloat[$\lambda=.002$]{\includegraphics[width=0.7\textwidth]{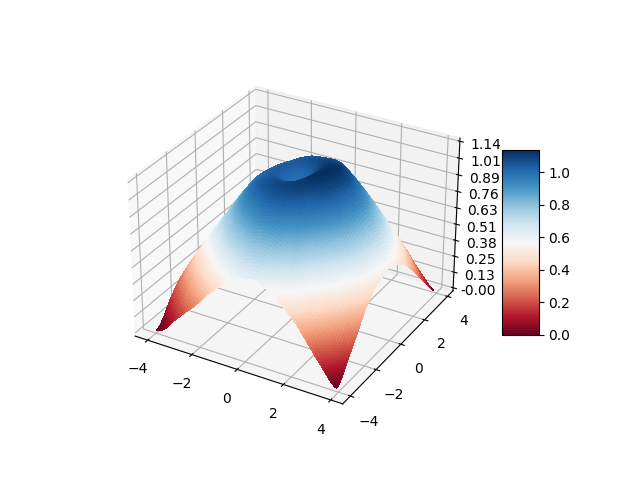}\label{fig:f2}}
  \caption{Two neural networks with 4 layers that were trained on the same task using different values for the weight of the L2-loss. A higher value of $\lambda$ leads to lower values for edge weights and thus lower lipschitz constants. The networks were trained to assign $0$ to all corners of the shown area and $1$ to the midpoint, i.e. $(0,0)$. Both fit the task well but do so differently. Network a (with a lower value of $\lambda$ yields a flat surface with a spike around $(0,0)$ which has a very high lipschitz constant. Network b chooses a flatter, more continuous version. Of course, the lipschitz constant of $b$ is much lower. }
\end{figure}
Lastly, this gives justification to the idea of layers themselves. As layers lead to a potentially exponential increase in the lipschitz constant of a neural network, they  are highly beneficial in the construction of powerful architectures that have to fit large datasets while only requiring a linear increase in neural network size.\\
By changing up the number of neurons and layers in a neural network, one can change the inductive biases present in the network, allowing for higher or lower lipschitz constants depending on the number of layers used. Note that this is not just a theoretical result but occurs in practice as demonstrated in the experiments conducted by \cite{scaman2018lipschitz} and also matched by experiments conducted for the sake of this work. Similarly, \cite{bianchini2014complexity} attain very similar results with regards to the power of additional layers, giving credence to the idea that additional layers allow for the approximation of exponentially more complex tasks\footnote{Interestingly, this holds for multiple different definitions of "complexity". As shown by \cite{bianchini2014complexity}, this holds if one uses topological complexity of $f^*$ and as shown by  \cite{scaman2018lipschitz} this holds using a lipschitz based definition of complexity.}.\\
For the remainder of this work, we consider those properties as the inductive biases that a neural network architecture should contain to be useful. The idea of nearby points being similar extends to many different tasks and is core to gradient descent and neural networks themselves. Moreover, the tunability of a neural networks inductive biases is of importance. Perhaps one of the biggest advantages of neural networks when compared to other machine learning algorithms is their extreme flexibility. One can easily change the inductive biases of a neural network without changing its fundamental architecture. Tuning knobs are for example the number of neurons per layer, the number of layers and the choice of regularizing loss function.  \\
Note also that this is not a complete list of the inductive biases present in fully connected neural networks. For example, initialization of parameters is a way of actively inducing biases and plays a huge role in how a neural network learns \cite{glorot2010understanding}. Also, neural networks can hardly be fully described by their lipschitzness.\\
However, for the remainder of this work we are concentrated on deriving new architectures and desirable properties of these architectures. The implicit biases presented in this section are some of the most fundamental implicit biases of neural networks and perhaps most important to their success. Thus, we desire each new architecture to solve some domain specific problem of fully connected networks while maintaining those following biases and properties on top of the always required trainability of neural networks:
\mybox{Inductive Biases of Neural Networks}{Fully connected neural networks are lipschitz continuous with some lipschitz constant $L$. Therefore, it holds that
	\begin{equation}
	f(x+v) \in [f(x)-L\dist{v}, f(x)+L\dist{v} ]
	\end{equation}
	Therefore, points that are nearby in a euclidean sense have similar function values under the neural network. A higher lipschitz constant means that more and more complex functions that can be approximated.	The lipschitz constant of a neural network depends on its structure and weights. It grows by a factor of $O((w_{\max}*b_{\max})^l$ where $w_{\max}$ is the largest weight in the neural network, $b_{\max}$ is the largest number of neurons in any layer and $l$ is the number of layers. $w_{max}$ is heavily influenced by regularization techniques and initialization. However, no guarantees about $w_{max}$ can be given.}
\begin{itemize}
\item "Continuity"
\begin{itemize}
\item Each function $f$ representable by a neural network is continuous
\item Moreover, $f$ is lipschitz continuous with some lipschitz constant $L(f)$
\item Therefore, nearby points by a euclidean metric are similar with respect to $f$, the degree to which they are similar is described by $f$'s lipschitz constant
\end{itemize}
\item "Tunability"
\begin{itemize}
\item The degree to which "nearby points" are similar, and thus the lipschitz constant $L(f)$, should be (at least indirectly) tunable
\item To scale well to complex tasks, there needs to exist a mechanism that exponentially (or at least super-linearly) scales to more complex tasks. This is most often achieved through some layered architecture.
\end{itemize}
\item "Completeness"
\begin{itemize}
\item All functions of interest should be able to be approximated by our neural network architecture.
\end{itemize}
\item "Trainability"
\begin{itemize}
\item As a baseline, a neural network needs to be trainable, i.e., parameterized and differentiable w.r.t. its parameters
\end{itemize}
\end{itemize}

\section{Problems with Feedforward Neural Networks}
\begin{figure}
\centering
\includegraphics[scale=0.4]{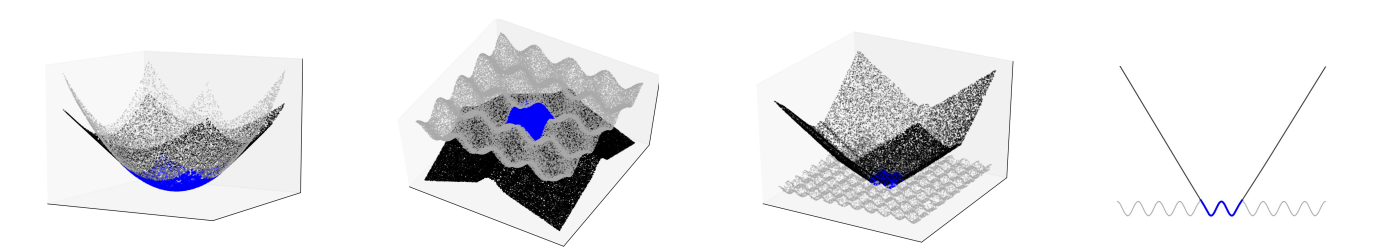}
\label{linex}
\caption{Some examples from \cite{xu2020neural} for the failure of extrapolation of nonlinear functions by fully connected neural networks. The neural network's predictions are presented in black, the actual function in gray and the training dataset in blue. Observe that the first task is quadratic in nature and thus the network's predictions start to undershoot outside of the training data. The second task is periodic which the neural network entirely fails to approximate, instead linearly continuing the trend of one period of the training data. Similar findings hold for the latter two tasks. }
\label{linearextrapolation}
\end{figure}
As outlined in the former section, feed forward neural networks possess some very convenient properties for function approximation, they can be tuned to approximate more or less complex functions with sublinear effort and they assume that nearby points are similar with respect to the function to approximate. In the absence of any additional information, this seems sensible. In most real world tasks, similar inputs yields similar results to at least some degree\footnote{In tasks where this is not the case, neural networks should also not be the method of choice or a different input representation should be chosen.}. However, there are some problems with fully connected neural networks which is why they are indeed not currently state of the art in most tasks of interest \cite{alom2019state} and domain specific architectures such as graph neural networks \cite{scarselli2008graph} are preferred.\\
The first problem with fully connected neural networks is their assumption of continuity. While we argued in the previous section that this is a useful property for many domains, there do exist other domains where either the domain or co-domain are not continuous. Neural networks are fundamentally functions $\real^k \rightarrow \real^j$ for some $k,j$. Frequently, neural networks are used for classification. There, the co-domain is not euclidean but rather a set of distinct labels. This does not fit the co-domain of fully connected neural networks and also poses problems for gradient descent based optimization, which requires gradients to be defined. As gradients cannot be defined on a set of labels, gradient descent is not usable. Still, neural networks are frequently employed in those domains, using various workarounds and these problems are not tackled. Therefore, we focus not on discrete domains but rather continuous domains with additional structure. For a more in-depth discussion of neural networks in discrete contexts, refer to the appendix.\\
For now, we consider fully connected neural networks in domains that can be represented as real-valued function approximation tasks. Although fully connected neural networks are usable and successful in these contexts, they have some problems in that fully connected neural networks do not extrapolate well for most tasks even if used in continuous domains. That means, if a neural network learns from a set of datapoints $D$, it does not learn well for points that are far away from all of the points in $D$. As shown in \cite{xu2020neural} in theory and in many other works in practice \cite{barnard1992extrapolation, haley1992extrapolation}, fully connected neural networks with ReLU activations, i.e. the by far most common choice of activation function, quickly become linear when moving away from the training data $D$ and thus have a very hard time extrapolating nonlinear functions $f^*$.\\
\begin{figure}
\label{extrap}

\centering
\includegraphics[scale=0.4]{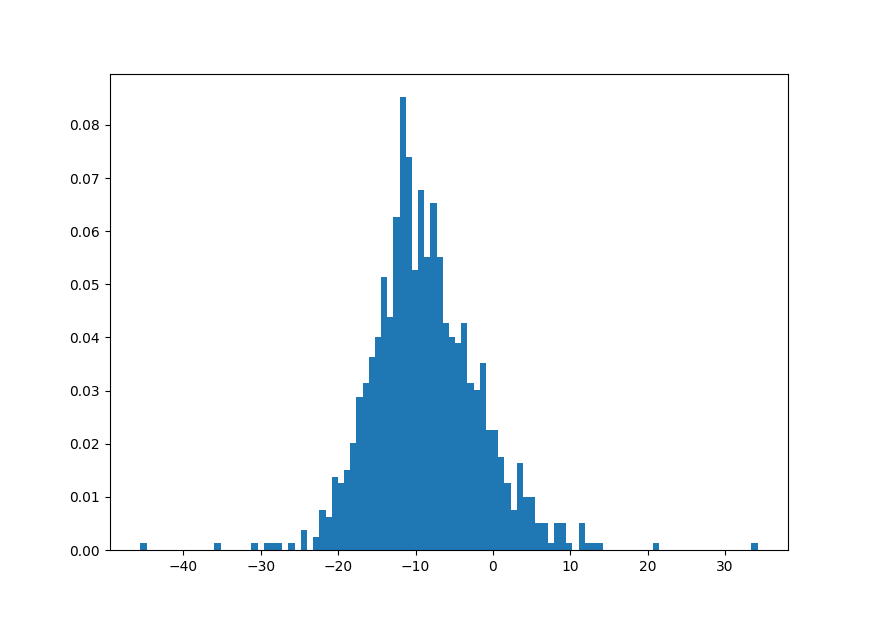}
\caption{A histogram showcasing the stability of neural networks linear extrapolation. The same neural network was trained 1000 times on a simple example problem where $f^*((0,0))=1$ and $f^*(	\pm 4, 	\pm 4)=0$ are given and eventually queried on the point $(50,50)$ which is far away from all training data points. Note that the distribution of $f(50,50)$ centers around 12.5, the value that would be expected by linear extrapolation. There exists some right leaning in the attained curve which is most likely attained as a result of initialization biases. }
\end{figure}
This gives a very clear picture of how neural networks tend to behave after being trained on some dataset $D$. In regions where datapoints are dense, neural networks can exploit their potentially high lipschitzness to approximate these points very closely, yielding highly complex functions. Outside of these datapoints they continue linearly and attain this linearity relatively quickly as shown by \cite{xu2020neural}.\\
To be exact, they show that any neural network with 2 layers, the ReLU activation function trained with gradient descent attains this behavior, given that no datapoints lie in the vicinity of the line induced by $x+hv$ where $x$ and $v$ are vectors.
\begin{equation}
|\frac{\dist{f(x)+hv}}{h} -\beta_v| < \epsilon
\end{equation}
where $\epsilon \in O(\frac{1}{h})$. Where $\beta$ is a linear coefficient independent of $h$. This means that, on any line that reaches outside the region containing the dataset, the neural network quickly approaches linear behavior. \\
This should not be understood as an inherent weakness of fully connected neural networks. If one lacks all kinds of domain knowledge, it is arguable that a linear continuation of the observed pattern is a very sensible choice. However, if one does possess domain knowledge, one should be able to improve on this linear continuation. This is perhaps one of the most important ideas for geometric deep learning: Extrapolation is not generally possible without domain knowledge.\\
For example, consider the last problem presented in Figure \ref{linearextrapolation}. The neural network extrapolates by linearly continuing the pattern it has observed instead of continuing in a periodic way which would be more apt for this problem. As a result, it horribly misestimates the actual $f^*$. If one knew beforehand of the periodic nature of $f^*$ and could force the neural network to instead periodically continue the pattern it has observed, one would attain a much better approximation of $f^*$.\\
There are two core challenges that result from this statement. First, one needs to identify properties of $f^*$ that one can exploit for function approximation without actual knowledge of $f^*$ by leveraging human domain expertise. Second, one needs to build a neural network architecture such that it extrapolates based on the properties of $f^*$, usually this entails having $f$ mimic the expected behavior of $f^*$.\\
Of course, doing so generally is an extremely hard task as domains in which neural networks are used are diverse and numerous. As a consequence, some functional properties that are assumed for $f^*$ can be extremely hard to enforce in $f$. Thus, geometric deep learning focuses on two domain specific properties that are both present in many different deep learning domains (and therefore generally useful) and also relatively easily enforceable. These properties are invariance of $f^*$ to some group of transformations (called symmetries) and a separability of $f^*$ into locally acting functions in different scale.\\
In the following sections, we will introduce these properties, justify that they are indeed found in many different domains before moving on to an example that showcases how they can be enforced in an estimator $f$.\\
\mybox{Extrapolation Behavior of Neural Networks}{Despite the fact that neural networks can approximate arbitrary functions, neural networks that are trained with gradient based optimization techniques extrapolate by linearly continuing the pattern they learned on their input data. The authors of \cite{xu2020neural} show that two layer neural networks trained with gradient descent and a squared error loss converge arbitrarily close to a linear functions outside of the dataset:
\begin{equation}
|\frac{\dist{f(x)+hv}}{h} -\beta_v| < \epsilon
\end{equation}
where $\epsilon \in O(\frac{1}{h})$. Further, for networks with more than two layers, this was supported by our experiments and other empiric results \cite{barnard1992extrapolation}. See Figure 2.8.\\
}

%% file: kapitel/gdl.tex
\chapter{An Introduction to Geometric Deep Learning}
Now that we have set a baseline of the general setting where we want to apply neural networks and gotten some insight into the behavior of fully connected neural networks that are explicitly \textit{not} built with domain knowledge in mind, we can start tackling the question of ``what should good domain specific neural networks look like''.\\
As we saw before, fully connected have some very nice properties that are useful almost independently of the domain we would want to use them in. If we want to design useful domain specific neural networks, it stands to reason that those should be retained by any good domain specific neural network as they appear useful almost independently of context. 
\begin{enumerate}
\item \textbf{Trainability.} Fully connected neural networks are trainable, which is necessary for usage in gradient based optimizations. This is necessary for usage in our framework.
\item \textbf{Continuity.} Fully connected neural networks assume that $f^*$ is continuous and, furthermore, that it is lipschitz continuous with some lipschitz constant that is not arbitrarily large. As a consequence, $f^*(x) \approx f^*(y)$ holds if $x$ and $y$ are relatively close. This is a useful property for many real-world tasks, as generally inputs that are close in a euclidean sense are also similar with respect to the task at hand. 
\item \textbf{Tunability.} The lipschitz constant, and therefore the degree to which nearby points can be assumed to be similar, can be controlled by adding more layers or initialization and regularization schemes.
\item \textbf{Completeness.} If given enough resources, a fully connected neural network is able to approximate every continuous function $\real^n \rightarrow \real^m$ to arbitrary precision,
\end{enumerate}
However, we also observed problems with fully connected neural networks that arise due to the lack of domain knowledge built into fully connected neural networks. There are two main problems that we observed: First, if a neural networks observes $f^*(x)$ for some $x$, it will assume that the nearby points $y$ of $x$ have similar values $f^*(y)$. While sensible, there are in most domains more points $y$ to which we could extend learning leveraging our domain knowledge. For example, if an image is mirrored, the euclidean distance between the original image and the mirrored version is very large. Still, they represent the same objects and thus learning should extend from one to the other. Second, neural networks extrapolate by a linear continuation of the pattern they observed. This makes sense without domain knowledge, but breaks down if we know more about the task at hand. For example, as a consequence, neural networks tend to generalize very badly on periodic tasks.\\
Thus, we have to identify properties of domains that we want to include in our neural network. At this point, we turn to geometric deep learning that identifies two properties of domains that many domains of interest appear to have in common and, which the authors argue, should be respected by domain specific neural networks \cite{geo}. 
\section{Symmetries and $\grp$-Invariance}
As noted earlier fully connected neural networks include the inductive bias that nearby points have similar values for $f^*$. For many tasks, this is a very intuitive and useful bias. For example, if one was tasked with recognizing objects in a picture, the classification of the object should change only little if one darkens each pixel by a small amount. The original picture and the darkened picture are very close to each other in a euclidean sense so the fully connected neural network correctly assumes that the object that they represent is similar as well.\\
However, while this is a widely useful bias, in many problems there exists stronger, domain specific structure that should be leveraged in function approximation.\\
To start, let us consider the problem domain of images of fixed size, where an image $x$ is formalized as a function $x: P \rightarrow \real^3$ where $P = \{1,2,...,j \} \times \{1,2,...,k\}$ is the set of pixels and $x$ assigns to each pixel the vector of its RGB values. Alternatively, as $x$ has discrete domain, one can also write the function table of $x$ in matrix form and we will switch between representations depending on the context.\\
For most computer vision tasks, there exists symmetries in these images that should not affect the output of the neural network. For example, linear shifts of the image
\begin{equation}
tx(p)=x(p+c)
\end{equation}
where $c$ is some constant vector, do not affect the image itself. Even if an image is shifted by a fixed amount (for boundaries, assume that some whitespace is added or removed to allow for movement of the image and assume that whitespace does not influence the predictions), it is still the same image for the purpose of most computer vision tasks.\\
Thus, we can safely assume almost all functions of interest $f^*$ to be invariant under linear shifts:
\begin{equation}
f^*(tx)= f^*(x)
\end{equation}
For another example, consider graphs $G=(V,E,l)$ where $l: V\rightarrow \real^d$ denotes labels for each node. To input this into a neural network, it is natural to represent $l$ by its function table as a matrix $\bar{l}\in \real ^{|V|,d}$\footnote{We use $\bar{x}$ to denote a matrix based representation of some object $x$.} and the set of edges $E$ as an adjacency matrix $\bar{E}\in \real^{|V|,|V|}$. Of course, this representation is heavily dependent on the ordering of the nodes. If one changes the ordering of the nodes, the matrix $\bar{V}$ is permuted
\begin{equation}
\bar{l}' =  \hat{l}P
\end{equation}
where $P$ is a permutation matrix, swapping the columns of the matrix it is multiplied with. A similar transformation happens to the adjacency matrix:
\begin{equation}
\bar{V}' = P^T \bar{V} P
\end{equation}
Intuitively, this denotes a swapping of the rows in $\hat{l}$ and a corresponding change in $\hat{E}$ preserve the structure of the graph. For almost all tasks, the order of nodes is irrelevant as the underlying graph is structurally the same (and, if drawn, can look the same). Thus, one can safely assume almost all functions $f^*$ to be invariant under these kinds of permutations. 
\begin{equation}
f^*(\bar{l}, \bar{E})= f^*(\bar{l}P, P^T\bar{E}P)
\end{equation}
This general pattern of symmetries in the data arises frequently in domains where one would traditionally apply deep learning techniques such as the aforementioned examples of computer vision and graph processing. This pattern occurs in many kinds of data very naturally, for many problem domains there exist transformations that only change the object's \textit{representation} but not the object itself. As an example, images that are rotated are still, in essence, the same image and thus equivalent with respect to most computer vision tasks. Similarly, if one processes triangle meshes in geometric modeling tasks, the modelled object is still the same object, regardless of where it is located in 3d-space or how it is rotated\footnote{In general, this can be extended to isometric transformations, i.e. transformations between meshes such that all angles and areas are preserved. Relocation of the mesh and rotations are examples of these. Deep learning models in these areas are thus heavily concerned with invariance to isometries \cite{bronstein2017geometric}.}.\\
Moreover, this also occurs as a consequence of technical difficulties arising from the use of deep learning. For a function to be trainable by gradient based optimization, it has to be differentiable and as a consequence have a real valued domain and co-domain. Thus, the input to a neural network has to be a matrix and so, too, must its output.\\
\begin{figure}
\centering
\includegraphics[scale=0.3]{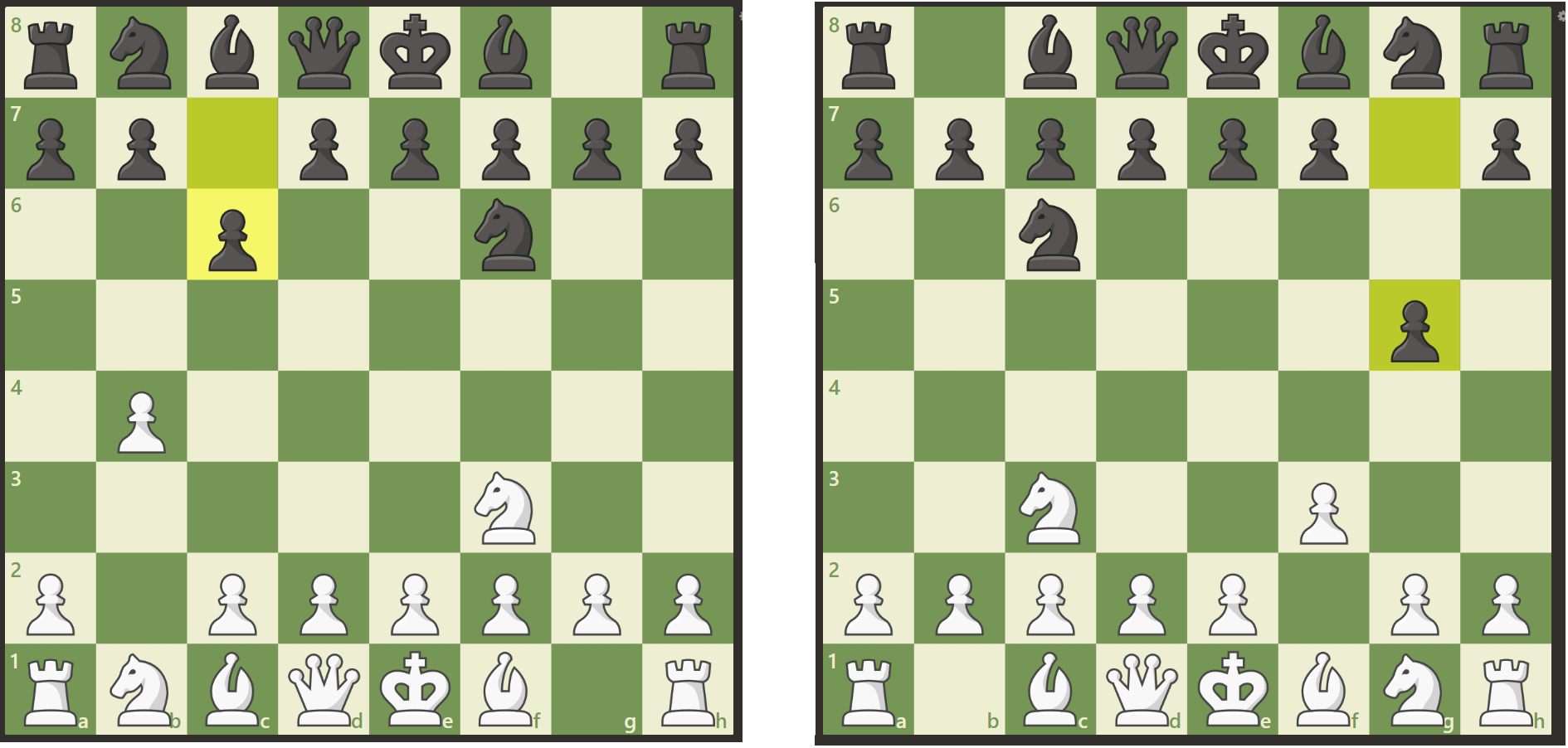}
\caption{An example of symmetry in a domain. In chess, one aims to approximate the function that maps a chess boardstate to the optimal move in that situation. $f^*$ is of course unknown, but it is clear from domain knowledge of chess that the optimal move for black in the leftmost image is the optimal move for white in the rightmost image as chess is symmetric with regard to color (except for whose turn it is). }
\end{figure}
Consider the domain of finite sets containing vectors $\Omega = \{x \; \mid \; x \subset R^n, \text{x finite} \}$. The intuitive approach of transforming such a set into a matrix is to simply "stack" the vectors of the set $x=\{x_1, ..., x_n\}$ into a matrix:
\begin{equation}
\bar{x} = \left( x_1 \; \;  x_2 \; \;  ... \;\;x_n \right)
\end{equation}
This representation faithfully contains all information of the original set $x$, but is not unique. To attain this representation, one chooses an arbitrary ordering of the elements of $x$, a structure that is not present in sets. If one chose a different ordering, a permutation $\pi$ of the above example is attained as a representation:
\begin{equation}
\pi(\bar{x}) = \left( x_{\pi(1)} \; \;  x_{\pi(2)} \; \;  ... \;\;x_{\pi(n)} \right)
\end{equation}
As both are representations of the same set, we know that 
\begin{equation}
f^*(\bar{x}) =  f^*(\pi(\bar{x}))
\end{equation}
In this case, the symmetry does not exist in the data itself but is a consequence of the fact that neural networks require an ordered input and thus impose an additional structure onto the data that is not actually present in the data and thus should not influence our estimation. Still, the general idea is the same: There exists some form of transformations that should not influence the estimation task.\\
In all of these examples, we immediately know just from basic knowledge\footnote{This is very important as neural network architectures are generally designed for a domain and not for a specific task. Thus, prior knowledge of the domain is the most useful kind of knowledge to exploit here.} of our domain that there exist some transformations $T$ that the target function $f^*$ is invariant to. These transformations will be called symmetries and are very useful as they are usually inherent to the data itself and thus known even if one does not know about $f^*$. \\
If one were to use a fully connected neural network, one might end up with an estimator that is not invariant to these transformations, i.e. there exists some $x \in \Omega$, $t \in T$ such that $f(x) \neq f(tx)$. Recalling the extrapolation biases of fully connected neural networks, this is very likely to happen as a linear extrapolation should in most cases violate this condition.\\
Thus, the first additional inductive bias that we seek to include in an architecture is invariance to a set of transformations called symmetries. These transformations usually follow a group structure, usually a permutation group. Therefore, we attain the following additional property that a domain specific neural network should adhere to:
\begin{center}
\textbf{$\grp$-Invariance}. If there exists some  group $\grp$ of transformations on the elements of $\Omega$ that $f^*$ is assumed to be invariant to, $f$ must be $\grp$-invariant, i.e. $f(x)=f(gx)$ for all $g \in \grp$. 
\end{center}
For more information on groups and specifically how our transformations of interest can be expressed as permutation groups, refer to the appendix.
\mybox{$\grp$-Invariance}{Geometric deep learning proposes that domain specific neural networks should be $\grp$-invariant, if $f^*$ is assumed to be $\grp$-invariant: If there exists a group $\grp$ of transformations $\Omega \rightarrow \Omega$ such that we can assume $f^*(x)=f^*(gx)$ for all $g \in \grp$, then for any domain specific neural network $f$ it should also hold that $f(x)=f(gx)$ for all $g$. 
    }
\section{Task Separation and Locality}
\begin{figure}
\centering
\includegraphics[scale=0.5]{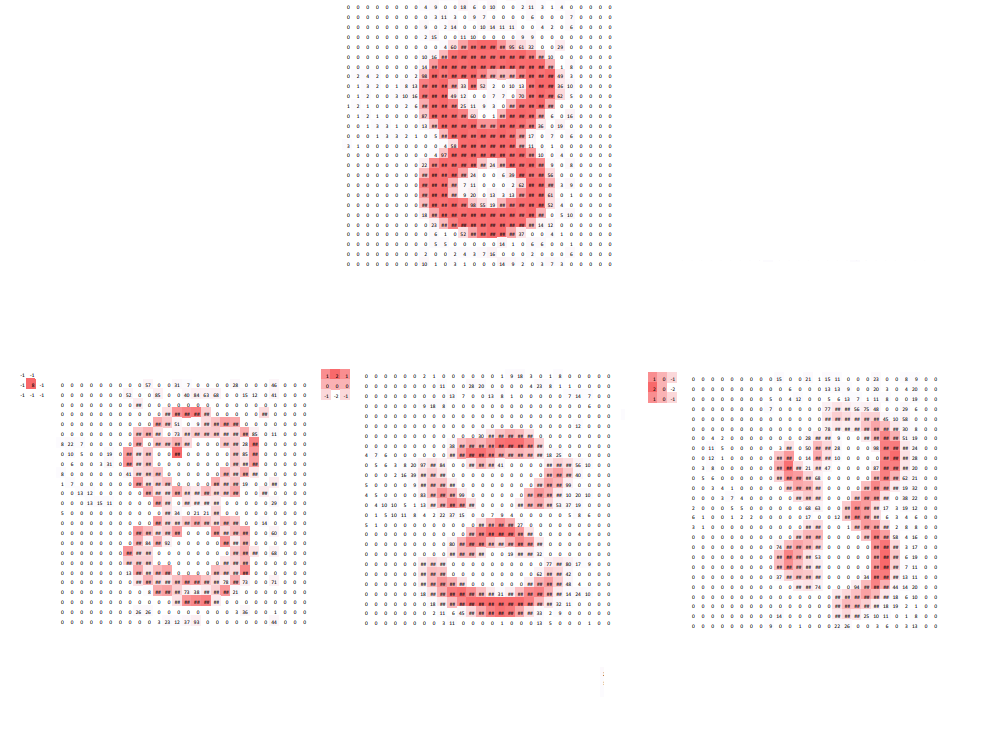}
\caption{The first layer of a CNN acting on an image of an "8" (non-linearity omitted for clarity). The CNN produces three feature maps from the input, the left one roughly capturing the boundary, the second roughly capturing horizontal lines and the third captures vertical lines. This matches the approach in standard image recognition and showcases that CNNs are built in a way that can easily approximate standard image recognition practices. Image from \cite{yamashita2018convolutional}.   }
\end{figure}

The second core idea of geometric deep learning lies in the observation that many deep learning tasks can be separated into many different tasks on different resolution scales.\\
The idea behind this is that in many domains there exists some notion of locality such that local structures can be summarized without losing much global information. One well known example of this occurs in convolutional neural networks made for the domain of computer vision \cite{cnn}. In computer vision, it is usually not one pixel that decides what object is represented by the image. Rather, it is a set of features that are characteristic for that object, i.e. eyes and lips for faces or wheels for cars. All of these characteristics are made up of  pixels that are close to each other in a euclidean sense.\\
Similarly, in many graph based tasks there exist local substructures of graphs that can be summarized without heavily losing information of the entire graph. Thus, one can summarize local substructures of a graph and thus reduce the dimensionality of the approximation problem without losing information that would be needed to properly approximate $f^*$.\\
This leads to the idea of scale separation in neural networks. Intuitively, there are some tasks that require a finely grained solution to solve. To classify small, local patterns in an image, every single pixel in a patch is needed. However, to combine those small patterns into bigger patterns, one does not need pixelwise information and can instead rely on a coarser resolution if the smaller patterns have already been found. For example, when classifying images of cars, a first step might be to find very small patterns such as edges and curves. This step can be easily done directly on the pixels of the image themselves. Then, the next step might be to find wheels (consisting of curves) and license plates (consisting of curves and straight lines). Then, the final classification task of 'does this image represent a car?' can be easily solved by asking "are there four wheels in this image beneath a car body with a license plate". This separates the image classification task, which is in itself decently hard, into a number of much simpler, modular tasks.\\
\begin{figure}
\centering
\includegraphics[scale=0.5]{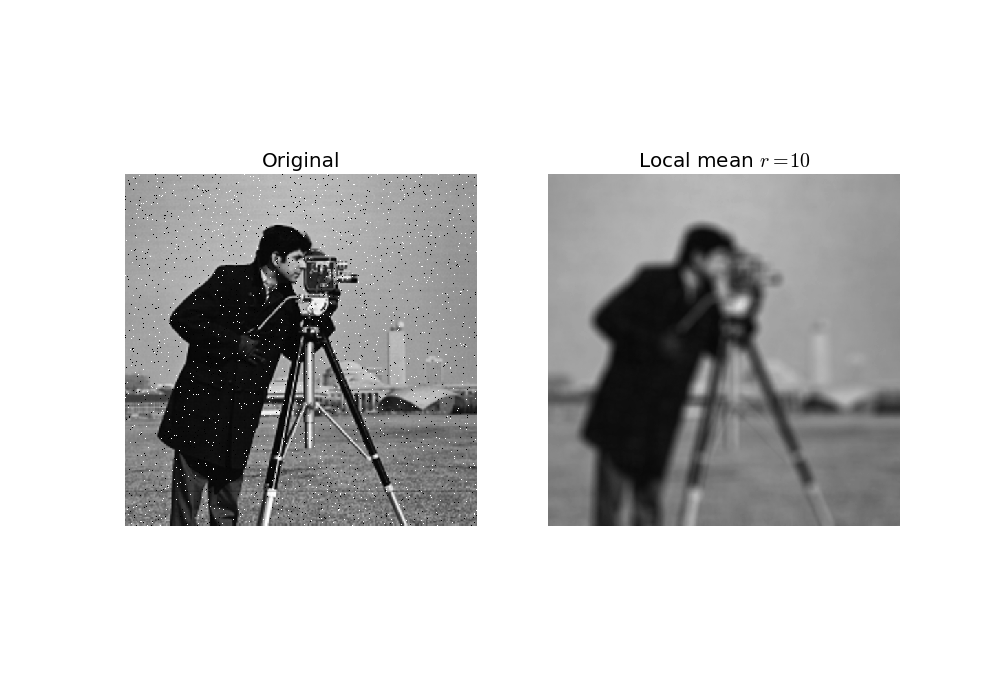}
\caption{An example application of a mean filter. The mean filter here is applied to an original image $x$ instead of the output of some function $\phi(x)$ as would usually be done in a neural network to more clearly showcase its workings. The patch size here is 10, i.e. neighborhoods of 10*10 pixels are aggregated at each step. Image from \cite{van2014scikit}.   }
\end{figure}
The authors of \cite{geo} formalize this as follows. Let $\Omega = \{x \; \mid \; x\in H \rightarrow \real \;, \; H \in S\}$ be the space of functions assigning real values to elements of some gridspace $H$ belonging to a set of gridspaces $S$. As an example, consider computer vision tasks. Here, the gridspace $H$ are indeed regular 2D-grids
\begin{equation}
S = \{ H \; \mid \; H = \{0,1,...,i\} \times \{0,1,...,k\} \}
\end{equation}
of varying size, representing the pixels of an image. The functions $x \in \Omega : H \rightarrow \real$ assign to each pixel of $H$ its grayscale value\footnote{To extend this to colored images, it is sufficient to introduce three channels for each pixel.}. We will use the word 'pixel' to refer to elements of $H$ even in domains that are not computer vision related but note that thus the realization of 'pixels' might be heavily different from actual pixels. \\In graph classification, the gridspaces $H$ are the node sets of respective graphs and the functions $x \in \Omega$ assign to each node a real values label. To also model the edges in this framework, one chooses $H=V \cup  V \times V$ where $x$ assigns labels to each $v \in V$ and $x( (v_1, v_2))=1$ if and only if there exists an edge between $v_1$ and $v_2$.\\
The resolution of some $x:H \rightarrow \real$ is then given as the cardinality of $H$. For example, in images this means that images with more pixels have higher resolution while images with less pixels have lower resolution. In graphs, more nodes means more resolution while less nodes mean a lower resolution.\\
We define a coarsening function $\rho: \Omega \rightarrow \Omega$ as a function that lowers the resolution of its input, i.e. if $\rho(x)=y$, then the gridspace of $x$ is a superset of the gridspace of $y$. In the context of images, this corresponds to an actual coarsening: The new image has less pixels and is thus more coarse, necessarily losing information of the original image. While not clearly indicated in this definition as it is hard to formalize without restricting $\rho$ too much, we also assume that $\rho(x)$ resembles the original $x$ as much as possible. An easy example of this in the context of computer vision is the so called mean pooling. Mean pooling uses one pixel to represent a neighborhood of pixels by their mean
\begin{equation}
\rho(x)(p)= \frac{1}{|N(d)|} \sum_{d' \in N(d)} 	x(d')
\end{equation}\\
Next, we define a locally operating function $\psi:\Omega \rightarrow \Omega$ to be a function that only operates on a certain neighborhood of each pixel. Specifically, we assume that both argument and output of $\psi$ operate on the same gridspace $H$ and that each pixel's value is influenced only by its neighborhood:
\begin{equation}
\psi(x)(p)= \psi'\left( x(p_1), ..., x(p_w)  \right)
\end{equation}
where the neighborhood of $p$ is given by: $N(p)=\{x(p_1), ..., x(p_w)  \}$.\\
Thus, we have defined two different types of functions, one that reduces the resolution of images and one that works locally on small patches of the image. By concatenating these, one attains a separation of a function into many smaller tasks on different scales.\\
We assume that $f^*$ follows such a structure:
\begin{equation}
f^*= \psi_{i+1} \circ \rho_i \circ \psi_i \circ ... \circ \rho_1 \circ \phi_1
\end{equation}
This structure heavily resembles the intuition presented at the beginning of this section. At first, a function $\psi_1$ operates on the local neighborhoods of pixels. Then, the result is coarsened into a coarser domain $H_2$ and another function $\psi_2$ operates on the local neighborhoods of pixels in $H_2$. As $H_2$ is coarser, the meaning of neighborhood also extends. 
For example, if one uses mean pooling to aggregate neighborhoods of four adjacent pixels, one attains an image with a quarter of the original resolution. Now, each point in the new image represents 4 points in the original image. If $\psi_2$ now acts on the four direct neighbors of points $p$, each of these points represents four points in the original image. Therefore, $\psi_2$ depends on a neighborhood of 16 points in the original image whereas it would have depended only on neighborhoods of four pixels before the coarsening. Therefore, with continued coarsening, larger parts of the original image are processed. However, direct interactions on a pixel-wise scale are not visible at more coarse scales. For example, $\psi_2$ depends on four means of four points each in the original image. It is impossible for $\psi_2$ to depend on the exact location of each pixel within its neighborhood as this information is lost in the coarsening. The concrete position of each of these pixels that make up one mean cannot be determined, only the fact that they are part of a given neighborhood.\\
This is the core of the assumption of scale separation: Finely grained interactions (for example interactions between two pixels) that depend on the exact position of pixels happen only between directly neighboring nodes while parts of the image that are further apart interact only on larger scales; the larger the distance; the more coarsely grained the interaction and the less the dependence on where the pixel actually is. This idea extends similarly to other domains than image recognition.\\
This assumption of scale separation is a much more specific assumption than invariance. First off, it pertains only to domains that have the structure of $\Omega$ that was described here, i.e. data consists of maps $H \rightarrow R$, and relies heavily on the fact that not only $f^*$ is separable in the way described here, but that one also knows the correct notion of neighborhood with respect to $f^*$. Nevertheless, it applies in many of the most common deep learning domains such as computer vision where the notion of neighborhood is trivial, while other domains such as sets do not really incorporate this structure\cite{geo,zaheer2017deep}.\\	
Hence, we will separate this into two different, less strict assumptions and denote the following inductive biases as desirable
\begin{center}
\textbf{Locality}. If there exists a meaningful notion of 'locality' within $\Omega$ and if it is reasonable to assume that direct interactions that heavily influence $f^*$  are local, then the estimator $f$ should be built from locally operating functions.
\end{center}
and
\begin{center}
\textbf{Task Separation}. If it is reasonable to assume that $f^*$ can be separated into multiple different simpler functions, then $f$ should adhere to a structure that follows a similar separation.
\end{center}
Both of these concepts are heavily important in deep learning. For example, the highly successful CNN \cite{cnn} relies very heavily on these concepts. The big advantage of the architectures that adhere to these inductive biases lies in the fact that they manage to reduce one very highly dimensional approximation task to a lower dimensional one. This is very useful, but the theoretical understanding of the effects of these inductive biases on the space of functions that can actually be approximated appears to be limited, which is why they will be covered only briefly in this work.\\
It is interesting to note however, that at least in some domains, multiscale structures are associated with some form of approximate invariance to deformations as outlined in \cite{geo}. In the case of CNNs, the authors of \cite{bietti2019group} managed to show that CNNs are relatively stable under diffeomorphisms using the above ideas.\\
In particular, their work pertains a real valued extension of CNNs, i.e. CNNs that process images on a continuous pixelspace such that $H \subseteq \real^2 $. They consider displacement maps $\tau :\real^2 \rightarrow \real^2$ that move the pixels of the image such that $\tau(p)$ indicates how much the pixel $p$ is moved. This induces a transformation of the image given by
\begin{equation}
\tau(x)(p) = x(p-\tau(p))
\end{equation}
For these differentiable displacement maps, they show that for all CNNs $f$ and all diffeomorphisms $\tau$ it holds that
\begin{equation}
\dist{f(\tau x) - f(x)} \leq  (C_1 \sup_{p \in H} \dist{\nabla \tau(p)} + C_2 \sup_{p \in H} \dist{ \tau(p)}) \dist{x}  
\end{equation}
Intuitively, the bound is made up of two separate parts, one depending on the size of the displacement itself, $\sup_{p \in H} \dist{ \tau(p)}$, and one depending on the gradient of the displacement,\\$\sup_{p \in H} \dist{\nabla \tau(p)}$. The latter is of special interest here as it pertains to deformations, i.e. non-linear transformations that go beyond just translations. As the authors of \cite{bietti2019group} show, the corresponding factor $C_1$ is dictated by a factor of $O(k^{l+1})$ where $l$ is the number of layers in the network and $k$ is the pooling size used, i.e. the number of pixels that are aggregated by pooling at each step. The quantity $O(k^{l+1})$ dictates the size of the receptive field of neurons in the last layers, i.e. how many pixels of the original image influence each neuron in the last layer before the final classification layer. Therefore, CNNs with a low amount of layer and very small pooling filters are approximately invariant to small deformations $\tau$.\\
Note however, that this invariance applies only very little to modern CNNs in practice, as these typically involve a very large amount of layers. As shown in \cite{ruderman2018pooling}, the approximate deformation invariance of CNNs is highly complex and not yet sufficiently understood.
\mybox{Task Separation and Locality}{Geometric deep learning proposes that domain specific neural networks should follow scale separation of $f^*$. This implies that any good domain specific neural network should follow task separation and locality of $f^*$. Task separation means that, if there exists some compositional structure that is assumed for $f^*$, then a good domain specific neural network $f$ should follow this structure. Locality means that, if there is a notion of locality that exists on $\Omega$, i.e. if there are local substructures of the entire input $x$, then those substructures should be closely connected in the domain specific neural network.
    }
\section{Domain Specific Neural Network Desiderata}
From this, we can add the domain specific ideas of geometric deep learning, i.e. task separation, locality and $\grp$-invariance, to the properties of fully connected neural networks that we found desirable. Thus, we attain a set of desiderata, i.e. some properties that a domain specific neural network should have, that combines the useful properties of the fully connected neural network and the domain specific properties that were proposed by geometric deep learning.
These desiderata are given by:
\begin{enumerate}
\item \textbf{Trainability.} The function needs to be trainable, i.e. parameterized with some parameter vector $\theta$ and differentiable with respect to said parameters.
\item \textbf{Continuity.} We assume that $f^*$ is continuous and, furthermore, that it is lipschitz continuous with some lipschitz constant that is not arbitrarily large. As a consequence, we assume that $f^*(x) \approx f^*(y)$ holds if $x$ and $y$ are relatively close. 
\item \textbf{Tunability.} As different problems in a domain require different strength of the lipschitz continuity, we desire this property to be tunable in some way. It is especially important, that there exists some way to exponentially increase the maximum lipschitz constant to scale to highly complex tasks with sublinear effort.
\item \textbf{Completeness.} If given enough resources, the architectural class should be able to approximate all functions that do not violate any of the other restraints to arbitrary precision, i.e. $rep(f)=S$ where $S$ is the set of all plausible candidates for $f^*$.
\item \textbf{$\grp$-Invariance}. If there exists some group $\grp$ acting on $\Omega$ that $f^*$ is assumed to be invariant to, $f$ must be $\grp$-invariant, i.e. $f(x)=f(gx)$ for all $g \in \grp$. 
\item \textbf{Locality}. If there exists a meaningful notion of 'locality' within $\Omega$ and if it is reasonable to assume that direct interactions that heavily influence $f^*$  are local, then the estimator $f$ should be built from locally operating functions.
\item \textbf{Task Separation}. If it is reasonable to assume that $f^*$ can be separated into multiple different simpler functions, then $f$ should adhere to a structure that follows a similar separation.
\end{enumerate}
This is the answer that we can give to the question ''What does a good domain specific neural network look like?'' using the ideas of geometric deep learning. Now, we will start trying to answer the question: ''How can such a neural network be constructed?''.
To do this, we will first start by deriving \cite{zaheer2017deep} from this set of desiderata. At first we will do this in an unstructured way and from there on motivate the geometric deep learning blueprint \cite{geo} because we believe this is one application where the ideas of said blueprint are very visible.
\section{Deep Sets as Domain Specific Neural Networks and the GDL Blueprint}
Sets are one of the most fundamental and important mathematical objects and it is hugely important to be able to have suitable neural network architectures to process them. As outlined before, neural networks can only ever take in some number of real values as input and, as a baseline, assume that they are ordered. In consequence, fully connected neural networks are in fact dependent on the order of their inputs, which poses problems when dealing with sets that do inherently not possess any notion of ordering of their elements.\\
We consider finite sets of real numbers which is the most basic use case for neural networks. Note that everything presented in this section applies to not only set but rather multisets, i.e. sets where values can occur multiple times. Nevertheless, we present it for the use case of sets this section presents deep sets that were designed for sets \cite{zaheer2017deep}. We consider $\Omega = \{ x \; \mid \; x \subset \real, x \text{ finite} \}$. As outlined in Section 3.3, we represent a set $x= \{x_1,...,x_n\} \in \Omega$ as a vector
\begin{equation}
\bar{x} = \left( x_1 \; \;  x_2 \; \;  ... \;\;x_n \right)
\end{equation}
which is necessary as neural networks need ordered real values as input. The ordering is of course arbitrary and we need to define a trainable function (i.e. a neural network) that is architecturally guaranteed to be invariant to permutations of this ordering.\\
We will start with property 4 of the domain specific neural network desiderata, i.e. the aforementioned invariance to permutations of the set elements. We seek to construct $f$ such that the ordering of the nodes does not matter. An easy way to do this is to simply sum over all of the values $x_i$. If we define
\begin{equation}
\label{sum}
f(x) = g(\sum_{i=1}^n x_i)
\end{equation}
the function $f$ is guaranteed to be invariant to permutations of the input. If $g$ is trainable, continuous and tunable, we also immediately fulfill the desired properties 1 and 2. Of course, there are immediately obvious problems with this approach. Using this formulation of $f$, input values with the same sum cannot be distinguished from eachother.  For example, $f(\{4,3\})= g(7)=f(\{2,5\})$, violating the sixth property. Therefore, the function needs to be more complex.\\
We can proceed by exploiting the Kolmogorov-Arnold representation  theorem\cite{kolmogorov1961representation}. According to the Kolmogorov-Arnold representation theorem, every continuous multivariate function $f^*$ can be expressed by composition of a sum and univariate functions:
\begin{equation}
f^*(x_1, x_2, ... ,x_n)\: =\: \sum_{q=0}^{2n} \:  \Phi_q( \sum_{p=1}^n\phi_{q,p}(x_p))
\end{equation}
Note the very peculiar construction of this: There exists a number of univariate functions $\phi$ that are applied to the singular inputs $x_p$, then summed, the passed to $\Phi$.\\
\begin{figure}
\centering
\includegraphics[scale=0.7]{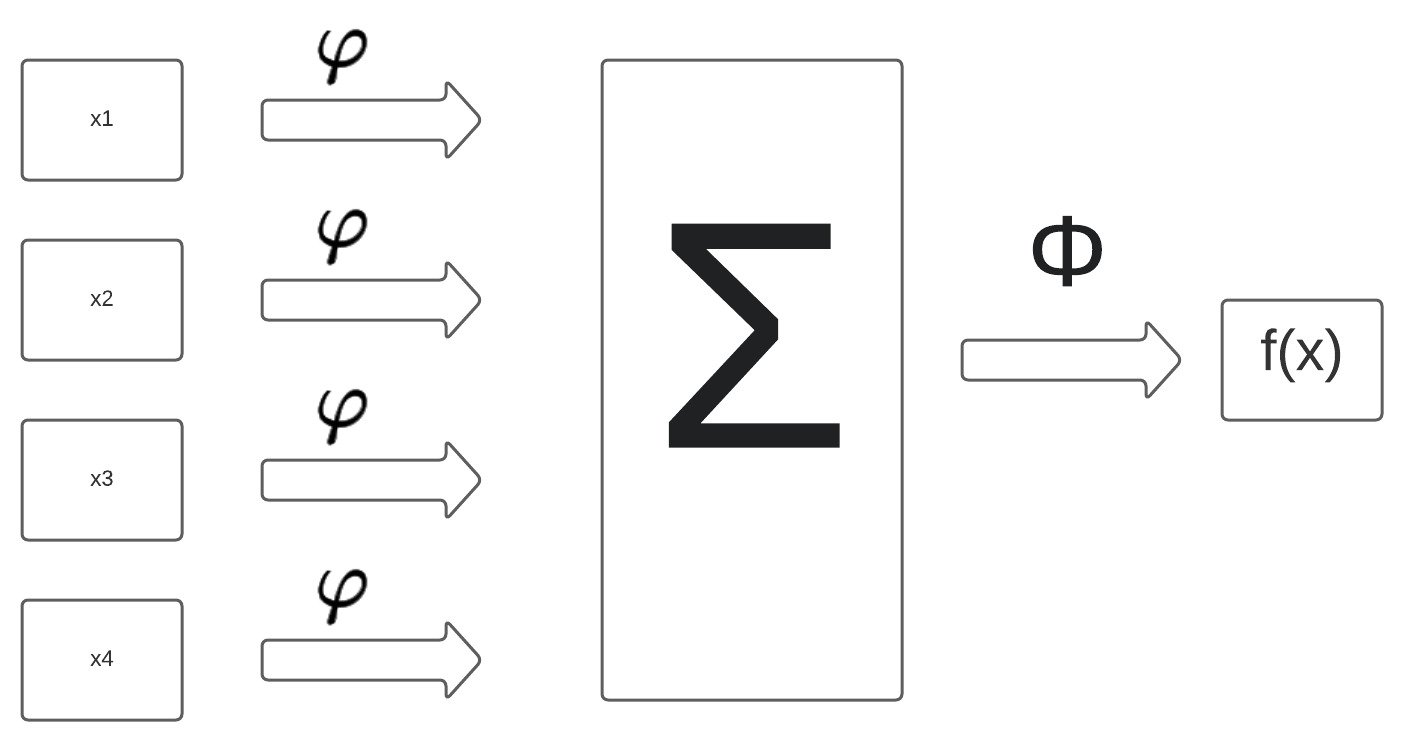}
\caption{The structure of a deep set architecture acting on a set with four elements $x=\{x_1, x_2, x_3,x_4\}$. to each element, $\phi$ is applied. The results are then summed to attain the final result $f(x)$. Note that, if one were to swap $x_2$ and $x_3$ in this architecture, the final result would remain unchanged as }
\end{figure}
This, at least partially, resembles Equation \ref{sum} and can be further simplified. The above is a Kolmogorov-Arnold representation for arbitrary functions, we are however interested in only permutation invariant functions. For those, the $\phi$ can be uniform as all $x_i$ behave in the same way and the outer sum can be dropped \cite{zaheer2017deep}:
\begin{equation}
f^*(x_1, x_2, ... ,x_n)\: =\:  \Phi( \sum_{p=1}^n\phi(x_p)) 
\end{equation}
where $\Phi : \real^n  \rightarrow C$ and $\phi : \real \rightarrow \real^n$ are continuous\footnote{One could attain simpler versions of this representation, but continuity is of crucial importance to neural networks.}. The difference to Equation \ref{sum} lies in the fact that the $x_i$ are not immediately summed, losing information, but mapped beforehand into a $n$-dimensional vector space. Intuitively, the vector $\sum_{p=1}^n\phi(x_p)$ contains just as much information as the original values did and therefore, the mapping induced by $d(x)=\sum_{p=1}^n\phi(x_p)$ can be continuous and invertible, up to permutation. If one chooses $\Phi=f^* \circ d^{-1'}$ one can represent every continuous, permutation invariant function $f^*$ in this way.\\
Thus, we build our network architecture in this manner:
\begin{equation}
f(x_1, x_2, ... ,x_n)\: =\:  \Phi( \sum_{i=1}^n\phi(x_p)) 
\end{equation}
and we can be sure to fulfill the completeness-property and the $\grp$-Invariance. Of course, we have to find fitting functions for $\Phi$ and $\phi$ to ensure that the final function is continuous, trainable and tunable and that we can indeed approximate $\Phi$ and $\phi$ such that every function $f^*$ is approximable. Luckily, this is relatively easy. Both $\Phi$ and $\phi$ are arbitrary functions without any kind of prior knowledge on our side. Therefore, we can use simple fully connected neural networks for both $\Phi$ and $\phi$. Thus, all of the properties of fully connected neural networks that were desirable are maintained. Trainability, tunability and continuity all are maintained through composition and summation. Similarly, completeness and $\grp$-invariance are maintained as shown by \cite{zaheer2017deep} for the former and as immediately follows from construction for the latter.\\
This leaves locality and task separation to be checked. Due to the very nature of sets, all elements behave symmetrically. Therefore, either all elements are local to each other or elements are local only to themselves. One can easily see that this formulation employs scale separation and therefore both task separation and locality. $\phi$ operates on individual elements, whereas $\Phi$ operates on all elements.\\
Deep sets are interesting for two reasons, one, in that they are of interest in themselves as a building block for other neural networks whenever sets need to be processed. Second, they give some intuition on how to build neural network architectures that fulfill all of the desired GDL properties.\\
\mybox{Deep Sets}{
	Deep sets are a domain specific neural network architecture for (Multi-)sets that follow the ideas of geometric deep learning. (Multi-)sets contain multiple elements without order.
	Deep sets are trainable functions $f :\real^i \rightarrow \real^o$ that are invariant to the order of their inputs. Deep sets $f$ are described by
	\begin{equation}
f(x_1, x_2, ... ,x_n)\: =\:  \Phi( \sum_{p=1}^n\phi(x_p)) 
\end{equation}
where $\Phi$ and $\phi$ are fully connected neural networks. Deep sets are universal approximators for the set of continuous functions $\real^n \rightarrow \real^k$ that are invariant to permutations of their inputs.	
    }
\subsection{The GDL Blueprint}
After we have seen how deep sets can be derived from the domain specific neural network desiderata and how they fulfill them, we want to take a look at a more structured approach to this. Geometric deep learning proposes to this end the geometric deep learning blueprint that we will motivate here from the example of deep sets.\\
The first three properties of the domain specific neural network desiderata, i.e. continuity, trainability and tunability were relatively easily achieved by incorporating fully connected neural networks wherever an arbitrary, unknown function had to be substituted. Similarly, locality and task separation were easy to enforce by simply constructing the network accordingly to the proposed structure.\\
The much harder task was the implementation of $\grp$-invariance and completeness. As was seen in the example of deep sets, invariance itself is not usually hard to enforce. For finite groups $\grp$, one can always enforce invariance by averaging over the orbit of $x$\footnote{For additional information on groups and orbits, consult the appendix.}
\begin{equation}
f(x)= f'( \sum_{g \in \grp} gx)
\end{equation}
As the orbits are an equivalence relation, one easily checks that this formulation is indeed invariant to transformations $g\in \grp$. This is a form that is guaranteed to yield $\grp$-invariant functions. Now, one could be led to believe that creating a $\grp$-invariant neural network would be trivial, if one simply applies the construction above and chooses a fitting function approximator for $f'$, for example a fully connected neural network.\\
However, this is not actually quite that easy. Let us consider the above term in the case of deepsets where $\grp$-invariance means invariance to permutations. Thus, if we have $x=(x_1, x_2, ..., x_n)$, then this sum becomes:
\begin{equation}
f(x)= f'( \sum_{\pi \in \grp} \pi(x))
\end{equation}
over all possible permutations of $x$, which is vector valued. Consider $a=\sum_{\pi \in \grp} \pi(x)$, the input to $f'$. For its first entry, it holds that:
\begin{equation}
a_1 = x_1 + \pi_1(x_1) + \pi_2(x_2)...
\end{equation}
Now, consider the fact that there exists $n!$ permutations $\pi$ that map $x_1$ to $x_1$, $n!$ permutations that map $x_1$ onto $x_2$ and so on. Therefore, this becomes:
\begin{equation}
a_1= n! x_1 +n!x_2+... + n!x_n=n!*\bar{x}
\end{equation}
The same holds for $a_2,...,a_n$ and thus $f'$ depends only on the average of $x$. This holds not only when $\grp$ is the symmetric group, but similarly for all of its subgroups. Of course, such an estimator cannot be complete as it can, for example, not distinguish the inputs $(1,2)$ and $(0,3)$, even though our assumptions do not prescribe this. Therefore, this violates our desired completeness property. \\
Thus, we need to preface this aggregation by some nonlinear function. If we write
\begin{equation}
f(x)= f'( \sum_{g \in \grp} gf''(x))
\end{equation}
where $f''$ is a well chosen continuous, nonlinear function $\Omega \rightarrow \Omega$, then, as shown by adapting the kolmogorov representation theorem, we can in fact represent all of our desired functions, given that the right $f'$ was chosen. However, again due to the kolmogorov representation theorem, this presentation also allows for $f(x)$ to be non-$\grp$-invariant, defeating the purpose. To prevent this, we need to restrict $f''$.\\
Consider the example with $f''(x)=(x_1^2, x_2)$. Then, $f''((2,1))= (4,1)$ and $f''((1,2)=(1,2)$. Then, for $(2,1)$: 
\begin{equation}
\sum_{g \in \grp} gf''(x))=(4,1)+(1,4)=(5,5)
\end{equation}
while for $\pi(2,1)=(1,2)$:
\begin{equation}
\sum_{g \in \grp} gf''(x))=(2,1)+(1,2)=(3,3)
\end{equation}
and therefore $f'$ is not guaranteed to be invariant to permutation. The intuitive reason for this lies in the fact that $f''$ can encode the input in a way that depends on the ordering and thus the permutation. Still, we require a nonlinear $f''$ to ensure completeness.\\
Thus, $f''$ needs to be a non-$\grp$ invariant function that is nonlinear but also cannot be influenced by group actions $g \in \grp$ in meaningful ways as otherwise the final function is not $\grp$-invariant anymore. Therefore, the embedding needs to be orthogonal to $\grp$, i.e.:
\begin{equation}
d(gx)=gd(x)
\end{equation}
That is, the result of $d$ is not influenced by $g$ and leaves $g$ unchanged. This property is called $\grp$-equivariance. For example, in deep sets, $d$ is given by the function $\phi$. $\phi$ works on each node separately. It is easy to see that applying $\phi$ to the vector of stacked $x_i$'s yields a vector of stacked $\phi(x_i)$. Then, observe that first applying $\phi$ and then permuting the rows is equivalent to first permuting the rows and then applying $\phi$, yielding the $\grp$-invariance that was described. This yields the  geometric deep learning blueprint that ensures $\grp$-invariance.\\
According to the geometric deep learning blueprint, a GDL-model should be built as:
\begin{equation}
\label{blueprint}
f=\Phi \circ \phi_n \circ \phi_{n-1} .... \circ  \phi_1
\end{equation}
where all $\phi_i$ are $\grp$-equivariant and $\Phi$ is $\grp$ invariant. This achieves an easy way to stack layers by increasing the number of $\phi_i$, yielding  tunability and very frequently completeness. Similarly, due to the compositional design of this network, continuity and trainability of the singular functions extends to the composition of all of them. Thus, if we restrict $\Phi$ and all $\phi_i$ to be trainable, lipschitz continuous functions, we retain those properties for $f$. Further, we need to ensure that every function can be approximated by this construct. To do this, it is important that the $\phi_i$ do not lose relevant information of the original input. Therefore, we demand that each $\phi_i$ can, given correct weights attained during tranining, be a bijective function $\Omega \rightarrow \Omega$.\\
Concerning the $\grp$-invariance, a simple inductive argument shows that any $f$ constructed in such a fashion is $\grp$-invariant:
\begin{align*}
f(gx)&= (\Phi \circ \phi_n \circ \phi_{n-1} .... \circ  \phi_1)(gx)\\
&=(\Phi \circ \phi_n \circ \phi_{n-1} .... \circ  \phi_2)( \phi_1(gx))\\
&=(\Phi \circ \phi_n \circ \phi_{n-1} .... \circ  \phi_2)(g \phi_1(x))\\ 
...\\
&=(\Phi (g (\phi_n \circ \phi_{n-1} .... \circ  \phi_2 \circ \phi_1)(x))\\ 
&= (\Phi \circ \phi_n \circ \phi_{n-1} .... \circ  \phi_2 \circ \phi_1)(x)\\
&=f(x)
\end{align*}
where the $\grp$-equivariance of the $\phi_i$ and the $\grp$-invariance of $\Phi$ is used. It is easy to check that deep sets obey this general blueprint.\\
Putting this together, the geometric deep learning blueprint reduces the problem of creating a $\grp$-invariant neural network to the definition of two functions
\begin{enumerate}
\item The feature representation function $\phi$ : $\Omega \rightarrow \Omega$. This function should be trainable, lipschitz continuous and $\grp$-equivariant.
\item The classification function $\Phi$ : $\real^k \rightarrow C$ that does the final classification. This function should be trainable and lipschitz continuous and $\grp$-invariant. It can, but does not have to, be constructed as $f(x)= f'( \sum_{g \in \grp} gx)$.
\end{enumerate}
This blueprint was proposed by the authors of \cite{geo} as a way of building neural networks that fulfil the domain specific deep learning properties, mainly $\grp$-invariance. As locality and task separation involve mirroring the assumed structure of $f^*$, it should usually be very clear how they should be implemented.\\
For the remainder of our domain specific neural network desiderata however, it is not clear that this blueprint necessarily fulfills them. For example, it is not guaranteed that this blueprint necessarily yields a complete architecture that can indeed approximate all functions of interest.
\section{Geometric Deep Learning Summary}
In this chapter, we tried answering the questions ''what do good domain specific neural networks look like?'' and ''how can we construct them?'' using the ideas of geometric deep learning.\\
For the first question, we drew up a list of properties that we believe domain specific neural networks should possess. This list is made up of some properties of fully connected neural networks that are useful \textit{almost} independently of context. Those contain trainability (which ensures that gradient based optimization can be used), completeness (the ability to approximate any function of interest to arbitrary degree) and the lipschitzness of the estimator which ensures that points in a euclidean neighborhood are treated similarly by our estimator. On top of these generically useful properties, we added three domain specific properties that were taken from the ideas of geometric deep learning. These include task separation, i.e. if the function to be approximated is assumed to have some compositional structure, then our estimator should follow the same structure, locality, i.e. the idea that, if there exists a notion of locality on $\Omega$, then our estimator should treat local neighborhoods as a combined entity, and lastly $\grp$-invariance, i.e. if there exists some group of transformations that $f^*$ is invariant to, then so should be $f$.\\
This list of desiderata is not necessarily complete and it is hard to say that it is. But, all of the non-domain specific properties in this list are necessary for an optimal estimator, for example, an incomplete estimator that cannot approximate all functions of interest has to necessarily fail at some tasks. For the non-domain specific properties taken from geometric deep learning, it is hard to evaluate how crucial they are to a neural networks success. For the example of $\grp$-invariance, we will discuss their impact in a later chapter.\\
For the question of ''how can we construct such architectures?'', we first started by showing how deep sets can be derived from these desiderata. Deep sets are a successful neural network architecture that deals with sets. In fact, in their original presentation the inventors of deep sets explicitly rely on the ideas of $\grp$-invariance and completeness \cite{zaheer2017deep}. Yet, we found that this process involved heavy manual work and domain expertise. Thus, we introduced the geometric deep learning blueprint, the proposed approach to designing such neural networks that was introduced as part of geometric deep learning \cite{geo}. This blueprint requires only the definition of two functions, one that is $\grp$-invariant and one that is $\grp$-equivariant. If both are trainable and lipschitz continuous, we are guaranteed to attain a $\grp$-invariant, continuous and trainable function. However, the other desired properties require additional work and are not guaranteed.\\
Now that we have answered the questions that we posed at the beginning with the ideas of geometric deep learning, we will try to evaluate their impact. To do this, we will observe them using graph neural networks as a case study. There, we will observe how graph neural networks are designed to fulfill the domain specific neural network desiderata and how these ideas are reflected in them. Afterwards, we will try to apply the geometric deep learning blueprint and find out whether one could feasibly create graph neural networks using said blueprint.\\
After this case study, we will try to evaluate the theoretical impact of the domain specific neural network properties that were proposed by geometric deep learning. As task separation and locality are harder to quantify and exceed the scope of this work, we will focus on $\grp$-invariance and showcase some theoretical results that quantify the impact of $\grp$-invariance on the performance of our estimator.

%% file: kapitel/gnn.tex
\chapter{Graph Neural Networks as Geometric Deep Learning Instances}
In the previous chapter, we laid out how geometric deep learning answers the questions of ''how do good domain specific neural networks look like and how do we build them?''.  In this section, we will use graph neural networks as a case study where we will try to see how these ideas are implemented in a practical, state of the art domain specific neural network architecture in graph neural networks.\\
Graph neural networks, as presented by Scarselli et. al \cite{scarselli2008graph}, are used as a case study here because they are one of the most prominent examples of domain specific neural network and have achieved interesting results in the last decade in fields where deep learning has traditionally struggled to achieve good results such as SAT solving \cite{selsam2018learning} and the traveling salesman problem \cite{prates2019learning}, showcasing the potential of domain specific neural networks.\\
Further, they adhere to the design philosophies of geometric deep learning very rigidly as described in \cite{geo} and are thus well suited as a case study to see how the ideas of geometric deep learning manifest in an actual state of the art neural network architecture.\\
Graph neural networks are a domain specific neural network architecture for dealing with graphs.\\ 
This section will proceed as follows: First off, graph neural networks will be introduced. This introduction will first present the Weisfeiler-Lehman test, a test that uses graph coloring to approximately decide whether two graphs are isomorphic or not. This test is the basis of graph neural networks, as graph neural networks essentially are a version of this test with certain parts of it replaced by fully connected neural networks.\\
After we have introduced this test, we will use it to derive graph neural networks from the domain specific deep learning desiderata and show how graph neural networks adhere to those principles.\\
Once this has been done, we will evaluate the geometric deep learning blueprint on the example of graph neural networks. As presented in this chapter, graph neural networks are derived from the Weisfeiler Lehman test and thus are based on strong expert domain knowledge. The geometric deep learning blueprint promises to alleviate this. If it is indeed useful, it should be possible to derive graph neural networks from it \textit{without} needing much additional domain knowledge. Thus, we will try to do just that and evaluate whether the geometric deep learning blueprint is sufficient for the derivation of graph neural networks. 

\section{The Weisfeiler-Lehman Test}
A graph isomorphism between two graphs $G_1=(V_1, E_1, l)$ and $G_2=(V_2, E_2,l)$ is a bijective function $\phi:V_1 \rightarrow V_2$ that respects edges and labels, i.e. 
\begin{equation}
l_1(v) = l_2(\phi(v))  \forall v \in V_1
\end{equation}
\begin{equation}
(u,v) \in E_1 \leftrightarrow (\phi(u),\phi(v)) \in E_2
\end{equation}
intuitively, isomorphic graphs, i.e. graphs between which a graph isomorphism exists, are structurally equivalent, only differing in the naming of their nodes. This is a highly interesting form of symmetry for the purposes of geometric deep learning as almost all meaningful tasks on graphs are concerned only with the structure of the graph and thus can safely be assumed to be invariant with respect to isomorphisms.\\
If there exists an isomorphism $\phi$ between two graphs such that  $u=\phi(v)$, we will call $u$ and $v$ isomorphic nodes. Further, with a slight abuse of notation, we will also let isomorphisms $\phi$ act directly on graphs $G=(V,E)$ such that $\phi(G)=G'$ where $G'=(V',E',c')$ with $V=\{ \phi(v_1), \phi(v_2),...\}$, $E'=\{\;(\phi(v),\phi(u)) \; \mid \;  (v,u) \in E\}$ and $l'(\phi(v))=l(v)$ and switch between the two representations, i.e. an isomorphism mapping nodes of two graphs to each other and an isomorphism mapping graphs onto each other, when context demands it.\\
While GNNs are supposed to be invariant under graph isomorphisms, it is hard to attain an exact invariance under graph isomorphisms. More specifically, it is hard to create neural networks that are both fast to evaluate, can discriminate all non-isomorphic graphs and are also invariant under isomorphism. This is rooted in the fact that graph isomorphism is in itself a hard problem to decide. While it is not known to be NP-complete, there also currently does not exist a polynomial time algorithm to solve it \cite{fortin1996graph}. As a consequence, a neural network architecture that is invariant exactly to isomorphisms would probably require an exponential runtime which is unfeasible for most tasks where one would employ a neural network.\\
Still, it is possible to use approximative isomorphism tests that either over- or underapproximate the true isomorphism relationship. One such example is the Weisfeiler-Lehman test, a very easy but useful approximation based on iterative abstraction refinement implemented as graph coloring \cite{morris2019weisfeiler}. The WL test is strongly related to GNNs and it was shown by \cite{morris2019weisfeiler} that GNNs are exactly as powerful as this test.\\
We will briefly review the WL-test before moving on to GNNs as it showcases the core ideas of GNNs with regards to graph isomorphism while not requiring the machine learning details to make it trainable.\\
The WL-Test computes a canonical signature of graphs that can be used to test if they are isomorphic. If two graphs have the same signature, they might be isomorphic. If two graphs have different signatures, they are guaranteed to be not isomorphic.\\
\begin{figure}[htbp]
\centerline{\includegraphics[scale=.5]{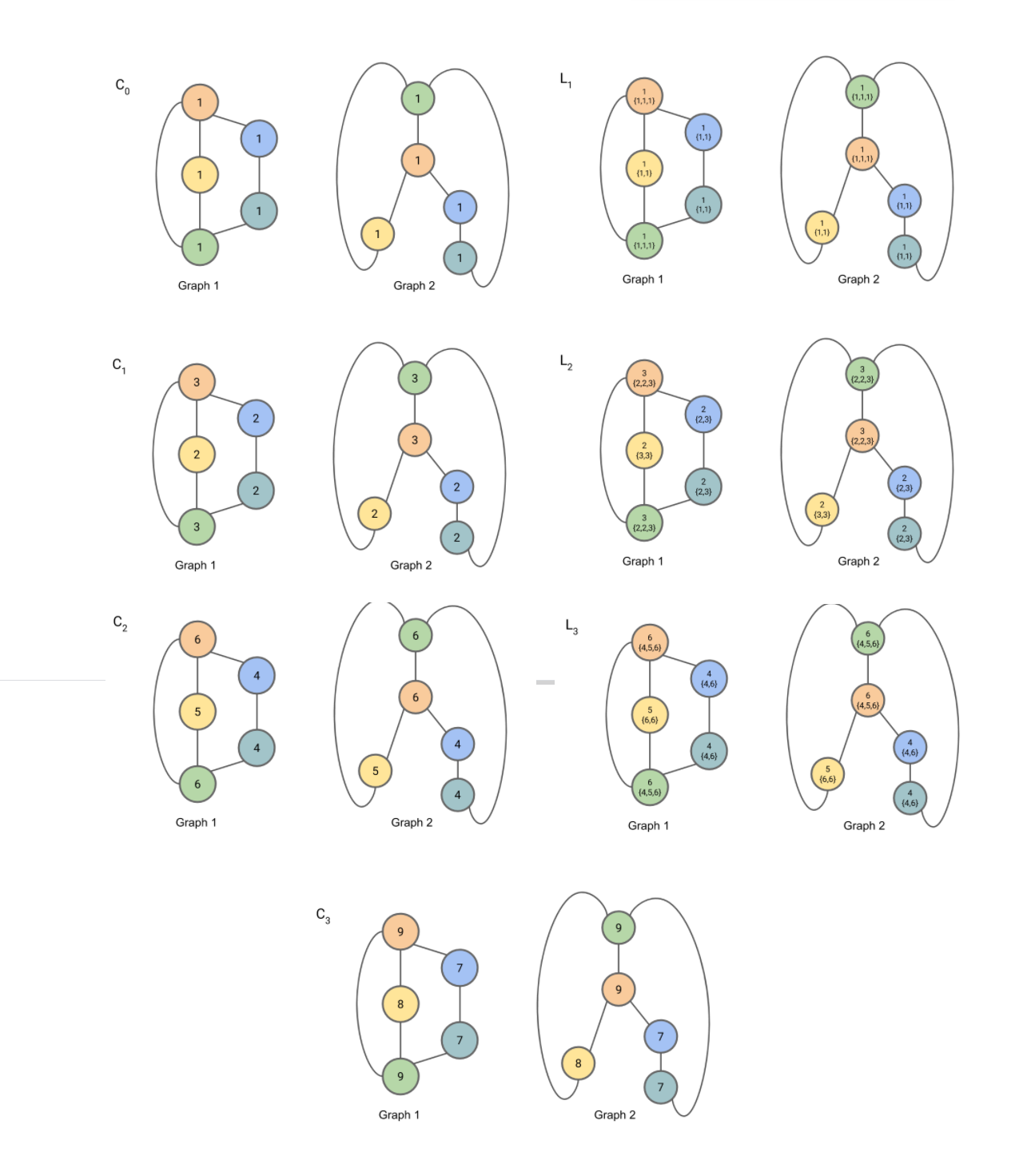}}
\caption{A run of the Weisfeiler Lehman Test on two isomorphic graphs. As both graphs are isomorphic, they yield the same signature "9,9,8,7,7". For both graphs, the coloring at each step is shown on the left hand side and the multisets representing each nodes neighborhood on the righthandside. Note that in the first iteration all nodes have the same color and that in the second iteration nodes are distinguished by their outdegree. Example taken from \cite{WL} }
\end{figure}
We will, for ease of notation, consider undirected, unlabeled graphs in this section but the WL test and as a consequence GNNs extend very easily to all kinds of different settings.\\
Consider a graph $G=(V,E)$. The WL-test assigns a color $c(v)$ to each node $v$ and performs multiple refinement steps where these colors are updated. Intuitively, if for two nodes it holds that $c_t(v)=c_t(u)$, the WL-test could not disprove at timestep $t$ that these nodes could be isomorphic to each other.
A priori, all nodes could potentially be isomorphic to one another, so the WL-test assigns the same color to each node:
\begin{equation}
c_0(v) = 0
\end{equation} 
For labeled graphs, nodes can only be isomorphically mapped onto each other if they have the same label. Therefore, in labeled graphs:
\begin{equation}
c_0(v)=l(v)
\end{equation}
Then, the WL-test iteratively refines these colors based on the following observation: Two nodes $u$ and $v$ can only be isomorphic to one another, if they have the same number of neighbors of each color. This is of course rooted in the fact, that any isomorphism $\phi$ that maps $u$ onto $v$, needs to map each neighbor of $u$ onto exactly one neighbor of $v$ and that nodes of different colors cannot be mapped onto each other. Thus, at each step the WL-test assigns to each node $v$ a new color based on the colors of its neighbors:
\begin{equation}
c(v)= h(c(u_1), c(u_2),..., c(u_n))
\end{equation}
where $N(v)= \{u_1, ..., u_n\}$ and $h$ is an injective hash function that assigns each configuration of neighborhood colors to a new, distinct color and is independent of the order of its inputs. This is repeated until convergence. It is straightforward to show by induction that at each timestep isomorphic nodes have the same color.\\
Thus, at the end of the WL test, the signature of a graph $G$ is the unordered list of colors of its nodes	.
\begin{equation}
s(G)=[c(v_1), c(v_2),..., c(v_n) ]
\end{equation}
It is easy to see that graphs can only be isomorphic if their signatures match. For an isomorphism to exist between two graphs $G_1$ and $G_2$, each node in $G_1$ needs to be mapped to a node in $G_2$. As for every isomorphism $\phi$ and all nodes $v$ it holds that $c(v)=c(\phi(v))$, it follows that each color has to occur in both graphs the same number of times and thus signatures have to be equivalent. For an example run of the Weisfeiler Lehman algorithm, see Figure 4.1.\\
Note that there exist graphs that yield the same WL-signature yet are not isomorphic. As a very simple example, consider the graphs shown in Figure \ref{WL-i}. Here, one can see that the WL-test is limited by its purely local nature, failing to differentiate cycles of three and cycles of six nodes respectively. 
\begin{figure}[htbp]
\centerline{\includegraphics[scale=.5]{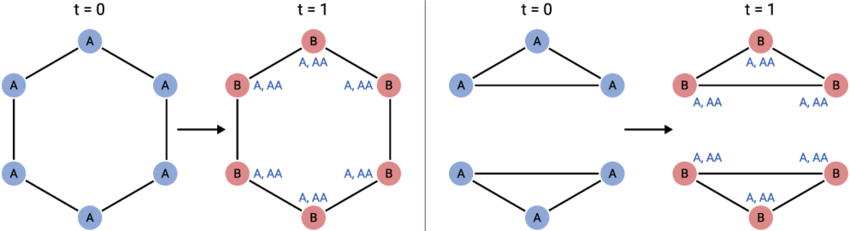}}
\caption{An example of two indistinguishable graphs for the WL-test. Note that both graphs are cyclic in nature but that due to the local nature the WL test cannot distinguish between a circle of length three and one of length 6. Example taken from \cite{damke2020novel}.}.
\label{WL-i}
\end{figure}
\mybox{Weisfeiler Lehman test}{The Weisfeiler Lehman test computes a signature $sig(G)$ for a graph $G$ that can be used to, approximatively, decide graph isomorphism. If two graphs have the same signature, they might be isomorphic. If they do not, they are not isomorphic. The signature is computed in an iterative process where each node is assigned a color at each step. In the first step, all colors are identical. Afterwards, colors are assigned such that nodes with differently colored neighborhoods are separated by their color. Colors are only used to characterize subsets of nodes and are otherwise arbitrary.
}
\section{Deriving GNNs from the WL test}
Consider the following setting: One is tasked with the estimation of some function $f^*$ from the space of graphs to a target space $C$ and desires to build a new neural network architecture to realize this. In this example we consider graphs with labelled vertices such that a graph is given as $G=(V,E,l)$ where $l: V \rightarrow \mathbb{R}^n$ is a function assigning labels to the vertices. The graphs are given as adjacency matrices and as a list of labels. We assume the function $f^*$ to be continuous with respect to the vertex labels. More specifically, we assume that for graphs $G_1 =(V, E, l_1)$ and $G_2=(V, E,l_2)$ that have the same structure with respect to nodes and edges, it holds that\footnote{Note that the notion of similarity used here is the euclidean notion of similarity. This is important as it is one of the core assumptions of gradient based optimization.}
\begin{equation}
l_1 \approx l_2 \implies f^*(x) \approx f^*(y)
\end{equation}
This is a justified assumption in many practical contexts and one that is, in some way or form, implicitly assumed in many deep learning contexts. For example, in a triangle mesh represented as a graph, the labels could indicate the geometric position of each vertex. It stands to reason that if two triangle meshes possess the same connectivity  and their vertex positions only slightly differ, they represent almost the same object and thus the prediction should change very little.\\
Also, as previously noted, we assume $f^*$ to be invariant under graph isomorphisms which is the case for almost all practical applications.\\ Of course, one could simply employ a fully connected neural network that takes as input the labels and the adjacency matrix. Doing so would however not make use of the many assumptions that were presented earlier and in the spirit of geometric deep learning we desire to create a neural network architecture that fulfills these assumptions by construction.\\
Using the domain specific neural network desiderata presented in Chapter 3 and substituting the assumptions that we have on our estimation problem yields these goals for graph neural networks:\footnote{Note that the ideas of task separation and locality are absent here as they are not strictly part of the assumptions of $f^*$ that we presented earlier. If one deals with a function $f^*$ that is assumed to fulfill these properties, they should be included here. However, it stands to note that in graph neural networks one attains locality "for free" and thus this is of little interest here.	}
\begin{enumerate}
\item \textbf{Trainability}. Our network $f$ needs to be a tarinable function (i.e. both parameterized and differentiable with respect to its parameters $\Theta$ to allow for gradient based optimization). 
\item \textbf{$\grp$-Invariance}. As we assume $f^*$ to be invariant under graph isomorphisms, we want $f$ to also be invariant under graph isomorphisms to reduce the space of approximable functions. Specifically, this means that $f(x)=f(y)$ if there exists an isomorphism between graphs $x$ and $y$.
\item \textbf{Continuity and Tunability}. We want to enforce that $f(x)\approx f(y)$ if $x$ and $y$ differ only in their labels and their labels are similar in a euclidean way as this property holds for $f^*$ by assumption. Further, to allow for task specific fine-tuning, we want the degree to which this is true (i.e., the lipschitzness of the function) to be easily tunable.
\item \textbf{Completeness}. Every function $f^*$ that is continuous w.r.t. node labels and invariant to graph isomorphisms should be able to be approximated by our neural network. As graph isomorphisms are hard to decide, it is unreasonable to demnad that this property is entirely fulfilled. Therefore, we demand that as many continuous, graph-isomorphism invariant functions $f^*$ as possible can be approximated.
\end{enumerate}
We will now derive from these targets the GNN architecture.\\
Note that in the realm of deep learning all but the second target are generally fulfilled by all successful deep learning models. It seems thus plausible that most of these targets can thus be met by adapting some tools in the standard deep learning toolbox. Thus, we will start off trying to solve target (2) first and from that point add onto that result to solve the other targets as well.\\ 
The core idea to building a function that is invariant under graph isomorphisms is to use the ideas of the Weisfeiler-Lehman test. As covered in the previous section, the WL-test computes a signature $sig(x)$ for each graph that is guaranteed to be the same for two graphs $x$ and $y$ if they are isomorphic. Thus, if the classification function is dependent only on a graphs signature, it is automatically invariant to isomorphisms. Thus, letting $\psi$ be some arbitrary classification function, we can define $\ft$ as follows and be sure that it is invariant to graph isomorphisms:
\begin{equation}
\ft(G)= (\psi \circ sig)(G)
\end{equation}
This immediately solves the problem of constructing a neural network architecture that is invariant to graph isomorphisms\footnote{Note that if a neural network is given as input the WL-signature of a graph it can distinguish only graphs that the WL-test can distinguish and is thus invariant to a bit more functions than just isomorphisms. However, in most practical applications this appears to be of little concern.}. If one uses a fully connected neural network to process the graphs signature the resulting function is also differentiable and parameterized. However, one of our targets is not yet met. The signature computed by the WL test consists of arbitrary numbers. Recall that the signatures are computed by repeatedly assigning hash values to each vertex depending on the multiset of signatures of its surrounding nodes but that the concrete hash function was entirely unrestricted. As a consequence, the signature of a graph can vary wildly even if only small changes are done to its labels.\\ Thus, we lack the desired continuity with regards to small changes in the vertex labels and cannot yet fulfill target (3).\\
To achieve target (3), it is important to have the colors of the WL-test depend in a continuous way on the original vertex labels. To enforce this, let us revisit how the graph colors were originally computed. Recall the update equation of the WL-test which was given by:
\begin{equation}
\label{upd2}
c(v)= h(c(u_1), c(u_2),..., c(u_n))
\end{equation}
The color of a node is updated depending on the colors of its neighbors with some function $h$ determining the concrete color that is used to represent the specific configuration of neighboring colors. For the WL-test to work, $h$ had to be just an arbitrary, injective function that was independent of the order of its arguments. For the purposes of our estimation, we also require $h$ to be continuous with respect to its inputs. Recall that the iterative process of the WL-test starts with colors being equivalent to the original labels and that colors are updated repeatedly using $h$. If $h$ was to be continuous, the final labels would also depend on the labels $l$ in a continuous way as the final signature is constructed by repeated applications of $h$ and  compositions of continuous functions are still continuous.\\
Moreover, for estimation purposes we can drop the requirement of $h$ being injective. While important for the task of distinguishing non-isomorphic graphs, there might be graphs in our use case that are not isomorphic yet still equivalent with respect to $f^*$ and thus this requirement is not necessary. It is only necessary for $h$ to be injective, if the concrete estimation task demands it.\\
In recap, to achieve a function $\ft$ that is parameterized, differentiable with respect to its parameters, invariant to graph isomorphisms and continuous with respect to the node labels, we can implement $\ft$ as follows:\\
Given an input graph $G$,  compute the WL-signature $sig_h(G)$ using a continuous update function $h$. Then, use the WL-signature as basis for final classification and compute $\ft(G)= \psi(sig_h(G)$ where $\psi$ is some arbitrary function that is continuous with respect to its arguments and differentiable with respect to its inputs.\\
This structure guarantees that we fulfill all of our targets. However, we still have to adequately choose $h$ and $\psi$ to actually attain a concrete realization of this structure.
\section*{Building h and $\psi$}
Recap the restrictions that were imposed on $h$. It is a multivariate function\footnote{Note that we do not even know the concrete input dimension of $h$ as it depends on the neighoring colors of a vertex. For different vertices, the neighborhood has a different size and we therefore need $h$ to be variable in its input dimension.}, depending on all vertex colors in a neighborhood, that is continuous with respect to its inputs and commutative\footnote{We will use the term "commutative function" to refer to functions that are not dependent on the order of their input.}. Otherwise, $h$ is basically unrestricted. We do not entirely know, which $h$ from the set of functions induced by these constraints is the optimal one. Thus, it seems necessary to have $h$ be a trainable function as well. Fortunately, we can already introduce a neural network architecture that is well suited for exactly this task in deep sets. Deep sets, as presented in Chapter 3, are specifically designed to approximate arbitrary functions that are invariant to permutations of their input.\\
Thus, according to the deep set architecture, we define
\begin{equation}
h(v) = \Phi_{update}\left(\sum_{u \in N(v)}\phi_{encode}( c(u)  )\right)
\end{equation}
as the update function to use to compute the WL-signature. Here, $\Phi_{update}$ and $\phi_{encode}$ are implemented as fully connected neural networks.\\
Intuitively, $h(v)$ for some vertex $v$ is then computed as follows: For each neighboring vertex $u$, compute $\phi_{encode}(c(u))$, encoding the information present in $c(u)$, and sum their results. Afterwards, apply $\Phi$ to their results to update the state of $h$. This directly yields the message passing analogy frequently used to describe graph neural networks\footnote{While message passing graph neural networks are merely one flavor of GNNs, they appear here as they are the most general kind of GNN \cite{geo}. In our derivation, we did not place any domain specific constraint on the update function $h$. If one does place restrictions on $h$, one ends up with a different version of GNNs such as attention based GNNs or graph convolutional neural networks \cite{kipf2016semi, thekumparampil2018attention}}\cite{gilmer2017neural,geo}.\\
There are two ways of interpreting this concrete implementation of $h$. The first one is the intuitive way of looking at $h$. If $h$ is implemented like this, one can visualize this process as vertices iteratively sending messages to each other that are encoded by the encoding fully connected network $\phi_{encode}$, then aggregated by sum and lastly used to create a new vertex color through $\Phi_{update}$.\\ The second way of thinking about $h$ constructed in this way is that $h$ is constructed such that it can approximate arbitrary, commutative multivariate functions. As proved by \cite{sannai2019universal}, if $\Phi_{update}$ and $\phi_{encode}$ are powerful enough (read: enough layers and neurons), $h$ can approximate arbitrary commutative, continuous functions to arbitrary precision.\\ On a final note, this section relatively quietly ignored the issue that the dimensionality of $h$ is variable as $h$ takes as input all colors in a neighborhood of some vertex. As vertices differ in their degree, it is previously unknown what the dimension of this function is. This is not generally a problem as $h$ uses a sum to aggregate incoming information anyway which works with arbitrary numbers of input. However, if we desire $h$ to approximate all functions with $n$ inputs, Equation 3.17 dictates that the last nonlinear layer of the network realizing $\Phi$ needs to posseess $2n$ neurons. If vertices with a larger outdegree than $n$ occur in our problems, $h$ can no longer approximate arbitrary functions. This does not seem to be a problem however as most problem domains do not contain vertices with extremely large degrees and neural networks are routinely built with very large amounts of neurons.
Recall that to build our full function $\ft$ we split the function into two parts:
\begin{equation}
\ft(G) = \psi(sig_h(G))
\end{equation}
The first one computing the WL-signature of graph $G$ using a learned, continuous hash function $h$. The second, $\psi$, performs the actual classification of the graph based on the previously computed signature. Luckily, constructing $\psi$ is relatively easy using the methods used for creating $h$. Similarly to $h$, we do not have any kind of prior restriction to impose onto $\psi$ except for the fact that it has to be a continuous, multivariate, trainable, commutative\footnote{This is rooted in the fact that the WL-signatures are inherently permutation invariant.} function where the exact dimension is not necessarily known. Thus, we use the same structure as for $h$ to construct $\psi$ to ensure that every multivariate function can be constructed.\\
Following this construction, $\psi$ is given by:
\begin{equation}
\psi(G)= \Phi_{final}( \sum_{v \in V} \phi_{vote}(c(u))
\end{equation} 
For each node, a final color has been computed as a result of $\sig(G)$. The voting network $\phi_{vote}$ calculates each nodes contribution to the final prediction. Those are aggregated by sum and used to form the final prediction by $\Phi_{final}$, both $\phi_{vote}$ and $\Phi_{final}$ being implemented as simple fully connected neural networks.\\
Again, keep in mind that this implementation basically means that, if enough layers/nodes are provided, $\Phi_{final}$ and $\phi_{vote}$ can approximate arbitrary commutative functions.\\
Note that, as $\psi$ can approximate arbitrary commutative functions, the universal approximation theorem can be attained almost for free: Every function $f^*$ fulfilling our assumptions can be arbitrarily closely approxmitated if one adds the condition that:
\begin{equation}
sig(x)=sig(y) \implies f^*(x)=f^*(y)
\end{equation}
for arbitrary graphs $x$ and $y$\footnote{$sig$ here refers to the signature computed by the actual WL-test, not the signature computed by a GNN with a learned hash function $h$.}. This follows as $h$ can be injective, given the correct parameters. If $h$ is injective, two graph signatures of the GNN are equivalent if and only if the signatures of the same graps computed by the WL-test are equivalent.\\
Lastly, $\psi$ can approximate arbitrary functions from the space of graph signatures to the space of possible classes and thus the universal approximation property follows.\\
As for training of a GNN, observe that the GNN consists only of concatenations of trainable functions. Thus, the GNN can be trained with all gradient based optimization methods.
\section*{GNNs Summary}
In this section, we introduced graph neural networks (GNNs) as domain specific neural networks for the domain of graphs that adhere to the domain specific neural network desiderata that were laid out in Chapter 3. Most importantly, graph neural networks are invariant to graph isomorphisms, which is required by the domain specific neural network desiderata.\\
To achieve this, GNNs are built as a variant of the WL-Algorithm where the color update function is implemented as a deep set (a permutation invariant neural network) and the final set of colors is given to another deep set for classification purposes. As GNNs implement the WL-Algorithm, they possess many of its properties. Most importantly, for a GNN $f$ it holds that $f(x)=f(y)$ whenever the WL-test cannot distinguish $x$ and $y$. As a consequence, GNNs are invariant to graph isomorphisms. However, there are also some non-isomorphic graphs that GNNs cannot distinguish. Thus, they do not fulfill the completeness property.\\
Still, they fulfill all of the other domain specific neural network desiderata and for many classes of graphs, the WL-test is indeed powerful enough to separate them.\\
This section showed how the domain specific neural network desiderata laid out in Chapter 3 are reflected in graph neural networks. The core insight of this section is twofold: First, GNNs do adhere to the desiderata laid out in Chapter 3. Second, the way in which they achieve it is by adapting an already known and understood algorithm from the domain of graphs by incorporating neural networks wherever some function can be chosen. This idea is also found in the derivation of Deep Sets \cite{zaheer2017deep}.\\
In the next section, we will see if the geometric deep learning blueprint can provide a shortcut here and replace the domain knowledge as a tool for neural network construction.
\mybox{Graph Neural Networks}{Graph Neural Networks are domain specific neural networks for the domain of graphs. Graph neural networks are functions $\text{Graphs} \rightarrow C$ that take as input graphs $G=(V,E)$ represented as a matrix of node labels $\hat{V} \in \real^{d \times |V|}$ and an adjacency matrix $\hat{E} \in \mathbb{B}^{|V| \times |V|}$. Graph neural networks compute a WL-signature of the graph using a deep set as a color update function $h$. This signature is passed to another deep set to achieve the final classification result. For all graph neural networks $f$ it holds that $f(x)=f(y)$ for all graphs $x,y$ that cannot be separated by the WL-test. As a consequence, graph neural networks are invariant to graph isomorphisms.}
\section{GNNs and the GDL Blueprint}
In this section, we will describe how GNNs fit the geometric deep learning blueprint and how GNN specific ideas can be found in their design beyond the GDL desiderata that were implicitly used in their derivation preceeding this section. To do this, first recall the structure of the geometric deep learning blueprint. As defined in Chapter 3, the geometric deep learning blueprint prescribes estimators $f$ to be constructed such that
\begin{equation}
f=\Phi \circ \phi_n \circ \phi_{n-1} .... \circ  \phi_1
\end{equation}
where the $\phi_i$ are trainable, lipschitz continuous functions that are equivariant to the actions of $\grp$:
\begin{equation}
\phi_i(gx)=g\phi_i(x)
\end{equation}
and $\Phi$ is trainable, lipschitz continuous and also invariant to actions of $\grp$
\begin{equation}
\Phi(gx)=\Phi(x)
\end{equation}
Recall that the actions $\grp$ that we are concerned with in the context of graph neural networks consist of the graph isomorphisms, i.e. permutations of the nodeset $V$ and an according change in the edges.\\
We will start with some function $\phi$\footnote{For brevity, we will omit the subscript but note that this applies to all functions $\phi_i$ in the blueprint.} where the condition $\phi(gx)=g\phi(x)$ needs to be fulfilled. First off, as for all $\grp$-equivariant functions, to be able to apply $g$ to $\phi(x)$ one requires that $\phi(x)$ belongs to the same type of object as $\phi(x)$. Therefore, $\phi(x)$ needs to be a function mapping labelled graphs to labelled graphs in a way that is orthogonal to permutations of the node order. In the case of GNNs, each step of the modified WL-algorithm implemented by GNNs constitutes such a function $\phi$. First, it is easy to check that a WL step maps one labeled graph to another as it only computes new labels, more specifically colors, for the graph. Further, we can observe that each WL step indeed fulfills the equivariance condition by observing the impact of a WL step, or message passing step, on a graph $G=(V,E,c)$ and one isomorphic graph $G'=(V',E',c')$ that originated from $G$ by some isomorphism/permutation $\pi$.\\
Recall that  each WL step computes a new graph $G_{new}=h(G)=(V,E,c_{new})$ by computing new colors for $G$ according to this function:
\begin{equation}
c_{new}(v) = \Phi_{update}\left(\sum_{u \in N(v)}\phi_{encode}( c(u)  )\right)
\end{equation}
for a node $v \in V$ with neighborhood $N(v)=\{u_1, ..., u_n\}$ with $\Phi_{update}$ and $\phi_{encode}$ being implemented as fully connected neural networks. Now consider the same update step for the isomorphic graph $G'$ and specifically the node $v'=\pi(v)$ that corresponds to $v$. The neighborhood of $v'$ is, as $\pi$ is an isomorphism, given by $N(v')=\{u'_1=\pi(u_1), ..., u'_n=\pi(u_n)\}$. As $G$ and $G'$ are isomorphic, it necessarily holds that
\begin{equation}
c(u)=c'(\pi(u))
\end{equation}
for all $u\in V$. Therefore, we can immediately see that
\begin{align*}
c_{new}(v) &= \Phi_{update}\left(\sum_{u \in N(v)}\phi_{encode}( c(u)  )\right)\\
&= \Phi_{update}\left(\sum_{u' \in N(\pi(v))}\phi_{encode}( c'(u')  )\right) \quad\text{(as $\pi$ is an isomorphism)} \\
&= c_{new}(\pi(v))
\end{align*}
As a consequence, one can immediately see that $\pi$ is still an isomorphism between the graphs $G_{new}$, $G'_{new}$ after this message passing step. Therefore, the isomorphism between them is preserved by this message passing step $h$ and 
\begin{equation}
\pi(h(G))=h(\pi(G))
\end{equation}
Therefore, the update function $h$ fulfills the equivariance property.\\
Finally, after computing the node colors via a trained WL-algorithm, a graph neural network computes its final result as by 
\begin{equation}
\psi(G)= \Phi_{final}( \sum_{v \in V} \phi_{vote}(c(u))
\end{equation} 
This formulation is of course invariant to graph isomorphisms. The reason for this lies in the fact that $\phi_{vote}$ is applied to each node separately, independent of graph structure, and that all of the node embeddings are aggregated by a sum. If one were to permute the order of the nodes, the aggregation via sum ensures that the final result does not change. Therefore, this final classification step takes the place of the invariant final function in the geometric deep learning blueprint and it follows that graph neural networks follow the geometric deep learning blueprint, which was also lined out in \cite{geo}.\\
\section{Can GNNs be Derived from the GDL Blueprint?}
As was just shown, GNNs adhere to the geometric deep learning blueprint. However, the most interesting question is not if graph neural networks adhere to the geometric deep learning blueprint as the geometric deep learning blueprint was built \textit{from} the ideas of graph neural networks. Rather, as we are interested in the idea of applying geometric deep learning to new areas, the question is if graph neural networks can be derived from this blueprint in a feasible manner. This question is of course hard to answer in hindsight as graph neural networks already exist and their ideas are known. Further, whether one can derive graph neural networks from the geometric deep learning blue print is, as said blueprint leaves non-trivial work to its user, heavily subjective and depends on the skillset and ideas of the user. Therefore, this section is more of an informal argument and cannot in itself be fully convincing. Nevertheless, we will try to evaluate how much work the geometric deep learning blueprint leaves to its user and how helpful it is in the hypothetical of GDL-blueprint guided design of graph neural networks.\\
Recall that the geometric deep learning blueprint requires three different functions:
\begin{enumerate}
\item The feature representation function $\phi$ : $\Omega \rightarrow \Omega$. This function should be trainable, lipschitz continuous, $\grp$-equivariant, i.e. $f(gx)=gf(x)$ and able to  be lossless, i.e. there exists some set of parameters $\theta$ such that $\phi_{\theta}$ is bijective.
\item The invariance ensuring aggregation function $\Sigma$ : $\Omega \rightarrow \real^k$ , which is a fixed $\grp$-invariant function and usually implemented as a simple sum or a similar aggregation function.
\item The classification function $\Phi$ : $\real^k \rightarrow C$ that does the final classification. This function should be trainable and lipschitz continuous. 
\end{enumerate}
We will first start off with the easiest ones to define. The function $\Sigma$ that ensures $\grp$-invariance is, relatively simply, a sum over all of the node embeddings. Using a sum in this way radically discards every notion of graph structure and is therefore invariant to isomorphisms (on top of many other transformations). Similarly, the function $\Phi$ is constrained only to being trainable and lipschitz continuous and can therefore be realized as a simple fully connected neural network.\\
Both of these choices match those made in graph neural networks and are, as we would argue sensible. The idea of using a sum to ensure invariance to permutations of some set of objects is found in deep sets and implicitly in every domain where mean pooling is used. Similarly, as we established earlier on in this work, whenever one wants to use a lipschitz continuous, trainable function for estimation purposes, fully connected neural networks are the standard method of choice in deep learning. Therefore, we believe that, if one used the geometric deep learning blueprint to come up with a neural network architecture fit for graphs, one could feasibly come up with those functions $\Sigma$ and $\Phi$ that are used in graph neural networks.\\
However, in the case of the equivariant functions $\phi$, this is much harder as there are many functions that fulfill the constraint of $f(gx)=gf(x)$. For example, one could implement $\phi$ by applying a neural network to each of the node labels 
separately as is done in the case of deep sets. This respects $\grp$-equivariance and all other restrictions on $\phi$, but can hardly be satisfying as it ignores edges entirely and thus makes it impossible for the entire construct to be a universal approximator for the desired domain. Similarly, one could achieve $\grp$-invariance by implementing message passing along those nodes that are \textit{not} connected, which would be unsatisfying in its own right. Further, even if one did not choose any of those unsatisfying ideas, there does not appear to be a structural way of identifying message passing as it is generally implemented in graph neural networks as the way to go. In itself, it is even possible that one would have the idea of implementing $\phi$ in such a way that it changes up the edge structure of the input graph, yielding an infinite amount of feasible, but probably useless versions of $\phi$.\\
This is hugely problematic as the feature representation functions $\phi$ make up the, by far, largest part of the neural network architecture in almost all instantiations of domain specific neural networks in practice (compare to CNN architectures in \cite{cnn,dl}). Choosing the wrong function here can lead to a critical loss in neural network performance, arguably a much greater one than if one made a mistake in the other two functions, $\Sigma$ and $\Phi$. In fact, at least in the context of CNNs, useful architectures exist that make no use of the latter functions and still yield decent results \cite{ronneberger2015u}, indicating that the heavy lifting is done by the feature representation function $\phi$ which corresponds to the WL-steps/message passing in graph neural networks.\\
Thus, we hold that the geometric deep learning blueprint lacks specificity in exactly the most critical area, i.e. the first couple of layers that compute an embedding for the input that can later on be used by much more simple classification functions. In fact, it contributes almost nothing in this area that is not covered by conventional deep learning intuition \cite{dl}. It does not seem feasible to expect that this blueprint can sufficiently aid domain specific architecture creation substantially beyond the traditional approaches and we believe that it marks only a first idea of what an actual architecture blueprint should look like instead of an approach that is actually useful in practice.

%% file: kapitel/invariantswhy.tex
\chapter{On the Impact of Enforced $\grp$-Invariance on Estimator Performance}
In the last chapter, we used graph neural networks as a case study to see how the geometric deep learning based ideas on domain specific neural network design are reflected in a state of the art neural network architecture. However, this only shows that these ideas are used in practice and it does not sufficiently show that the ideas of geometric deep learning are necessary for a good domain specific neural network architecture. In short, in the last chapter we saw that it is possible to attain good neural network architectures using the ideas of geometric deep learning but we could not show that those ideas are responsible for the success of the architecture.\\
In this chapter, we will try to answer just this question, what is the impact of the ideas of geometric deep learning. Recall that the core ideas of geometric deep learning for domain specific neural networks are task separation, locality and $\grp$-invariance. Covering all three would exceed the scope of this work and as task separation and locality are more difficult to properly capture, we will focus on $\grp$-invariance as our property of interest. Therefore, this chapter will try to answer the question: ``What is the impact of enforcing $\grp$-invariance in an estimator, given that $f^*$ is $\grp$-invariant?''.\\
First of all, note  that giving theoretical guarantees (and thus evaluating the impact of $\grp$-invariance) on a large class of neural networks is generally a very hard problem.
As noted by C. Zhang in two very influential papers in deep learning \cite{zhang2021understanding,zhang2016understanding}, deep learning models  are simply put way too powerful with regards to their expressivity for many standard analyses. Standard neural networks that are heavily regularized can trivially fit even the most absurd training data as shown by \cite{zhang2021understanding}, generally attaining a training error of almost 0 on random tasks without any structure\footnote{The experiments conducted by \cite{zhang2016understanding} very directly concern this work. In their work, they employ data augmentation, a less powerful version of incorporating $\grp$-invariance than the one used in this work and show that it does not sufficiently regularize a neural network. Instead, they show that these networks that were trained to be $\grp$-invariant can still attain a training error of almost 0 on non $\grp$-invariant tasks}.\\
To measure the success of neural networks, we can look at two things: First, the training error, i.e. is this neural network able to correctly fit the training data? As  just mentioned, this is almost a given for modern neural networks as correctly fitting the training data is trivial for modern neural networks that benefit from having extremely large amounts of layers and neurons \cite{zhang2016understanding}. Therefore, this question is not interesting in practice. Second, the generalization error, i.e, given that we achieved a low training error, how sure can we be that the performance is also good for the datapoints that we did not observe. This question is much more interesting for practical application, but also much harder to answer.\\
Traditional statistical wisdom generally assumes that less powerful estimators, i.e. those, that can fit only a limited number of functions, generalize better. The idea is,  if the function $f^*$ does not belong to the very limited number of functions that my estimator can approximate, then it is very unlikely that we drew random samples from $f^*$ that our estimator could properly fit. Therefore, if we attain a low training error, we can be reasonably sure that the true error is not too far off. An example of this philosophy are the widely used bounds using the Vapnik-Chervonenkis dimension  \cite{vapnik2013nature}.\\ As noted in \cite{zhang2021understanding}, these results do not extend well to modern neural networks as neural networks are \textit{not} limited with regards to what they can approximate. Therefore, traditional statistical bounds on 
neural network generalization do not work.\\
Still, neural networks tend to generalize well in practice, especially in those domains where they are frequently employed. Therefore, we have to use novel techniques as was demanded in \cite{zhang2021understanding}. Behind most of these, there is the idea of incorporating specific properties of the domain. If we assume that our function $f^*$ has some properties, then we can give bounds on the generalization performance and a better description of how $\grp$-invariance aids in learning.\\
This chapter will attempt to give some insights and quantify as best as possible how $\grp$-invariance shapes the learning process. We will especially focus on techniques that were outlined as possible answers to the analysis of neural networks in \cite{zhang2021understanding}, namely Bayesian PAC Bounds and the biases induced by gradient based optimization.\\
This chapter will first formalize the effects of the restriction to $\grp$-invariant functions and cover the effects of $\grp$-invariance on neural networks from three different angles, one focussing on gradient based optimization and the biases induced through this training method, the topological characterization of the classification space and lastly bayesian PAC analysis of the generalization properties.
\section{Characterizing the $\grp$-Invariant Estimation Problem}
\label{character}
First off, we have to understand how $\grp$-invariance changes the estimation problem. For all of the remaining chapter, we will assume that we have some function $f^*$ to approximate that is in fact invariant to some group of transformations $\grp$.\\
The original, unrestrained estimation problem that we were tasked with was posed as:
\begin{equation}
 \hat{f} = \argmin_{f \in F}   \int_{x \in \Omega }  l(t(x),f(x)) d\mu(x) 
\end{equation}
where the integral notably ranges over the full space of possible inputs $\Omega$, usually the vector space $R^n$, $t : \Omega \rightarrow C$ denotes the ground truth label for $x$, $l$ is some loss function, $\mu$ some distribution on $\Omega$.  By introducing the condition on $\grp$-Invariance as
\begin{equation}
\forall x \in \Omega \forall g \in \grp f(x) = f(gx) 
\end{equation}
one only has to estimate a function $\orbit(\Omega)\rightarrow C$ where $\orbit(\Omega) = \{ \orbit(x) \mid x \in \Omega \}$ the domain being simplified from $\Omega$ to the $\grp$-induced partition represented by the orbits\footnote{For more on groups and orbits, consult the appendix.}.\\
This can correspond to a drastic simplification of the problem. For example, if we consider the group of invariances of shifts by three acting on $\mathbb{R}$, i.e. $f(x)=f(x+3)$ for all $x$, the orbits are given as 
\begin{equation}
\orbit(x)=\{... x-3, x, x+3,...\}
\end{equation}
It can easily be seen that this set of orbits is isomorphic to the space $(0,3) \subset \mathbb{R}$ which is not in itself a vector space. While estimation problems $(0,3) \rightarrow \real$ can certainly be difficult, they should be easier than a corresponding estimation problem $\real \rightarrow \real$. How this manifests in the case of neural networks, we will see in the next sections.
\mybox{The $\grp$-invariant Estimation Problem}{
	Introducing $\grp$-invariance simplifies the estimation problem. Instead of estimating a function $\Omega \rightarrow C$, $\grp$-invariance allows us to approximate a function $\orbit(\Omega)\rightarrow C$ where $\orbit(\Omega)$ is the set of $\grp$-induced orbits. In our example of a function that was invariant to addition of multiples of 3, this led to a drastic reduction in complexity. 
    }
\section{Invariance and Gradient Descent}
\label{graddesc}
As neural networks are first and foremost machine learning models, they are frequently viewed from a statistical point of view as the end result of a stochastic sampling and training process \cite{dl}. As an example, neural networks are frequently treated as maximum likelihood estimators as described in \cite{dl}. This point of view is very generic as it abstracts from the concrete optimization process. This is useful to justify loss functions and the concrete idea of optimizing said loss function, but neglects the fact that neural networks are usually not entirely optimized, i.e. the global minimum of the loss function is not usually found, and that there exist more than one local minimum the optimizer could converge to. Thus, the concrete behavior of the optimization function is of interest to see what local minimum is reached\\
As pointed out by \cite{zhang2021understanding}, understanding the process of gradient based optimization is one possible avenue to explain how they learn and how they generalize. While standard statistical bounds tend to not work well with neural networks, optimization processes are not in fact random and it might be biases induced by the training process that explain why neural networks work well.
 
In all kinds of gradient based optimizers such as gradient descent \cite{dl} or ADAM \cite{kingma2014adam}, updates take the form
\begin{equation}
\Theta_{new} \leftarrow \Theta_{old} + \alpha\nabla_{\Theta} l_x(\Theta)
\end{equation}
Where $x$ is a given sample. For infinitesimal learning rates $\alpha$ this can be easily shown to always improve the error. \\
Instead of inspecting just the error, we will take a look at how the function value itself changes under gradient based updates as we are interested in the learning process and how it is influenced by $\grp$-invariance of the network. A taylor series expansion and substituting the update given above yields: 
\begin{align}
 f_x(\Theta_{new})  &\approx f_x(\Theta_{old})+\nabla_{\Theta} f_x(\Theta)\alpha*\nabla_{\Theta} l_x(\Theta) &&\\  &=
f_x(\Theta_{old})+\alpha \nabla_{\Theta} f_x(\Theta) \ddx{l}{f} \nabla_{\Theta}  f_x(\Theta) &&\\ &=  
 f_x(\Theta_{old})+\alpha\ddx{l}{f}\dist{\nabla_{\Theta} l_x(\Theta)}^2 && \\
\end{align}

This can be further extended to points in the direct neighborhood of $x$, as gradients are continuous and thus similar in a (very close) neighborhood of $x$.\\
\begin{equation}
\label{gradneighbor}
 f_y(\Theta_{new}) \approx f_y(\Theta_{old})+\nabla_{\Theta} f_y(\Theta) \alpha\ddx{l}{f}\nabla_{\Theta} f_y(\Theta) \approx f_x(\Theta_{old})+\alpha\ddx{l}{f}\dist{\nabla_{\Theta} l_x(\Theta)}^2
\end{equation}
This yields a very intuitive description of gradient based optimization algorithms that will be useful to inspect the effect of $\grp$-invariance on the learning process. Fundamentally, if given a sample (x,$label$) the function value is changed proportionally to $\alpha\dist{\nabla_{\Theta} l_x(\Theta)}^2$ and the term $\ddx{l}{f}$, which implicitly depends on $label$, denotes the direction of the change. This makes intuitive sense as the term $\alpha\dist{\nabla_{\Theta} l_x(\Theta)}^2$ consists of the squared length of the gradient at point $x$, measuring how sensitive $f$ is to change in $\Theta$ and the learning rate and thus describes how the value of $f$ changes at point $x$ when $\Theta$ is changed. In contrast, the term $\ddx{l}{f}$ describes how the loss changes when $f$ changes and thus prescribes the desired direction of change in $f$. As Equation \ref{gradneighbor} shows, for points $y$ where the gradient $\nabla_{\Theta} f(x)$ is similiar in direction to the gradient at $x$, a similar change is done. For all points with an opposite direction of the gradient, the inverse is done. As previously noted, this most notably applies to points in the very direct euclidean neighborhood of the point $x$, as gradients are continuous. For all other points however, it is in a general setting difficult to obtain anything about their respective gradients $\nabla_{\Theta}$ and thus hard to predict, how the optimization will work.\\
This point directly illustrates, why standard neural network architectures do not generalize well to non-euclidean data. The optimization process assumes points to be similar, if they are close to each other in a euclidean sense. Obviously, this does not make a lot of sense for non-euclidean data; points that are close in a euclidean metric can be very dissimilar and points that are not close in a euclidean metric can be almost the same datapoint. As an example, consider images shifted by one single pixel. Obviously, an image shifted in such a fashion is still very similar to the original image, but the euclidean distance between both of these images can be very large.\\ 
This changes with the added $\grp$-invariance. In fact, as $\grp$-invariance is introduced into the network, the identity $f_{\Theta}(x)=f_{\Theta}(gx)$ holds for all $g$, $x$ and $\Theta$.  As a consequence, we know that $\nabla_{\Theta} f_{\Omega}(x)=\nabla_{\Theta} f_{\Omega}(gx)$. Therefore, the same change that happens to $f$ at point $x$ also happens at each shifted point $gx$ for all $g\in \grp$. Further, Equation \ref{gradneighbor} yields that the direct neighborhood of $gx$ also experiences a similar change. This shows, that the invariance adds additional structure to the learning process. We can interpret this as introducing a new notion of neighborhood to the learning process:
\begin{equation}
\label{dist}
dist(x,y) = \min_g \dist{y-gx}
\end{equation} 
instead of the standard euclidean distance that gradient based optimization obeys in unrestrained settings.\\
This directly relates to one of the core goals of geometric deep learning which is to extend the success of deep learning to non-euclidean domains \cite{geo}, in their respective work to domains such as triangle meshes and graphs. It appears however notable, that the above formulation of distance in Equation \ref{dist} still contains euclidean distance as a factor. The above formulation is very sensible for almost all use cases that were examined in their work, as almost all of them share a graph like structure where there exists a number of objects, encoded by real numbers, in a certain ordered structure. Images are represented as grids of pixels (each represented as some real numbers), graphs consists of nodes (each represented by a feature vector) and triangle meshes consist of vertices, each represented as points in a 3d space. The introduction of invariance in these cases concerns the ordering of these euclidean representations; but the representations themselves are still euclidean in nature. For example, consider the case of rotating images and Equation \ref{dist}. This notion of similarity denotes images as similar, if they are similar in a euclidean sense to rotated versions of each other.\\
This is a very sensible notion of neighborhood for this use case and this generally extends to all of the use cases presented in \cite{geo}, but it does not seem admissible to claim that the introduction of invariance allows for completely non-euclidean domains to be covered, for discrete inputs, non-continous inputs etc. the introduction of invariances does not seem sufficient yet. Rather, it lends itself well to a certain kind of domains, namely those, where \ref{dist} represents a sensible notion of distance.
\\
\mybox{Gradient Descent on $\grp$-invariant Estimators}{Assume an infinetesimal learning rate $\alpha$. If given a sample $(x,f^*(x))$, gradient descent changes $f(x)$ be closer to $f^*(x)$. Due to continuity of $f$, a euclidean neighborhood of $x$ experiences a similar change. For $\grp$-invariant networks this changes. Instead of the euclidean neighborhood of $x$ experiencing the same change, the neighborhood is given by:
	\begin{equation}
dist(x,y) = \min_g \dist{y-gx}
	\end{equation}}
\section{Invariance and Topological Structure}
The topological complexity of the function spaces that neural networks can emulate has been the repeated focus of attention in deep learning literature. In their works, \cite{bianchini2014complexity} and  \cite{guss2018characterizing}, the respective authors show upper bounds for the number of topological holes that a neural network can approximate in its preimage.\\
Consider a binary classification problem that a neural network is tasked with. Thus, there exist two classes, $C^+ \subset \Omega$ and $C^- \subset \Omega$ that should be separated by a classifying function.
\begin{figure}[htbp]
\centerline{\includegraphics[scale=.5]{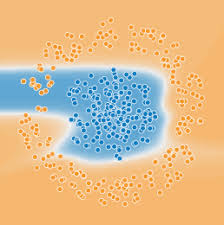}}
\caption{A number of points in a binary classification problem \cite{johnson2018deep}. The blue points belong to class $C^+$, the yellow ones to $C^-$. This space is topologically relatively easy to learn for a neural network, as both the subspace $C^+ \subset \Omega$ and the subset $C^- \subset \Omega$ posess only one topological hole.}
\label{binary}
\end{figure}
As we are interested in binary classification, the task of correctly classifying all $x \in \Omega$ to their respective classes can be easily posed as such:
\begin{equation}
f(x) > 0 \Leftrightarrow x \in C^+ 
\end{equation}
Thus, our neural network classifier should yield a function that is positive in exactly the area $C^+$ and negative in the area of $C^-$. This is, as shown in \cite{bianchini2014complexity} only possible, if the subspaces $C^+$ and $C^-$ possess a limited amount of topological holes. The exact number of topological holes that can still be approximated by our neural network depend on its architecture, activation function and size, but it is necessarily finite and both \cite{bianchini2014complexity} and  \cite{guss2018characterizing} show, that larger, more powerful neural networks and more data points are needed to approximate spaces that have more topological holes.\\
Note that this argument is independent of the optimization algorithm, instead only referring to the expressive power of the neural network architecture, but raises similar issues to those made in Section \ref{graddesc}; general neural networks assume some form of continuity and struggle with problems that lack this property.\\
In the sense of topological structure of the preimage, it is very easy to see that a lack of $\grp$-invariance in a neural network can have devastating effects. For example, consider the function $f^*(x)=x \mod 3 > 1$ that is to be approximated by a neural network $f$. On the surface, this function looks incredibly simple but it is deceptively hard to learn for a neural network. In our experiments, fully connected neural networks with differing activation functions and up to 8 layers failed to properly learn this function in its entirety. 
\begin{figure}[htbp]
\centerline{\includegraphics[scale=.5]{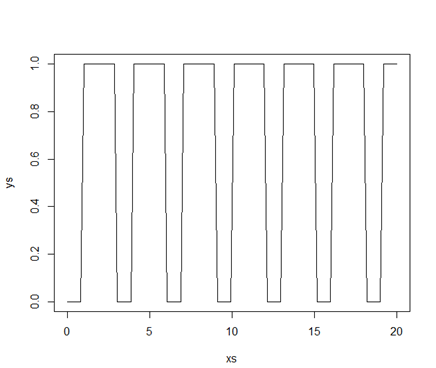}}
\caption{The image of $f^*(x)=x \mod 3 > 1$. Note that there are infinitely many gaps between the areas where $f^*(x)=1$. To more easily graph this function, we use $1$ to represent the boolean value of "true" and $0$ to represent "false".}
\label{mod3}
\end{figure}
This is however unsurprising if one considers the space $C^+ = f^{-1}(1)$  which is given by
\begin{equation}
...(-2,0) \cup (1,3) \cup (4,6) \cup..
\end{equation}
and contains an infinite number of holes. Thus, no matter, how large our neural network is, it can never sufficiently approximate the infinitely large structure of $f^*$. Of course, with added layers and neurons the neural network can approximate more and more of $f^*$, as shown in \cite{bianchini2014complexity}, but it can never actually reach this infinite structure and small networks fall incredibly short of the actual problem. \\
In contrast, when one considers the shift invariance of $f^*$, the problem becomes much, much easier. As noted in Section \ref{character}, after taking $\grp$-invariance into account, the domain of the approximation is reduced from $\Omega$ to $\orbit(\Omega)$. Therefore, instead of approximating $f(x)$ for all $x$, it is sufficient to approximate $f(\orbit(x))$. In this specific case, due to the $\mod 3$ in $f^*$, we know $f^*$ to be invariant under shifts by 3, i.e. $f^*(x) = f^*(x+3)$ for all $x$. Thus, the orbit of $x$ is given by:
\begin{equation}
\orbit(x)=\{... x-3,x , x+3, x+6,...\}
\end{equation}
The set of all orbits is thus isomorphic any open interval of length 3, especially the interval $(0,3)$. Our approximation task thus is equivalent to approximating $f^*$ over the interval $(0,3)$ which contains only one hole (c.f. Figure \ref{mod3}), drastically reducing the complexity of this problem.\\
This is an obviously extreme example and it is difficult to quantify in a general manner, how topological holes behave under $\grp$-invariance as the exact behavior of the network depends on how this invariance is implemented. Of course, one could implement $\grp$-invariance in such a way that yields another highly complex domain. Still, enforcing $\grp$-invariance should  always drastically reduce the topological complexity.\\
\mybox{$\grp$-Invariance and Topological Complexity}{Neural networks $\real^k \rightarrow \real$ induce two topological spaces on their domain $\Omega$ by $C^+=\{x \; \mid \; f(x)>0\}$ and $C^-=\{x \; \mid \; f(x)\leq0\}$. For standard neural network architectures, these spaces have a finite amount of topological holes and the number of holes depends on the activation function and the number of layers. For the ReLU activation function, the number of topological holes is in $O(2^l)$ where $l$ is the number of layers \cite{bianchini2014complexity}. $\grp$-invariance switches the domain from $\Omega$ to $\orbit(\Omega)$. The latter never has more topological holes than the former and can have drastically less topological holes than the former.}
\section{Generalization Properties of Invariant Neural Networks}
While the prior two sections dealt with the expressivity and learning procedure of $\grp$-invariant neural networks, i.e. the concrete mathematical optimization, we will now look at their statistical generalization properties. As argued in the beginning of this section, traditional statistical analyses struggle with justifying the generalization performance of neural networks. Thus, in this section we will refer to bayesian PAC techniques. Like traditional statistical approaches, bayesian PAC techniques try to give a bound on the actual risk over the entire dataset given the risk that was achieved on the training set. The advantage of bayesian PAC techniques is, that they allow for ''custom'' bounds that the user can improve, if they have domain knowledge to leverage.\\	
This section presents results achieved by \cite{lyle2020benefits}.\\
First and foremost, it is important to note that the results of \cite{lyle2020benefits} are attained with a statistical mindset and thus reframe the estimation problem as follows:
\begin{center}
Given some samples of data $d \in X \times T \sim D$ for inputs $X$ and labels $T$ where $D$ is the distribution generating the data, find the (probabilistic) estimator $f$ belonging to some estimator class $F$ such that the risk $R(f,D)=\Ex_{d \sim D}(l(f,d))$ is minimized.
\end{center}
Note that this assumes that neural networks model probability distributions in some way. While this differs from the way neural networks are usually presented in computer science settings and is indeed a restriction on a neural network's function, it is applicable for almost all classification architectures and in fact most loss functions used for neural networks are both derived from this point of view and directly let us view the neural network as a probability distribution. Further note that, as $D$ is the joint distribution of both data samples and corresponding labels, it also implicitly encodes the density of samples $x \in X$ as the marginal likelihood
\begin{equation}
D(x)=\int_{t \in T} D(x,t) dt
\end{equation}
and the conditional density of $t$ given $x$
\begin{equation}
D(y|x) = \frac{D(x,y)}{D(x)}
\end{equation}
Of course, while we seek to minimize the actual risk $R(f,D)$, it is almost never actually known. Hence, the empiric risk over the training set $\hat{D}$ is optimized instead:
\begin{equation}
\hat{R}(f,D) = \frac{1}{n} \Sigma_{(x,t) \in \hat{D}} l(x,t)
\end{equation}  
 \cite{vapnik2013nature}. For example, consider the function space of all possible functions $X \rightarrow T$. Of course, one could, given a sensible training set, define a probability distribution $f(x,y)$ such that $p(y|x)=1$\footnote{Note the difference of $f$ and $p$ in this case. $f$ refers to the (continuous or discrete) distribution, whereas $p$ refers to actual probability. In discrete distributions, $p(x)=1$ implies $f(x)=1$, but in continuous distributions $p(x)=1$ implies $f(x)= \infty$ as continuous measures differ from discrete measures.} $\leftrightarrow (x,y) \in D$ and trivially attain $\hat{R(f,D)}=0$ whereas the actual risk could still be arbitrarily high. This is, of course, due to the function class $F$ being entirely unrestricted.\\
As $\grp$-invariance encodes a restriction on the function class $F$ to which our estimator belongs, it seems intuitively plausible, that this restriction would yield an improved generalization error.\\
To attain their results, the authors of \cite{lyle2020benefits} consider symmetrized estimators $f^{\circ}$ for some non $\grp$-invariant estimator $f$. Here, $f^{\circ}$ is given by:
\begin{equation}
f^{\circ}(x)= \int_{g \in \grp} f(gx) d \lambda(g) = \Ex_{g ~ \lambda} [f(gx)]
\end{equation}
Intuitively, $f^{\circ}$ is a transformed version of the original estimator $f$, averaged over the orbits of each element. This ensures $\grp$-invariance (check that the integral in the above equation is equivalent for each $y \in \orbit(x)$) and the restriction to functions of the form $f^{\circ}$ entails just the $\grp$-invariant functions of $F$, assuming $f^{\circ} \in F$. Note also that this resembles a possible invariant final layer in the GDL blueprint. In fact, it is proposed in that work \cite{geo} to construct final layers in just that way.\\
The authors then derive a number of interesting results, the first ones pertaining to the risk $R$ and its empiric counterpart $\hat{R}$.  First, they show that for all functions $f \in F$ and all convex loss functions $l$ it holds that
\begin{equation}
\hat{R}(f^{\circ}, D^N) \leq R(f, D^N)
\end{equation}
showing that for each non-restricted function $f$, the empiric risk can be reduced via symmetrization. This is in itself interesting, but not too helpful in the context of neural networks. Due to the scale of modern neural networks, it is generally possible to get $\hat{R}$ arbitrarily close to zero, with or without $\grp$-invariance. Note that all of these results assume $l$ to be convex. This is a limitation of these theorems, but still applicable to almost all neural network architectures as deep learning losses are designed to be convex \cite{dl}.\\
More interestingly, they show that this inequality extends to the variance of $\hat{R}$, assuming that said variance is finite:
\begin{equation}
Var[\hat{R}(f^{\circ}, D^N)] \leq Var[R(f, D^N)]
\end{equation}
Implying that symmetrization of $f$ yields a better estimation of the risk. This has two major implications. First off, the estimated error over the testset is more accurate, as the variance is reduced. More importantly however, during learning, when $f$ is not yet entirely fitted to the training set, the estimate of the gradient $\nabla_{\Theta} R$ is improved as well, as $\hat{R}(f^{\circ}, D^N)$ more closely matches the true value of $R(f^{\circ}, D^N)$ . Intuitively, this implies that the learning process for $\grp$-invariant functions should make steps that are more closely aligned with improvement of the actual, global risk and not just the empirical risk.\\
It is notable however, that none of these inequalities are strict and there is little notion of how much estimates are improved. Rather, it could only be shown that some certain metrics do not worsen after symmetrization.\\
A more accurate improvement can be shown to affect a PAC-Bayes Bound of the estimator. A PAC-Bayes Bound is a bound on the actual risk of an estimator that is probably almost accurate (PAC), i.e. a bound $R(f, D) \leq \hat{R}(f, D^N) + K	$ that holds with arbitrarily high probability if $K$ is increased. Note that the term PAC-Bayes Bound is just a term for a family of bounds rather than one specific bound in particular. It is also important to note that PAC-Bayesian methods refer to Gibbs-estimation regimes \cite{catoni2007pac}. Gibbs estimation assumes that not one singular classifier is trained, but rather a distribution $Q$ over the class of possible classifiers $F$ is trained. This is not immediately applicable to neural networks as neural networks generally encode a singular classifier, but an extension can easily be achieved by adding random noise to its parameters or by choosing $Q(f)$ as the probability of choosing initial parameters $\theta_0$ such that $f$ is the estimator attained by gradient based optimization starting at $\theta_0$. Gibbs estimation is sometimes more easily theoretically handled and also accounts for the very much present randomness in neural network training, usually induced by initialization or early stopping (which can be understood as random noise added to the final weights). \\
The authors use the PAC Bound of \cite{catoni2007pac} which states the following:
\begin{center}
If $D^n$ is i.i.d. sampled according to some data generating distribution $P_D$, $Q$ is a posterior distribution over possible estimators $f \in F$ then for all prior distributions $P$ and all $\delta \in (0,1)$ and $\beta >0$ it holds with probability at least $1-\delta$ over all possible samples that 
\begin{equation}
\label{catoni}
R(Q) \leq \frac{1- e^{-\beta \hat{R}(Q,D^N)- \frac{1}{n} ( KL(Q||P)+\log \frac{1}{\delta} }}{1-e^{-\beta}}
\end{equation}
\end{center}
$KL(Q||P)$ denotes the Kullback-Leibler divergence, a function that measures the distance between two probability distributions $Q$ and $P$.\\	
As bayesian PAC bounds are notoriously unintuitive, we briefly review this bound and give some intuition as to why it makes sense.\\
First off, note that the above bound is more easily interpreted as a statistical experiment. If one chooses an arbitrary $P$, $\beta$, $\delta$ and samples $D^n$ from $P_d$, then trains the estimator distribution $Q$ and calculates this bound, there exists a $1-\delta$ chance of the bound being accurate.\\
It is very important to note that due to its statistical nature this bound is very sensitive to the order in which each component is defined. Therefore, it is not admissible to calculate this bound for some estimator class and later on choose $P$ to minimize this bound, it has to be chosen without knowledge of everything else. Otherwise, one could simply choose $P=Q$ to set $KL(P||Q)=0$. This is, of course, forbidden as $Q$ depends on itself.\\
Next, note that the prior distribution $P$ behaves very interestingly in this bound. It occurs only in the term $KL(Q||P)$ and the final distribution $Q$ does not actually depend on it, as would usually be indicated by the term "prior". Instead, it acts more as a generic measure on $F$ and is thus interchangable with all possible distributions. Intuitively, $P$ acts as a reference distribution that is chosen before the training of $Q$. This allows for PAC bayesian bounds to 'cheat' the limitations of traditional statistical approaches using domain knowledge. $P$ can be defined by an expert user that has knowledge of the domain of interest. While $P$ has to be chosen before $Q$ is trained, if one \textit{knows}, what kind of estimators are plausible for the given task, one can choose a fitting $P$ that yields a tight bound for $R(Q)$.\\
In essence, the idea behind traditional approaches is: ''If the classifier can only fit a limited amount of functions and if $f^*$ does not belong to those functions, then it is unlikely that we draw samples that $f$ can fit''. However, for bayesian PAC approaches, the idea is: ''If we train an estimator on data, then the empirical risk does not estimate the true risk as the estimator is dependent on the data. However, if we can correctly guess the final estimator before seeing the data, then this dependence does not exist and therefore the true risk is estimated well by our empiric risk.\\ In this formulation, human expertise is involved in guessing the final estimator that results after training the data. Thus, PAC bayesian approaches can use domain knowledge to give usable bounds for neural network estimators.\\
Interpreting the top part of the fraction, one can observe that the upper bound is increasing in the KL-term and the empiric risk. $\beta$ dictates how strongly the empiric risk influences the bound, whereas the influence of the $KL$ term is determined by the factor $\frac{1}{n}$, leading credence to the idea that generalization is more reliable if a greater $n$ is observed. Lastly, the bound is decreasing in $\delta$ in a logarithmic fashion which is intuitive as an increasing $\delta$ also lowers the probability of this bound holding.\\
As the authors of \cite{lyle2020benefits} stress and as is shown by \cite{zhang2021understanding}, neural networks, even if regularized, trivially fit even the most absurd sets of training data nigh perfectly. As a consequence, the empirical risk is usually (close to) 0 and thus negligible. The interesting part of this equation is thus the $KL$-term that directly encodes the assumed generalization error with respect to $P$.\\
This term specifically can be improved upon by $\grp$-symmetrizing both $Q$ and $P$ \footnote{Note that $P$ should probably be already chosen symmetrically already in any meaningful setting that is assumed $\grp$-invariant as one can make use of this domain knowledge while choosing $P$.}. While the actual proof given in \cite{lyle2020benefits} is cumbersome, the intuition behind this is relatively simple.\\
Symmetrizing the distribution $Q$ entails transforming $Q$ from a distribution over $F$ to a distribution $Q^{\circ}$ over $F^{\circ}$, the symmetrized estimators, where $Q^{\circ}$ is given by
\begin{equation}
Q^{\circ}(g)= \int_{f \in F, f^{\circ}=g} Q(f) df
\end{equation}  
i.e. $Q^{\circ}$ sums the probabilities for all $f$ that are identical under $\grp$-symmetrization. This intuitively allows for less differences in $Q^{\circ}$ and $F^{\circ}$, as the disagreements of $P$ and $Q$ on $\grp$-equivalent estimators are integrated out.\\
Further, the authors of \cite{lyle2020benefits} manage to show a concrete bound that defines the difference in KL terms before and after symmetrization. This so-called symmetrization gap is given by:
\begin{equation}
 \Delta(Q||P)= KL(Q||P)-KL(Q^{\circ} || P^{\circ}) = \Ex _{f \sim Q} [\log \frac{q(f)}{q^{\circ}(f^{\circ})}]
\end{equation}
Note that this gives credence to the intuition behind the KL-reduction given above, the difference between the KL-terms is determined by the expected logarithmic difference in $q(f)$ and $q^{\circ}(f^{\circ})$, representing the difference between the integrated probabilities of all $q(gf)$ for $g \in \orbit(f)$ and $q(f)$.\\
This result gives a very clear indication as to how $\grp$-invariance benefits neural network performance. The reduction in generalization in the bound of \ref{catoni} stems from a reduction in the KL term. This reduction in the KL term is attained via a reduction in the space of estimators, abstracting from the concrete estimators to their symmetrized variants, thus giving an improvement in the KL term which encodes the expressive strength of the estimator.
\mybox{$\grp$-Invariance and Statistical Generalization Properties}{
	If one chooses a distribution $P$ over possible estimators $f$ before seeing the data, chooses scalar values $\delta \in [0,1]$, $\beta>0$ and then attains through training a distribution over estimators $P$, then with a probability of $1-\delta$:
\begin{equation}
\label{catoni}
R(Q) \leq \frac{1- e^{-\beta \hat{R}(Q,D^N)- \frac{1}{n} ( KL(Q||P)+\log \frac{1}{\delta} }}{1-e^{-\beta}}
\end{equation}
If one knows beforehand, that the final estimator $f$ and the function $f^*$ to approximate are $\grp$-invariant, then this bound is improved by symmetrizing the distributions $Q$ and $P$ (i.e., averaging over all estimators that are equivalent if made symmetrical). This improvement occurs in the KL-term and is equal to:
\begin{equation}
 \Delta(Q||P)= KL(Q||P)-KL(Q^{\circ} || P^{\circ}) = \Ex _{f \sim Q} [\log \frac{q(f)}{q^{\circ}(f^{\circ})}]
\end{equation}
This applies to all estimators and, if one attains a good guess of $Q$ that closely matches the final estimator distribution $P$, these bounds can be relatively tight.
}
\section{Summary}
This section set out to examine the impact of $\grp$-Invariance as an example of the impact of the ideas of geometric deep learning for domain specific networks.\\
 Interestingly, it could be shown that there are at least three different perspectives from which the impact of $\grp$-invariance can be evaluated, each yielding a different perspective as to how neural networks benefit from incorporated $\grp$-invariance.\\
First off, it could be shown that $\grp$-invariance changes the training process induced via gradient based optimization. Due to the nature of gradient based optimizers, points where the gradients $\nabla_{\theta} f(x)$ are similar, are changed in equivalent ways. For unrestrained neural networks, this implies that euclidean neighbors are assumed to be similar in output. For $\grp$-invariant neural networks optimization works differently, instead considering the minimum distance between the orbits of points. This yields, for most problems, a much more sensible notion of distance than just euclidean distance, giving credence to the idea that invariant networks both train more easily and generalize better on most tasks of interest.\\
This relates to the idea of continuity in fully connected neural networks: In fully connected networks, datapoints $x$ and $y$ have similar values $f(x) \approx f(y)$ if they are close in a euclidean sense. In a $\grp$-invariant network, this holds if $x$ is close to $gy$ for any $g \in \grp$, which shows, that $\grp$-invariant neural networks use domain structure to extend learning further.\\
Next is the topological view on neural networks, examining topological qualities of the spaces induced by the networks classification function. Here, research indicates that functions can only be approximated well by neural networks if the datapoints belonging to each class form a simple topological space with little topological holes and that stronger networks are needed to approximate more complex spaces with more holes \cite{bianchini2014complexity, guss2018characterizing}. While it is hard to give a general expression for the change of topological complexity through $\grp$-invariance, the number of holes never increases by enforcing $\grp$-invariance and it could be shown by example that a very considerable reduction in complexity is possible.\\
Lastly, the statistical point of view is concerned with the statistical properties of estimations done in training and evaluation of a neural network. As shown in the work of \cite{lyle2020benefits}, risk estimation can be shown to improve the theoretical bounds for the generalization error attained by the neural network, independent on the prior that one uses.\\
These results show that, no matter through what lense one views deep learning, incorporating domain specific structure, specifically invariance properties, almost always improves generalization in a quantifiable way, giving credence to the idea that incorporating invariance into neural networks is heavily required. Moreover, the effects can be extremely drastical, as seen in the case of learning a function involving the modulo operator. Even a simple function such as  $ x \mod 3>1$ becomes impossible to learn for even large neural networks if one does not enforce invariance in the neural network. Thus, it seems not the case that enforcing domain knowledge into neural network is merely a way of improving the learning procedure, rather, it appears that learning without domain knowledge is doomed to fail for almost every domain that is sufficiently complex and structured.

%% file: kapitel/localpooling.tex
\relax 

%% file: kapitel/conclusion.tex
\chapter{Summary, Conclusion and Outlook}
In this work, we set out to evaluate the ideas of geometric deep learning, a first attempt at structurally classifying deep learning architectures from a domain specific point of view. The questions that we sought to answer using the ideas of geometric deep learning were: ''What should domain specific neural networks look like, how can we construct them and what is their impact?''.\\
We started out by defining the scope of this work, restricting it to the use case of function approximation using gradient descent as a training algorithm. Then, we examined the properties of fully connected neural networks, the original, non-domain specific version of neural networks. We found that fully connected networks have a number of convenient properties that good estimators should possess. Those are completeness, i.e. the ability to approximate all functions of interest to arbitrary degree, assuming enough neurons and layers are present, continuity and tunability, i.e. the estimator is lipschitz continuous and the degree to which it is can be influenced by choosing additional layers or neurons.\\
However, we also argued that fully connected neural networks are not able to use additional structure in a domain and therefore can be improved by adding additional domain knowledge.\\
At this point, we took the ideas of geometric deep learning and used them to add to the domain specific neural network ''wishlist''. Geometric deep learning proposes three properties that a domain specific neural network should have: First, if there exists a group of transformations in the domain that should leave classification unchanged, the final estimator should be invariant to those transformations. An example of such transformations are graph isomorphisms that change the representation of a graph but not its fundamental properties. Second, if there exists a notion of locality in the domain, local structures should be more closely connected in the network architecture. Third, if we assume that the true function $f^*$ has a  compositional structure, then our estimator should follow this structure.\\
We combined those ideas with the useful properties of fully connected neural networks and attained a new list of desirable properties for domain specific neural networks. Afterwards, we used this list to derive deep sets, a very simple neural network architecture for the domain of sets. The example of deep sets showed how one can use these properties to justify an already existing, widely used domain specific neural network but also showed that, even if one knows what properties a domain specific neural network should have, it is still difficult to actually construct it.\\
Based on deep sets, we introduced the geometric deep learning blueprint. An approach to creating domain specific neural network that was proposed by geometric deep learning.\\
Afterwards, we applied these ideas to graph neural networks as a case study to showcase how these ideas are reflected in a more complicated, modern domain specific neural network. We could show that, again, graph neural networks adhere to our domain specific  network desiderata. However, while trying to apply the geometric deep learning blueprint to graph neural networks, we found it lacking due to the many degrees of freedom that it leaves unanswered.\\
Finally, we evaluated the impact of enforced invariance to a group $\grp$ of transformations on an estimator's performance. There, we could see that enforcing $\grp$-invariance on an estimator can, under the right circumstances, drastically improve performance, especially with regards to very complex tasks. 
\section{Conclusion}
As set out earlier in this work, domain specific neural network architectures are one of the most important developments in modern deep learning and appear to be a large reason to its success. We could observe this effect very strongly on the example of $\grp$-invariance. As was shown in Chapter 4, not enforcing $\grp$-invariance in a domain that exhibits this property can make even simple problems extremely complex and it stands to reason that other domain specific properties can have similarly strong effects.\\
While the effectiveness of domain specific neural networks cannot be denied, constructing them, even using the ideas of geometric deep learning, still seems hard.\\
Combining the domain specific ideas of geometric deep learning and combining them with the convenient properties of fully connected neural networks yields a surprisingly useful characterization of domain specific neural networks. In this work, we showed that both deep sets and graph neural networks adhere not only to the ideas of geometric deep learning (as was also shown in \cite{geo}), but also retain the strong points of fully connected neural networks. Therefore, they entirely fulfill our domain specific neural network desiderata\footnote{In the case of graph neural networks, they do violate the completeness property in the sense that they can approximate all functions of interest. However, due to the complexity of deciding graph isomorphism, this can hardly be expected.}. The ideas of geometric deep learning proved especially useful here, for both deep sets and graph neural networks, invariance properties and locality are very noticable characteristics of the architecture.\\
However, as for the actual construction of such architectures, the geometric deep learning blueprint, the geometric deep learning approach to construction, seemed very lacking. As lined out in Chapter 3, in the case of graph neural networks, it leaves too many freedoms to the designer and still requires just as much expertise to build an actually useful network architecture than if one were to do it manually using the desiderata that were set out by geometric deep learning.\\
Thus, it appears that geometric deep learning accurately describes some useful domain properties that a domain specific neural network architecture can exploit and by what properties of the estimator they \textit{can} be exploited. However, as for construction of domain specific neural networks, there seems much work to be done still.\\
In addition, it stands to note that the domain properties used in geometric deep learning are very focused on domains that are traditionally domains of geometric deep learning. Most notably, they implicitly focus strongly on domains that are similar to image recognition, i.e. inputs are given as vectors $x \in \real^k$ of real numbers, with some relationships existing between the different entries of said vector. For these ideas, geometric deep learning should work well. Still, for different domains, they might not apply. For example, in discrete domains without real valued inputs, deep learning still relies on using work-arounds and using continuous models to approximate discrete processes, which is of course not optimal. Thus, while geometric deep learning makes some very useful contributions, they are still restricted to more standard usecases of deep learning and are thus not necessarily complete.\\
This is of course natural, as geometric deep learning is just one first step into the direction of structurally designing domain specific neural network architectures and cannot be expected to give answers to every possible question at this point. 
\section{Outlook}
While the ideas of geometric deep learning prove to be a good first step towards the creation of domain specific neural networks, much research is still lacking.\\
First off, geometric deep learning does not offer solutions to all of the problems encountered by neural networks. In fact, geometric deep learning is restricted to domains with real valued domains that possess some additional structure such as images and graphs with real valued node labels. In practice, deep learning is used in many fully discrete fields. For example, in language recognition, words are discrete objects. As was shown in Chapter 5, geometric deep learning cannot circumvent the fact, that neural networks operate on euclidean spaces and thus struggle with discrete tasks. To extend neural networks to fully discrete domains in a conceptually sound way, further research needs to be done.\\
Second, while geometric deep learning does, for the domains that is applicable to, provide a useful set of criteria as to what characterizes a good domain specific neural network architecture, the actual construction of said neural network architectures is still very hard and the geometric deep learning blueprint does not sufficiently structure it. Therefore, additional research into more strongly structured ways of creating domain specific neural network architectures is needed.

%% file: kapitel/anhang.tex
\chapter{Appendix}
\section{A. Groups and Symmetries}
One of the most crucial ideas of \cite{geo} concerns the idea of symmetries that exist in certain forms of data, i.e. transformations of some data object that leaves the important characteristics of said data intact. For example, in graphs, if one reorders the nodes of some graph, the resulting graph behaves equivalently for most tasks of interest. Similarly, in image processing, if an image is shifted by a certain amount of pixels (assuming sufficient whitespace around it), the object represented by the image should stay the same.\\
These transformations can be described mathematically as groups and, more specifically, as subgroups of the symmetric group\cite{grouptheory}, providing the necessary mathematical generalization.\\  
Formally, a group is defined as a tuple $\langle \grp, \cdot\rangle$ containing a set $\grp$, associated with a binary operator $\cdot : \grp \times \grp \rightarrow \grp$ that fulfills the following three properties \footnote{When referring to a group $\langle\grp,\cdot \rangle$ we will frequently refer to just the set $\grp$ and assume that the corresponding operator is clear from context.}. First off, $\grp$ needs to be associative:
\begin{equation}
a\cdot ( b\cdot c) = ( a \cdot b) \cdot c
\end{equation}
Associativity is a frequently assumed property of algebraic structures and allows for brackets to be omitted. Further, and more interestingly, groups require a neutral identity  element $id \in \grp$ that leaves others unchanged:
\begin{equation}
id \cdot x = x \cdot id = x
\end{equation}
Note that the neutral element is both left- and right-neutral, i.e. no matter whether it is applied as the left or right side of the operation, it behaves equally in a neutral way. This is special to the neutral element as groups do not generally commute (commutative groups are called \textit{abelian}). Lastly, groups are required to contain inverses for each of their elements. For each $g \in \grp$ there exists $g^{-1}$ such that
\begin{equation}
g \cdot g^{-1} = g^{-1} \cdot g = id
\end{equation}
It directly follows from this definition that the inverse of $g^{-1}$ is again $g$ and thus $(g^{-1})^{-1} = g$. \\
Further, if $\langle \grp, \cdot\rangle$ is a group, any subset $G \subseteq \grp$ that also forms a group $\langle G, \cdot\rangle$ with the operation $\cdot$ is called a subgroup of $\grp$.\\
As previously noted, the most important group for the context of this work is the symmetric group, $S(\Omega)$ and its subgroups, the permutation groups, for some set $\Omega$. The symmetric group is defined as 
\begin{equation}
S(X) := \langle\{\pi \; \mid \; \pi \; : \; \Omega \rightarrow \Omega, \text{$\pi$ bijective}  \}, \circ \rangle
\end{equation}
the set of all possible bijections, or permutations, of $X$ onto itself, associated with the binary operator of function composition. It is easy to check that this fulfills the conditions of a group as the composition of two bijections is again a bijection (closedness), the identity function $f(x)=x$ is a bijection (neutral element) and bijective functions are fundamentally invertible (inverse elements).\\
Symmetric groups have been the object of much prior research and they, or more specifically their subgroups, the permutations groups, cover almost all groups of interest for this work and also possess some interesting properties.\\
First off, for finite sets $\Omega=\{ x_1, x_2,...,x_n\}$, the symmetric group is generated by the set of adjacent transpositions. An adjacent transposition $\sigma_i: \Omega \rightarrow \Omega$ is the bijection that swaps only the i-th and i+1-th element of $\Omega$, defined by
\begin{equation}
\sigma_i(x_i)=x_{i+1}
\end{equation}
\begin{equation}
\sigma_i(x_{i+1})=x_i
\end{equation}
\begin{equation}
\sigma_i(x)= x  \text{ if } x \neq x_i, x \neq x_{i+1}
\end{equation}
Therefore, for any homeomorphism on $S(\Omega)$, we can derive its entire behavior simply from how it acts on the generating set of $S(X)$.\\
Lastly, let us inspect some symmetries on data that were mentioned previously and check that they correspond to some subgroup of $S(X)$.\\
First off, let us inspect images that are translated by some number of pixels. We represent images as a set of pixels $P=\{1,2,...,n\}^2$. Translations to the right are then generated by the permutation that shifts every pixel by one pixel to the right\footnote{To ensure bijectivity, we assume that the pixels behave periodically, i.e. the n+1-th pixel is equivalent to the first pixel again.}
\begin{equation}
\pi_r((i,j))=(i+1 \mod n,j)
\end{equation}
For the other directions, this is equivalent. Then, the set of all translation is generated by the one-pixel translation for each direction. It is easy to check that this set fulfills the conditions of a group and therefore is a subgroup of $S(P)$ and thus a permutation group.\\
Second, consider the case of graphs $G=(V,E)$ and reordering of their nodes. Here, all permutations of that set are considered and therefore the subgroup of interest is in fact $S(V)$. Both of these are prime examples of symmetries that are considered in \cite{geo}.
\subsection{Orbits and Equivalence Classes}
Having introduced the symmetric group, we are now interested in the structure that it and its subgroups induce on the set $\Omega$\footnote{Note that most of the content of this section extends to arbitrary groups acting on some domain, but we are interested mainly in the symmetric group and its subgroups.}.\\
As noted previously, we are heavily concerned with the ideas of symmetries, i.e. transformations that leave intact the properties of interest, that are given as elements of some permutation group, i.e. a subgroup of the symmetric group. Therefore, we assume that in some way, shape or form there exists some equivalence between a set $\Omega$ and its image, $p(X)$.
This is formalized as the orbit of $x$:
\begin{equation}
\orbit(x)= \{gx \mid g \in \grp \}
\end{equation}
The orbit of $x$ includes all elements that are $\grp$-transformations of $x$. As an example, consider the group of rotations acting on 2d images. Here, the orbit of an image consists of all of its possible rotated versions. It is also easy to see in this example, that for an element $y \in \orbit(x)$ it holds that $\orbit(y)=\orbit(x)$ as the rotated versions of an image $y$ that is a rotated version of $x$ are again rotated versions of $x$. This holds universally as:
\begin{equation}
y \in \orbit(x)\implies y = gx \Leftrightarrow g^{-1}y =x
\end{equation}
Therefore, each $z \in \orbit(x)$ can be written as
\begin{equation}
z = g'x = (g' g^{-1}) y \in \orbit(y) 
\end{equation}
The converse can be shown in an equivalent way.\\
Interpreting orbits as a relation, the above clearly shows that the orbit relation
\begin{equation}
x \gcong y \Leftrightarrow \orbit(x) = \orbit(y)
\end{equation}
is symmetric. Similarly to the proof above one can show its transitivity and reflexivity, showing that the orbits indeed represent an equivalence relation. For the remainder of this work, elements within the same orbit will thus be called $\grp$-equivalent.\\
\section{B. Neural Networks in Discrete Domains}
Neural networks, even domain specific ones that are described in \cite{geo}, are made for domains that are naturally continuous. In all of the domains that are considered there, the estimation problem can be posed as:
\begin{center}
Estimate a continuous function $f^* : \real^k \rightarrow \real^j$ for some $k,j \in \mathbb{N}$. Further, assume that there exists some known properties $P$ that are fulfilled by $f^*$.
\end{center}
The estimation problem is always a real-valued estimation problem where both domain and co-domain of $f^*$ are real. The domain specificity here is characterized by the properties $P$ that are assumed for $f^*$. However, in many use cases of deep learning, this is not sufficient.\\
Consider image recognition. Usually, in image recognition, the neural network is tasked with assigning a label to each image. Images can be represented as vectors of real numbers and are therefore compatible to the above estimation problem. However, labels for images are discrete and unordered. Therefore, they cannot be represented by real values and if one did so, one would add notions of distance and ordering to the labels that are not actually present. This estimation problem has a discrete co-domain and thus is not immediately suited for fully connected neural networks and gradient descent. For example, consider the case where one represented the label ''dog'' as a $1$,''cat'' as a $2$ and the label ''car'' as a $3$. Then suppose that an image has label ''car'' and the neural networks outputs the value $0.9$, leading to the prediction of ''dog''. Then, gradient descent with an infinitesimal learning rate changes the prediction of the network by a small fraction towards $3$. Given a small enough learning rate, the neural network would now output $1$ for the same image. Thus, its output has gotten even closer to ''dog'' which was the misestimation in the first place. Therefore, gradient descent does not extend well to a naive, real-valued encoding of the output.\\
Consider the task of language processing. In language processing, the inputs are words. Words are also fundamentally discrete objects and thus one has to estimate a function $\text{words} \rightarrow C$. This makes learning hard. All neural networks that are described in this work learn by an assumption of continuity. That means, they assume that if some datapoint $x$ has output $o$, then its euclidean neighborhood has a similar output. Of course, for words there does not immediately exist a euclidean distance that can be used for learning.\\
We now just presented two examples of problematic estimation problems, one with discrete inputs and one with discrete outputs. Both language processing and image processing are premier use cases of deep learning \cite{pak2017review,young2018recent}. In this section, we will give an overview of the difficulties of using deep learning for function approximation where the input or output are discrete valued and the heuristics that are usually used in deep learning to overcome them.
\subsection{Discrete Inputs}
Suppose that we are tasked with the estimation of a function $f^*: W \rightarrow \real$ where $W$ is some set of discrete values. For this section, we will stick with words as an example for $W$ but note that everything in this section can be adapted to other use cases. As outlined before, this makes learning, in the way neural neworks usually do it, very hard. Thus, to use deep learning, there exist two challenges that one has to solve: First off, the inputs must be represented as vectors of real numbers. Fundamentally, to even use gradient based optimization, this has to be fulfilled. Second, this representation has to make sense for neural network learning. As a baseline, we require that inputs $x$ and $y$ be close in a euclidean sense if and only if we assume that learning from $x$ transfers to $y$.\\
Thus, the only way of (meaningfully) using deep learning for function approximation with discrete inputs is to construct a representation function $r: W \rightarrow \real^d$ for some $d$ with the additional constraint of $r(w) \approx r(v)$ if and only if we assume that $f^*(w) \approx r(v)$. These representations are called \textit{embeddings} and are used in various machine learning applications to solve the problem of discrete inputs \cite{li2018word,goyal2018graph}.\\
There are multiple way of going about this. For small sets $W$, one can manually decide what input words should be similar and what input words should not. Given a set of tuples of words that should be similar: $S=\{(w_1,v_1), ..., (w_n, v_n)\}$, one only has to solve the optimization problem:
\begin{equation}
\sum_{ (w,v) \in S)} \dist{r(w), r(v)} \rightarrow \max
\end{equation}
to attain suitable representations for each word.
Having these representations, one can create an estimator of the form
\begin{equation}
f(w)=f'(r(w))
\end{equation}
where $f'$ is a neural network $\real^d \rightarrow \real$ and use gradient based optimization.\\
This is the most simple version of word embeddings and is only feasible for small sets of words. For example, if one took the entire english language, it seems unfeasible that one could manually create an exhaustive list of all words that should be similar to one another.\\
For more larger input sets $W$, one needs another way of defining what words should be similar. One widely used example of this is word2vec \cite{mikolov2013distributed}.\\
In contrast to the previous example of word embeddings, word2vec is specifically made for words in text. Due to the complexity of this approach, we will simply sketch the idea behind it.\\
We start off by assigning arbitrary representations $r(w)$ to each word $w$ and will iteratively improve them. We assume that we have a large amount of example text. Then, we take random example wordstrings of a fixed, odd length $k$ of the example text. Let such a wordstring be
\begin{equation}
w_1 \quad w_2 \quad w_3... w_{mid} ...  \quad w_n
\end{equation}
with $w_{mid}$ being at the center of this wordstring. Then, we maximize the cosine-similarity between the representation of $w_{mid}$ and all of its surrounding words:
\begin{equation}
\sum_{i=1}^n \omega_i \frac{r(w_i)^T r(w_{mid})}{\dist{r(w_i)^T r(w_{mid})}} \rightarrow \max
\end{equation}
\begin{figure}[htbp]
\centerline{\includegraphics[scale=.5]{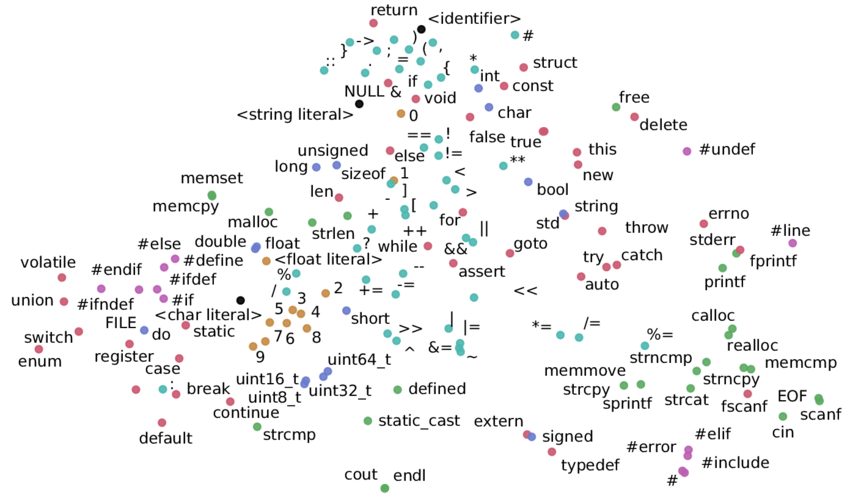}}
\caption{An example of word embeddings for the language C++, computed via word2vec and projected into two dimensions, image taken from \cite{harer2018automated}. Words are program statements. Note especially what statements are close to each other: Examples include ''return'' and ''identifier, which are frequently written together, and that operands such as addition, subtraction and comparators are each clustered together.  }
\end{figure}
where the weights $\omega_i$ are typically based on the distance of word $w_i$ from $w_{mid}$. Similarly, one can add additional terms that penalize similarity of $r(w_i)$ and all other words that are not in its neighborhoods. This is iteratively repeated. Word2vec achieves two things: First, words that frequently occur together have similar representations. This is directly enforced by the equation above. Second, words occur in similar sentences, also have similar representations. This is enforced by transitivity of the similarity. If it makes sense for us to assume that words that occur frequently together and that can be replaced by one another are similar with regards to $f^*$, then this representation makes sense. Of course, the word2vec algorithm can also be extended by additional constraints.\\
This approach is very popular for word embeddings, but does not immediately extend to other domains. Thus, while embedding approaches generally offer a good solution for discrete inputs in deep learning domains, it remains challenging to create a good embedding and good embeddings are heavily domain specific.
\subsection{Discrete Outputs}
Now consider that we are tasked with the estimation of a function $f^*: \real^k \rightarrow L$ where $L=\{l_1, l_2, ..., l_n\}$ is some finite set of labels. Here, the same problem applies: To use neural networks and gradient based optimization, we require the output domain to be some real valued space $\real^d$. Thus, we need to construct a model that outputs real values and a way of translating these outputs to the labels in $L$. Similarly to the case of discrete inputs, this needs to be constructed in such a way that it works with continuous models, i.e., we can reasonably assume that a small euclidean change in the inputs is associated with a small euclidea change in the outputs. \\
The most frequently used workaround to achieve this, is to estimate a probabilistic version of $f^*$. Here, we do not directly estimate $f^*$ but rather a probability distribution $P(f^*(x)=l_i)$ that denotes the probability of a datapoint $x$ having the label $l$. Our estimator $f$ takes as input a datapoint $x$ and returns one real value for each label $l\in L$:
\begin{equation}
f : \real^k \rightarrow \real^n
\end{equation} 
where $n=|L|$. We interpret the $i$-th component of $f(x)$ as the probability of $x$ having label $l_i$:
\begin{equation}
P(f^*(x)=l_i) := f(x)_i
\end{equation}
In our setting, $f^*$ is a deterministic function. Therefore, this follows the bayesian interpretation of probablity, describing not a truly uncertain event but a subjective degree of belief about the value of $f^*(x)$.\\
Of course this means that $f(x)$ must denote a proper probability distribution:
\begin{equation}
\sum_{i=1}^n f(x)_i =1
\end{equation}
\begin{equation}
f(x)>0
\end{equation}
for all $x$. Most estimators do not by default fulfill these properties. A common solution is to construct $f$ with a softmax function as the final operation\cite{dl}. The softmax function $\softmax: \real^d \rightarrow \real^d$ is defined by:
\begin{equation}
\softmax(a_1,..., a_n)_i= \frac{e^{a_i}}{\sum_{j=1}^n e^{a_j}}
\end{equation}
\begin{figure}[htbp]
\centerline{\includegraphics[scale=.5]{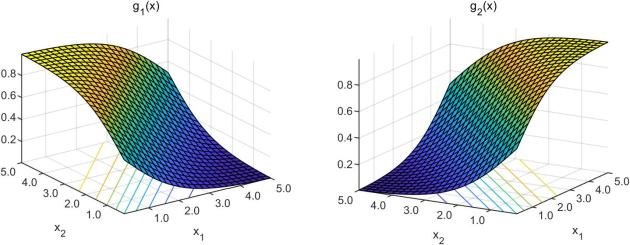}}
\caption{The softmax function plotted for two inputs.  }
\end{figure}
This ensures that each component of the result vector is positive by applying the exponential function and that it sums to one by normalizing it. Thus, if one builds an estimator of the form
\begin{equation}
f(x)=\softmax(f'(x))
\end{equation}
with $f':\real^k \rightarrow \real^n$, one attains an estimator that outputs a proper probability distribution over the labels of $x$. Furthermore, the softmax function works with the exponentials of $x$ and therefore $f'$ has to output the logarithmic label probabilities. This is convenient, as logarithmic spaces permit very natural distances between probabilities.\\
Of course, this also means that a new loss function is needed. As outlined in Chapter 2, we use as loss function the negative likelihood of the model $f$. For a dataset $D=\{(x_1, y_1), ..., (x_n, y_n)\}$, the log-likelihood of $f$ is given as
\begin{align*}
\log(P(D|f))&= \sum_{i=1}^n \log(P(f^*(x_i)=y_i))\\
&= \sum_{i=1}^n \log(f^*(x_i)_ {y_i})
\end{align*}
where $f^*(x_i)_{y_i}$ is the component of $f^*(x_i)$ that corresponds to $y_i$. This loss function is the cross entropy loss with deterministic labels, the most commonly used loss function in classification tasks~\cite{dl}.\\
In summary: To deal with discrete outputs, neural networks are made not to approximate $f^*(x)$, but a probability distribution over $f^*(x)$. This is achieved by building an estimator $\real^k\rightarrow \real^n$ and applying the softmax function to the output. This makes the estimation problem continuous and allows for the use of gradient based optimization algorithms. The final probability distribution can be used to do the actual classification by using the label with the highest probability.\\
There are some weaknesses to this construction. First, it uses probabilities to encode uncertainty about a function that is not in itself probabilistic. This corresponds to the bayesian interpretation of probability which is not uncontroversial in itself. Second, this only works for finite amounts of labels. Any infinite, discrete space such as the whole numbers cannot be handled in this way and these spaces prove difficult for use with neural networks. Third, the probabilities $P(f^*(x))$ do not actually exist and are entirely subjective. Therefore, the actual function to be estimated is not well defined which makes it hard to know whether $P(f^*(x))$ is suited for estimation with neural networks and also whether estimating it generalizes well. 